\documentclass[11pt,reqno,english]{article}
\usepackage{lmodern}

\usepackage[T1]{fontenc}
\usepackage[latin9]{inputenc}
\usepackage{color}
\usepackage{babel}
\usepackage{array}
\usepackage{mathrsfs}
\usepackage{multirow}
\usepackage{amsmath}
\usepackage{amsthm}
\usepackage{amssymb}
\usepackage{graphicx}
\usepackage[unicode=true,pdfusetitle,
 bookmarks=false,
 breaklinks=false,pdfborder={0 0 0},pdfborderstyle={},backref=false,colorlinks=true]
 {hyperref}
\hypersetup{
 citecolor=blue,linkcolor=purple}

\makeatletter

\providecommand{\tabularnewline}{\\}

\usepackage{paralist}
\usepackage{fullpage}
\usepackage{tikz}

\allowdisplaybreaks[3]

\@ifundefined{showcaptionsetup}{}{%
 \PassOptionsToPackage{caption=false}{subfig}}
\usepackage{subfig}
\makeatother

\begin{document}
\title{Mean Field Limit of the Learning Dynamics of\\Multilayer Neural Networks}
\author{Phan-Minh Nguyen\thanks{Department of Electrical Engineering, Stanford University.}}
\maketitle
\begin{abstract}
Can multilayer neural networks \textendash{} typically constructed
as highly complex structures with many nonlinearly activated neurons
across layers \textendash{} behave in a non-trivial way that yet simplifies
away a major part of their complexities? In this work, we uncover
a phenomenon in which the behavior of these complex networks \textendash{}
under suitable scalings and stochastic gradient descent dynamics \textendash{}
becomes independent of the number of neurons as this number grows
sufficiently large. We develop a formalism in which this many-neurons
limiting behavior is captured by a set of equations, thereby exposing
a previously unknown operating regime of these networks. While the
current pursuit is mathematically non-rigorous, it is complemented
with several experiments that validate the existence of this behavior.
\end{abstract}

\section{Introduction}

The breakthrough empirical success of deep learning \cite{lecun2015deep}
has spurred strong interests in theoretical understanding of multilayer
neural networks. Recent progresses have been made from various perspectives
within and beyond traditional learning-theoretic frameworks and tools
\textendash{} a very incomplete list of references includes \cite{arora2014provable,choromanska2015loss,mhaskar2016deep,safran2016quality,mallat2016understanding,schoenholz2016deep,zhong2017recovery,shwartz2017opening,soltanolkotabi2018theoretical,nguyen2018optimization,geiger2018jamming,ho2018neural,wei2018margin}.
Analyzing these networks is challenging due to their inherent complexities,
first as highly nonlinear structures, usually involving a large number
of neurons at each layer, and second as highly non-convex optimization
problems, typically solved by gradient-based learning rules without
strong guarantees. One question then arises: given such complex nature,
is it possible to obtain a succinct description of their behavior?

In this work, we show that under suitable scalings and stochastic
gradient descent (SGD) learning dynamics, the behavior of a multilayer
neural network tends to a non-trivial limit as its number of neurons
approaches infinity. We refer to this limit as the \textsl{mean field
(MF) limit}. In this limit, the complexity of the network becomes
independent of the number of neurons, and the network admits a simplified
description that depends only on other intrinsic characteristics,
such as the number of layers and the data distribution. Interestingly
this implies that two networks, which differ by the number of neurons
and hence the degree of over-parameterization, can perform almost
equally, so long as both have sufficiently many neurons. A similar
phenomenon has been recently discovered and studied in two-layers
neural networks \cite{mei2018mean,chizat2018,rotskoff2018neural,sirignano2018mean}.

\begin{figure}
\begin{centering}
\includegraphics[width=0.7\columnwidth]{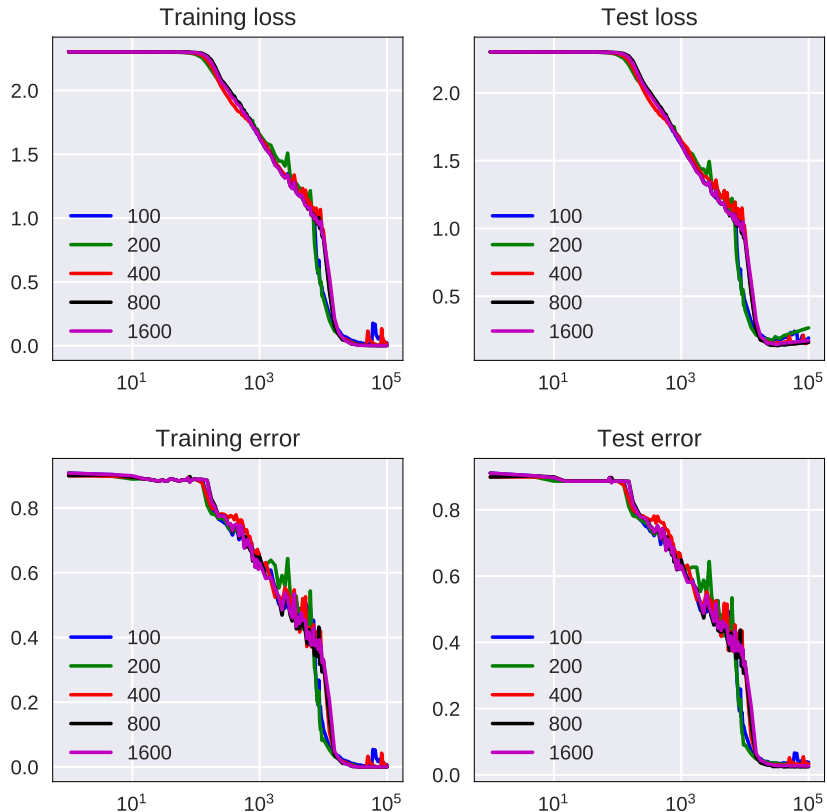}
\par\end{centering}
\caption{The performance of five $4$-layers fully-connected networks on MNIST
classification, plotted against training iterations. The number of
neurons at each hidden layer is 100, 200, 400, 800 or 1600, for each
network. Details are available in Section \ref{sec:Validation}.}
\label{fig:glimpse}
\end{figure}

To give a glimpse into the MF limit, in Fig. \ref{fig:glimpse}, we
plot the evolution of the performance of several $4$-layers networks,
each with a distinct number of neurons per layer, on the MNIST classification
task, under the chosen scalings. Observe how well the curves with
large numbers of neurons coincide, while the networks still achieve
non-trivial performance.

In the following, we shall give a motivating example via the two-layers
network case, before presenting the contributions of our work, discussing
related works and outlining the rest of the paper. Before we proceed,
we introduce some mathematical conventions.

\subsection{Notations, definitions and conventions}

We use boldface letters to denote vectors if lowercase (e.g. $\boldsymbol{x}$,
$\boldsymbol{\theta}$) and matrices if uppercase (e.g. $\boldsymbol{W}$).
For $n\in\mathbb{N}_{>0}$, we use $\left[n\right]$ to denote the
set $\left\{ 1,2,...,n\right\} $. For a scalar mapping $f:\;\mathbb{R}\mapsto\mathbb{R}$
and a vector $\boldsymbol{u}\in\mathbb{R}^{n}$, we use $f\left(\boldsymbol{u}\right)$
to denote $\left(f\left(u_{1}\right),...,f\left(u_{n}\right)\right)^{\top}$
entry-wise. For two vectors $\boldsymbol{u},\boldsymbol{v}\in\mathbb{R}^{n}$,
$\left\langle \boldsymbol{u},\boldsymbol{v}\right\rangle $ denotes
the usual Euclidean inner product. For a set $E\subseteq\mathbb{R}$
and $x\in\mathbb{R}$, we use $E+x$ to denote $\left\{ u+x:\;u\in E\right\} $.

We reserve the notation $\mathscr{P}\left(\Omega\right)$ for the
set of probability measures on the set $\Omega$. Strictly speaking,
one should associate $\Omega$ with a sigma-algebra to define $\mathscr{P}\left(\Omega\right)$.
We ignore this important technical fact in this paper. We will also
make use of the concept of stochastic kernels: $\nu$ is a stochastic
kernel\footnote{Not to be confused with a kernel function that is randomly generated.}
with a source set $\Omega$ (associated with a sigma-algebra ${\cal F}$)
and a target set $S$ (associated with a sigma-algebra ${\cal S}$)
if $\nu\left(\cdot\middle|\cdot\right)$ is a mapping ${\cal S}\times\Omega\mapsto\left[0,1\right]$
such that $\nu\left(\cdot\middle|\omega\right)\in\mathscr{P}\left(S\right)$
for each $\omega\in\Omega$ and $\nu\left(E\middle|\cdot\right)$
is ${\cal F}$-measurable for each $E\in{\cal S}$. We reserve the
notation $\mathscr{K}\left(\Omega,S\right)$ for the set of all such
kernels. For a stochastic kernel $\nu\in\mathscr{K}\left(\Omega,S\right)$,
we define
\[
{\rm CE}\left\{ \nu\right\} \left(\cdot\right)=\int_{S}x\nu\left({\rm d}x\middle|\cdot\right),
\]
i.e. the conditional expectation operator. We assume that ${\rm CE}\left\{ \nu\right\} $
exists almost everywhere for all stochastic kernels $\nu$ to be considered
in the paper.

We use ${\rm Emp}\left(\left\{ x_{i}\right\} _{i\in\left[n\right]}\right)$
to denote the empirical distribution $\left(1/n\right)\cdot\sum_{i=1}^{n}\delta_{x_{i}}$.
For a random variable $X$, the distributional law of $X$ is denoted
by ${\rm Law}\left(X\right)$. For a measure $\mu$, we use ${\rm supp}\left(\mu\right)$
to denote its support. A statement of the form $A\left(x\right)=B\left(x\right)$
for $\mu\text{-a.e. }x$ for a probability measure $\mu$ means that
\[
\int A\left(x\right)\phi\left(x\right)\mu\left({\rm d}x\right)=\int B\left(x\right)\phi\left(x\right)\mu\left({\rm d}x\right)
\]
for all smooth and bounded $\phi$.

For a functional $f:\;{\cal F}\mapsto\mathbb{R}$ on a suitable vector
space ${\cal F}$, we use $\mathscr{D}f$ to denote its differential:
for each $g\in{\cal F}$, $\mathscr{D}f\left\{ g\right\} $ is a linear
functional from ${\cal F}$ to $\mathbb{R}$ such that
\[
\lim_{\epsilon\to0}\frac{1}{\epsilon}\left(f\left(g+\epsilon\varphi\right)-f\left(g\right)\right)=\mathscr{D}f\left\{ g\right\} \left(\varphi\right)
\]
for $\varphi\in{\cal F}$. We shall ignore the fact that ${\cal F}$
is not arbitrary for this concept to apply.

A subscript in the differential operator ($\partial$, $\nabla$ or
$\mathscr{D}$) indicates the partial differentiation w.r.t. the respective
argument. For example, for $f\left(u,\boldsymbol{v},g\right):\;\mathbb{R}\times\mathbb{R}^{m}\times{\cal F}\mapsto\mathbb{R}$
where ${\cal F}$ is a set of functions, we use $\partial_{1}f$,
$\nabla_{2}f$ and $\mathscr{D}_{3}f$ (or $\partial_{u}f$, $\nabla_{\boldsymbol{v}}f$
and $\mathscr{D}_{g}f$ respectively) to denote its partial derivative
w.r.t. $u$, $\boldsymbol{v}$ and $g$ respectively.

We use $\left\Vert \boldsymbol{u}\right\Vert _{2}$ to denote the
Euclidean norm of a vector $\boldsymbol{u}$, $\left\Vert \boldsymbol{W}\right\Vert _{{\rm F}}$
the Frobenius norm of a matrix $\boldsymbol{W}$ and $\left\Vert f\right\Vert _{\infty}$
the max norm of a function $f$.

\subsection{A motivating example: two-layers neural networks\label{subsec:two-layers}}

\begin{figure}
\subfloat[]{\begin{centering}
\begin{minipage}[t]{0.5\columnwidth}%
\begin{center}
\begin{center}
\begin{tikzpicture}[yscale=-0.02,xscale=0.02]

\draw  [fill={rgb, 255:red, 155; green, 155; blue, 155 }  ,fill opacity=1 ] (50.08,107) .. controls (50.08,103.69) and (52.77,101) .. (56.08,101) -- (74.08,101) .. controls (77.4,101) and (80.08,103.69) .. (80.08,107) -- (80.08,244.08) .. controls (80.08,247.4) and (77.4,250.08) .. (74.08,250.08) -- (56.08,250.08) .. controls (52.77,250.08) and (50.08,247.4) .. (50.08,244.08) -- cycle ; 

\draw  [fill={rgb, 255:red, 255; green, 255; blue, 255 }  ,fill opacity=1 ] (210.67,174.67) .. controls (210.67,166.38) and (217.38,159.67) .. (225.67,159.67) .. controls (233.95,159.67) and (240.67,166.38) .. (240.67,174.67) .. controls (240.67,182.95) and (233.95,189.67) .. (225.67,189.67) .. controls (217.38,189.67) and (210.67,182.95) .. (210.67,174.67) -- cycle ; 

\draw    (160.67,174.67) -- (210.67,174.67) ;

\draw    (160.67,124.67) -- (210.67,174.67) ;

\draw    (210.67,174.67) -- (160.67,224) ;

\draw  [dash pattern={on 4.5pt off 4.5pt}]  (80.08,107) -- (130.67,124.67) ;

\draw  [dash pattern={on 4.5pt off 4.5pt}]  (80.08,244.08) -- (130.67,125.67) ;

\draw  [dash pattern={on 4.5pt off 4.5pt}]  (80.08,107) -- (130,175.67) ;

\draw  [dash pattern={on 4.5pt off 4.5pt}]  (80.08,244.08) -- (130,175.67) ;

\draw  [dash pattern={on 4.5pt off 4.5pt}]  (80.08,244.08) -- (130.67,225) ;

\draw  [dash pattern={on 4.5pt off 4.5pt}]  (80.08,107) -- (130.67,225) ;

\draw  [fill={rgb, 255:red, 155; green, 155; blue, 155 }  ,fill opacity=1 ] (130.67,124.67) .. controls (130.67,116.38) and (137.38,109.67) .. (145.67,109.67) .. controls (153.95,109.67) and (160.67,116.38) .. (160.67,124.67) .. controls (160.67,132.95) and (153.95,139.67) .. (145.67,139.67) .. controls (137.38,139.67) and (130.67,132.95) .. (130.67,124.67) -- cycle ; 

\draw  [fill={rgb, 255:red, 155; green, 155; blue, 155 }  ,fill opacity=1 ] (130,174.67) .. controls (130,166.38) and (136.72,159.67) .. (145,159.67) .. controls (153.28,159.67) and (160,166.38) .. (160,174.67) .. controls (160,182.95) and (153.28,189.67) .. (145,189.67) .. controls (136.72,189.67) and (130,182.95) .. (130,174.67) -- cycle ; 

\draw  [fill={rgb, 255:red, 155; green, 155; blue, 155 }  ,fill opacity=1 ] (130.67,224) .. controls (130.67,215.72) and (137.38,209) .. (145.67,209) .. controls (153.95,209) and (160.67,215.72) .. (160.67,224) .. controls (160.67,232.28) and (153.95,239) .. (145.67,239) .. controls (137.38,239) and (130.67,232.28) .. (130.67,224) -- cycle ;

\draw    (217.67,174.75) -- (234.5,174.67) ;

\draw    (226.21,183) -- (226.33,166.03) ;

\draw (65.5,262) node   {$\boldsymbol{x}$}; 

\draw (254.17,172.33) node   {$\hat{y}_n$}; 

\draw (149.5,94) node   {$\boldsymbol{\theta}_{j}$};
\end{tikzpicture} 
\par\end{center}
\par\end{center}%
\end{minipage}
\par\end{centering}
}\subfloat[]{\begin{centering}
\begin{minipage}[t]{0.5\columnwidth}%
\begin{center}
\begin{center}
\begin{tikzpicture}[yscale=-0.02,xscale=0.02] 

\draw  [fill={rgb, 255:red, 155; green, 155; blue, 155 }  ,fill opacity=1 ] (70.08,127) .. controls (70.08,123.69) and (72.77,121) .. (76.08,121) -- (94.08,121) .. controls (97.4,121) and (100.08,123.69) .. (100.08,127) -- (100.08,264.08) .. controls (100.08,267.4) and (97.4,270.08) .. (94.08,270.08) -- (76.08,270.08) .. controls (72.77,270.08) and (70.08,267.4) .. (70.08,264.08) -- cycle ; 

\draw  [fill={rgb, 255:red, 155; green, 155; blue, 155 }  ,fill opacity=1 ] (150.67,145.67) .. controls (150.67,137.38) and (157.38,130.67) .. (165.67,130.67) .. controls (173.95,130.67) and (180.67,137.38) .. (180.67,145.67) .. controls (180.67,153.95) and (173.95,160.67) .. (165.67,160.67) .. controls (157.38,160.67) and (150.67,153.95) .. (150.67,145.67) -- cycle ; 

\draw  [fill={rgb, 255:red, 155; green, 155; blue, 155 }  ,fill opacity=1 ] (150,195.67) .. controls (150,187.38) and (156.72,180.67) .. (165,180.67) .. controls (173.28,180.67) and (180,187.38) .. (180,195.67) .. controls (180,203.95) and (173.28,210.67) .. (165,210.67) .. controls (156.72,210.67) and (150,203.95) .. (150,195.67) -- cycle ; 

\draw  [fill={rgb, 255:red, 155; green, 155; blue, 155 }  ,fill opacity=1 ] (150.67,245) .. controls (150.67,236.72) and (157.38,230) .. (165.67,230) .. controls (173.95,230) and (180.67,236.72) .. (180.67,245) .. controls (180.67,253.28) and (173.95,260) .. (165.67,260) .. controls (157.38,260) and (150.67,253.28) .. (150.67,245) -- cycle ; 

\draw  [fill={rgb, 255:red, 255; green, 255; blue, 255 }  ,fill opacity=1 ] (311.67,195.67) .. controls (311.67,187.38) and (318.38,180.67) .. (326.67,180.67) .. controls (334.95,180.67) and (341.67,187.38) .. (341.67,195.67) .. controls (341.67,203.95) and (334.95,210.67) .. (326.67,210.67) .. controls (318.38,210.67) and (311.67,203.95) .. (311.67,195.67) -- cycle ; 

\draw    (261.67,195.67) -- (311.67,195.67) ;

\draw    (261.67,145.67) -- (311.67,195.67) ;

\draw    (311.67,195.67) -- (261.67,245) ;

\draw  [dash pattern={on 4.5pt off 4.5pt}]  (100.08,127) -- (150.67,144.67) ;

\draw  [dash pattern={on 4.5pt off 4.5pt}]  (100.08,264.08) -- (150.67,145.67) ;

\draw  [dash pattern={on 4.5pt off 4.5pt}]  (100.08,127) -- (150,195.67) ;

\draw  [dash pattern={on 4.5pt off 4.5pt}]  (100.08,264.08) -- (150,195.67) ;

\draw  [dash pattern={on 4.5pt off 4.5pt}]  (100.08,264.08) -- (150.67,245) ;

\draw  [dash pattern={on 4.5pt off 4.5pt}]  (100.08,127) -- (150.67,245) ;

\draw  [fill={rgb, 255:red, 155; green, 155; blue, 155 }  ,fill opacity=1 ] (231.67,145.67) .. controls (231.67,137.38) and (238.38,130.67) .. (246.67,130.67) .. controls (254.95,130.67) and (261.67,137.38) .. (261.67,145.67) .. controls (261.67,153.95) and (254.95,160.67) .. (246.67,160.67) .. controls (238.38,160.67) and (231.67,153.95) .. (231.67,145.67) -- cycle ; 

\draw  [fill={rgb, 255:red, 155; green, 155; blue, 155 }  ,fill opacity=1 ] (231,195.67) .. controls (231,187.38) and (237.72,180.67) .. (246,180.67) .. controls (254.28,180.67) and (261,187.38) .. (261,195.67) .. controls (261,203.95) and (254.28,210.67) .. (246,210.67) .. controls (237.72,210.67) and (231,203.95) .. (231,195.67) -- cycle ; 

\draw  [fill={rgb, 255:red, 155; green, 155; blue, 155 }  ,fill opacity=1 ] (231.67,245) .. controls (231.67,236.72) and (238.38,230) .. (246.67,230) .. controls (254.95,230) and (261.67,236.72) .. (261.67,245) .. controls (261.67,253.28) and (254.95,260) .. (246.67,260) .. controls (238.38,260) and (231.67,253.28) .. (231.67,245) -- cycle ;

\draw    (318.67,195.75) -- (335.5,195.67) ;

\draw    (327.21,204) -- (327.33,187.03) ;

\draw    (181.67,145.67) -- (231.67,145.67) ;

\draw    (181,195.67) -- (231,195.67) ;

\draw    (181.67,245) -- (231.67,245) ;

\draw (85.5,282) node   {$\boldsymbol{x}$}; 

\draw (355.17,193.33) node   {$\hat{y}_n$}; 

\draw (168.83,115.33) node   {$(\boldsymbol{w}_{j} ,b_{j})$}; 

\draw (250.5,115) node   {$\beta_{j}$};
\end{tikzpicture} 
\par\end{center}
\par\end{center}%
\end{minipage}
\par\end{centering}
}

\caption{(a): A graphical representation of a two-layers network, as in Eq.
(\ref{eq:two-layers-net}). (b): An equivalent representation for
$\sigma\left(\boldsymbol{x};\boldsymbol{\theta}_{i}\right)=\beta_{i}\varphi\left(\left\langle \boldsymbol{w}_{i},\boldsymbol{x}\right\rangle +b_{i}\right)$.}
\label{fig:Two-layers-net}
\end{figure}
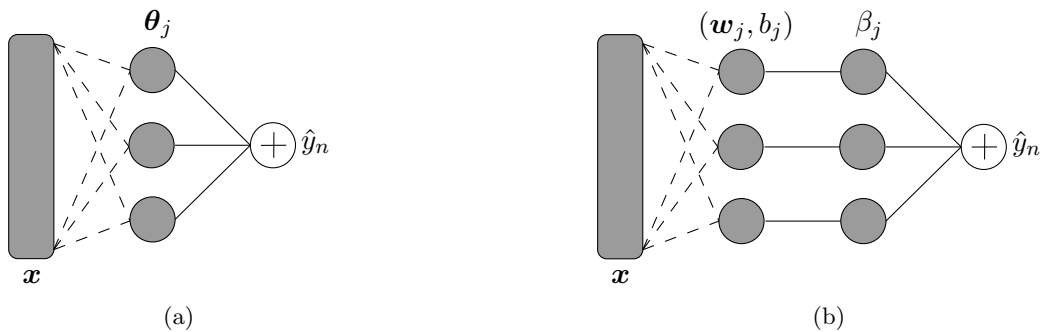

We give a brief and informal overview of relevant results from \cite{mei2018mean}.
Consider the following two-layers neural network:
\begin{equation}
\hat{y}_{n}\left(\boldsymbol{x};{\cal W}\right)=\frac{1}{n}\sum_{i=1}^{n}\sigma\left(\boldsymbol{x};\boldsymbol{\theta}_{i}\right),\label{eq:two-layers-net}
\end{equation}
where $\boldsymbol{x}\in\mathbb{R}^{d}$ is the input, ${\cal W}=\left\{ \boldsymbol{\theta}_{i}\right\} _{i\in\left[n\right]}$
is the collection of weights $\boldsymbol{\theta}_{i}\in\mathbb{R}^{D}$,
and $\sigma:\;\mathbb{R}^{d}\times\mathbb{R}^{D}\mapsto\mathbb{R}$
is the (nonlinear) activation. Here each term $\sigma\left(\boldsymbol{x};\boldsymbol{\theta}_{i}\right)$
is a neuron. With $\boldsymbol{\theta}_{i}=\left(\beta_{i},\boldsymbol{w}_{i},b_{i}\right)\in\mathbb{R}\times\mathbb{R}^{d}\times\mathbb{R}$
and $\sigma\left(\boldsymbol{x};\boldsymbol{\theta}_{i}\right)=\beta_{i}\varphi\left(\left\langle \boldsymbol{w}_{i},\boldsymbol{x}\right\rangle +b_{i}\right)$
for a scalar nonlinearity $\varphi$, this network reduces to the
usual two-layers fully-connected neural network. An illustration is
given in Fig. \ref{fig:Two-layers-net}.

Suppose that at each time $k\in\mathbb{N}$, the data $\left(\boldsymbol{x}^{k},y^{k}\right)\in\mathbb{R}^{d}\times\mathbb{R}$
is drawn independently from a probabilistic source ${\cal P}$. We
train the network with the loss ${\cal L}\left(y,\hat{y}_{n}\left(\boldsymbol{x};{\cal W}\right)\right)$
for a loss function ${\cal L}:\;\mathbb{R}\times\mathbb{R}\mapsto\mathbb{R}$.
In particular, starting from an initialization ${\cal W}^{0}=\left\{ \boldsymbol{\theta}_{i}^{0}\right\} _{i\in\left[n\right]}$,
for a learning rate $\alpha>0$, we perform (discrete-time) SGD:
\begin{equation}
\boldsymbol{\theta}_{i}^{k+1}=\boldsymbol{\theta}_{i}^{k}-\alpha n\nabla_{\boldsymbol{\theta}_{i}}{\cal L}\left(y^{k},\hat{y}_{n}\left(\boldsymbol{x}^{k};{\cal W}^{k}\right)\right)=\boldsymbol{\theta}_{i}^{k}-\alpha\partial_{2}{\cal L}\left(y^{k},\hat{y}_{n}\left(\boldsymbol{x}^{k};{\cal W}^{k}\right)\right)\nabla_{\boldsymbol{\theta}}\sigma\left(\boldsymbol{x}^{k};\boldsymbol{\theta}_{i}^{k}\right).\label{eq:two-layers-SGD}
\end{equation}
We take a note on the scalings by $n$ in Eq. (\ref{eq:two-layers-net})
and (\ref{eq:two-layers-SGD}). The MF limit behavior can then be
observed in the following two senses: statics and dynamics.

\subsubsection*{Statics.}

Observe that the sum in (\ref{eq:two-layers-net}) exhibits symmetry
in the role of the neurons. In the limit $n\to\infty$, one can replace
this sum with an integral:
\begin{equation}
\hat{y}\left(\boldsymbol{x};\rho\right)=\int\sigma\left(\boldsymbol{x};\boldsymbol{\theta}\right)\rho\left({\rm d}\boldsymbol{\theta}\right),\label{eq:two-layers-formal-net}
\end{equation}
for $\rho\in\mathscr{P}\left(\mathbb{R}^{D}\right)$. In particular,
on one hand, for a given ${\cal W}$, the identification $\rho={\rm Emp}\left(\left\{ \boldsymbol{\theta}_{i}\right\} _{i\in\left[n\right]}\right)$
results in $\hat{y}\left(\boldsymbol{x};\rho\right)=\hat{y}_{n}\left(\boldsymbol{x};{\cal W}\right)$.
On the other hand, for a given $\rho$, taking $\boldsymbol{\theta}_{i}\sim\rho$
i.i.d., one gets $\hat{y}_{n}\left(\boldsymbol{x};{\cal W}\right)\approx\hat{y}\left(\boldsymbol{x};\rho\right)$.
An intriguing observation is that, given ${\cal L}$ convex in the
second argument, $\mathbb{E}_{{\cal P}}\left\{ {\cal L}\left(y,\hat{y}\left(\boldsymbol{x};\rho\right)\right)\right\} $
is convex in $\rho$ \cite{bengio2006convex}. As another interesting
fact, \cite{mei2018mean} proves that under certain regularity conditions
and with ${\cal L}$ being the squared loss,
\begin{equation}
\left|\inf_{{\cal W}}\mathbb{E}_{{\cal P}}\left\{ {\cal L}\left(y,\hat{y}_{n}\left(\boldsymbol{x};{\cal W}\right)\right)\right\} -\inf_{\rho}\mathbb{E}_{{\cal P}}\left\{ {\cal L}\left(y,\hat{y}\left(\boldsymbol{x};\rho\right)\right)\right\} \right|=O\left(\frac{1}{n}\right)\xrightarrow{n\to\infty}0.\label{eq:two-layers-statics}
\end{equation}
In short, $\rho$ is a surrogate measure for ${\rm Emp}\left(\left\{ \boldsymbol{\theta}_{i}\right\} _{i\in\left[n\right]}\right)$.

\subsubsection*{Dynamics.}

Using the same idea of replacing the sum with an integral, one can
use the above infinite-$n$ representation to describe the SGD dynamics.
In particular, let $\hat{\rho}_{n}^{k}={\rm Emp}\left(\left\{ \boldsymbol{\theta}_{i}^{k}\right\} _{i\in\left[n\right]}\right)$
the empirical distribution of neuronal weights at time $k$ on the
SGD dynamics. Suppose that given some $\rho^{0}$, at initialization,
$\hat{\rho}_{n}^{0}\Rightarrow\rho^{0}$ as $n\to\infty$. For $\alpha\downarrow0$
as $n\to\infty$, under suitable conditions, \cite{mei2018mean} shows
that almost surely $\hat{\rho}_{n}^{\left\lfloor t/\alpha\right\rfloor }$
converges weakly to a deterministic limit $\rho^{t}$. This limit
is defined via the following differential equation with random initialization:
\begin{equation}
\frac{{\rm d}}{{\rm d}t}\boldsymbol{\theta}^{t}=-\mathbb{E}_{{\cal P}}\left\{ \partial_{2}{\cal L}\left(y,\hat{y}\left(\boldsymbol{x};\rho^{t}\right)\right)\nabla_{\boldsymbol{\theta}}\sigma\left(\boldsymbol{x};\boldsymbol{\theta}^{t}\right)\right\} ,\label{eq:two-layers-dynamics}
\end{equation}
where $\boldsymbol{\theta}^{t}\sim\rho^{t}$ and $\boldsymbol{\theta}^{0}\sim\rho^{0}$.
More specifically, given $\rho^{0}$, we generate $\boldsymbol{\theta}^{0}\sim\rho^{0}$.
Then we let $\boldsymbol{\theta}^{t}$ evolve according to Eq. (\ref{eq:two-layers-dynamics})
from the initialization $\boldsymbol{\theta}^{0}$ with $\rho^{t}={\rm Law}\left(\boldsymbol{\theta}^{t}\right)$
at any time $t$. Note that while \cite{mei2018mean} defines $\rho^{t}$
via a partial differential equation, what we present here is an equivalent
definition that is more convenient for our discussion.

The behavior of the network throughout the SGD dynamics thus tends
to a non-trivial limit, given in an explicit formula, as the number
of neurons tends to infinity. In fact, \cite{mei2018mean} proves
a more quantitative statement that holds so long as $n\gg d$ the
data dimension and $t\leq T$ not too large.

\subsubsection*{Heuristic derivation.}

A heuristic to derive Eq. (\ref{eq:two-layers-dynamics}) from Eq.
(\ref{eq:two-layers-SGD}) is by firstly, identifying $t=k\alpha$
and recognizing that $\alpha\downarrow0$ leads to time continuum
and in-expectation property w.r.t. the data, for $i\in\left[n\right]$:
\[
\frac{{\rm d}}{{\rm d}t}\boldsymbol{\theta}_{i}^{t}\approx-\mathbb{E}_{{\cal P}}\left\{ \partial_{2}{\cal L}\left(y,\hat{y}_{n}\left(\boldsymbol{x};{\cal W}^{t}\right)\right)\nabla_{\boldsymbol{\theta}}\sigma\left(\boldsymbol{x};\boldsymbol{\theta}_{i}^{t}\right)\right\} ,\quad{\cal W}^{t}=\left\{ \boldsymbol{\theta}_{i}^{t}\right\} _{i\in\left[n\right]}.
\]
Secondly, we again replace a sum with an integral, wherever possible.
Here $\hat{y}_{n}\left(\boldsymbol{x};{\cal W}^{t}\right)\approx\hat{y}\left(\boldsymbol{x};\tilde{\rho}^{t}\right)$
from the statics, with $\tilde{\rho}^{t}$ being the surrogate measure
for ${\rm Emp}\left({\cal W}^{t}\right)$ at each time $t\geq0$,
which yields
\begin{equation}
\frac{{\rm d}}{{\rm d}t}\boldsymbol{\theta}_{i}^{t}\approx-\mathbb{E}_{{\cal P}}\left\{ \partial_{2}{\cal L}\left(y,\hat{y}\left(\boldsymbol{x};\tilde{\rho}^{t}\right)\right)\nabla_{\boldsymbol{\theta}}\sigma\left(\boldsymbol{x};\boldsymbol{\theta}_{i}^{t}\right)\right\} .\label{eq:two-layers-dynamics-1}
\end{equation}
Thirdly, we observe symmetry among the neurons in the above. If this
symmetry is attained at $t=0$ by proper initialization, it should
be maintained at all subsequent $t$, and hence one has ${\rm Law}\left(\boldsymbol{\theta}_{i}^{t}\right)\approx\tilde{\rho}^{t}$
in the limit $n\to\infty$, in which case we drop the subscript $i$.
If at initialization $\tilde{\rho}^{0}=\rho^{0}$, then by comparing
the resultant dynamics with Eq. (\ref{eq:two-layers-dynamics}), one
identifies $\tilde{\rho}^{t}\approx\rho^{t}$.

\subsection{Contributions}

In this work, we aim to develop a formalism which describes and derives
the MF limit for multilayer neural networks under suitable scalings.

From the derivation for two-layers networks, we observe that symmetry
among the neurons plays a key role in the MF limit. Intuitively one
may expect the same for multilayer networks in that there is symmetry
among neurons of the same layer \textendash{} see Fig. \ref{fig:Three-layers-net}
of a three-layers network and Fig. \ref{fig:Multilayer-net} of a
generic multilayer one, for visualization. Yet when one attempts to
extend the argument from the two-layers case, several difficulties
and questions arise:
\begin{itemize}
\item In the two-layers case, the network output $\hat{y}_{n}\left(\boldsymbol{x};{\cal W}\right)$
is a sum of signals from individual neurons which do not share weights.
However in a multilayer network, neurons at layer $\ell$ receive
signals that are constrained to come from the same set of neurons
of layer $\ell-1$ or $\ell+1$.
\item In the two-layers case, each neuron is represented by its respective
weight $\boldsymbol{\theta}_{i}^{k}$ at each time $k$. This representation
is natural: $\hat{y}_{n}\left(\boldsymbol{x};{\cal W}\right)$ assumes
the simple form of a sum, and once the approximation $\hat{y}_{n}\left(\boldsymbol{x};{\cal W}\right)\approx\hat{y}\left(\boldsymbol{x};\rho\right)$
is made, each neuron is updated separately under the SGD dynamics,
in light of Eq. (\ref{eq:two-layers-dynamics-1}). However, as said
above, the multilayer case presents a certain structural constraint.
Moreover due to the layering structure, the update of each neuron
is influenced by other neurons of the adjacent layers. In what way
can we give a quantitative representation for each neuron that respects
the complexity in the structure, and at the same time, exploits the
neuronal symmetry to make simplifications?
\item Observe that the scalings are chosen so that quantities of interests
remain $O\left(1\right)$, roughly speaking. In the particular case
of two-layers networks, the scaling $1/n$ in $\hat{y}_{n}\left(\boldsymbol{x};{\cal W}\right)$
and the factor $n$ in the gradient update in Eq. (\ref{eq:two-layers-SGD})
ensure that $\hat{y}_{n}\left(\boldsymbol{x};{\cal W}\right)$ and
the differential change $\left(\boldsymbol{\theta}_{i}^{k+1}-\boldsymbol{\theta}_{i}^{k}\right)/\alpha$
are $O\left(1\right)$. Under what scalings does the MF limit behavior
occur for the multilayer case?
\end{itemize}
To make a first step, we postulate that neuronal symmetry gives rise
to two crucial properties, which we call \textit{marginal uniformity}
and \textit{self-averaging}. While they are used already in the two-layers
case, the complexity in a multilayer network requires more extensive
and explicit use of these properties, especially self-averaging. This
enables a heuristic derivation of the MF limit, revealing the answers
to the aforementioned questions:
\begin{itemize}
\item We propose that each certain neuron is represented not directly by
its corresponding weights, but by a stochastic kernel which outputs
at random the corresponding weights, conditional on the neurons of
the previous layer. We show how this representation, which changes
from layer to layer, can adapt to the constraint of multilayer hierarchies
and formalize the properties arising from neuronal symmetry.
\item Despite the somewhat complex representation, interestingly to describe
the MF limit of multilayer fully-connected networks, one requires
only one simple statistic of the stochastic kernel: its conditional
expectation. A key insight is the following: since a neuron at layer
$\ell$ receives an aggregate of signals from a set of neurons of
layer $\ell+1$ or $\ell-1$, this aggregate simplifies itself when
one views these neurons as one whole ensemble, thanks to the self-averaging
property.
\item We also find that the appropriate scalings are not uniform across
different layers.
\end{itemize}
We must caution the readers that our current development is non-rigorous.
As such, an outstanding challenge remains: under what regularity conditions,
as well as what precise mathematical sense, can the formalism hold?
On the other hand, the formalism is meant to be informative: firstly,
it is predictive of the MF limit behavior to be observed in real simulations
when the number of neurons is sufficiently large; secondly, it explores
a regime, associated with specific scalings, that is under-studied
by current experimental and theoretical pursuits; thirdly, it signifies
the potential for a theoretical framework to analyze and design multilayer
neural networks \textendash{} a quest that has recently witnessed
progresses in the two-layers case.

\subsection{Outline}

In Section \ref{sec:MF-three-layers}, we present the formalism in
the particular case of a three-layers network. In particular, Section
\ref{subsec:three-layers-setting} and Section \ref{subsec:three-layers-formalism}
run in parallel, each describing the forward pass, the backward pass
and the (learning or evolution) dynamics. The former section is on
the neural network with scalings, and the latter is on its MF limit.
We give a heuristic derivation of their connection in Section \ref{subsec:three-layers-derivation}
and several remarks in Section \ref{subsec:Discussions-three-layers}.
The case of general multilayer networks is presented in Section \ref{sec:MF-Multilayer},
with its heuristic derivation deferred to Appendix \ref{sec:multilayer-derivation}.
Since the treatments of these two cases are similar in spirit, the
readers are urged to read Section \ref{sec:MF-three-layers}, where
the key ideas are explained in greater details. In Section \ref{sec:Validation},
we present several experiments and a theoretical result to validate
the existence of the MF limit. We particularly do not aim for achieving
competitive empirical results in our experiments. It remains open
to find good practices to train a network in this regime, a task that
deserves another investigation.

While our main focus is fully-connected multilayer networks, the generality
of the principles allows us to draw similar conclusions on certain
other settings. See Appendix \ref{sec:LLN-CNNs} where we discuss
the case of multilayer convolutional neural networks.

In the following, we discuss related works.

\subsection{Related works}

As mentioned, several recent works have studied the MF limit in the
two-layers network case. The works \cite{mei2018mean,chizat2018,sirignano2018mean,rotskoff2018neural}
establish the MF limit, and in particular, \cite{mei2018mean} proves
that this holds as soon as the number of neurons exceeds the data
dimension. \cite{mei2018mean,chizat2018} utilize this limit to prove
that (noisy) SGD can converge to (near) global optimum under different
assumptions. For a specific class of activations and data distribution,
\cite{javanmard2019analysis} proves that this convergence is exponentially
fast using the displacement convexity property of the MF limit. Taking
the same viewpoint, \cite{wei2018margin} proves a convergence result
for a specifically chosen many-neurons limit. \cite{rotskoff2018neural,sirignano2018CLT}
study the fluctuations around the MF limit. Our analysis of the multilayer
case requires substantial extension and new ideas, uncovering certain
properties that are not obvious from the two-layers analysis (see
also Section \ref{subsec:Discussions-three-layers}).

We take a note on the work \cite{hazan2015steps}, which shares a
few similarities with our work in the forward pass description (for
instance, in Eq. (1) of \cite{hazan2015steps} as compared to Eq.
(\ref{eq:three-layers-formal-net}) in our work). \cite{hazan2015steps}
differs in that it takes a kernel method perspective and develops
a Gaussian process formulation, which makes strong assumptions on
the distribution of the weights. Its formulation does not extend beyond
three layers. Meanwhile our work focuses on the MF limit, points out
explicitly the appropriate scalings, proposes new crucial ideas to
address the backward pass and the learning dynamics, and is not limited
to any specific number of layers.

There is a vast literature on settings that assume a large number
of neurons \textendash{} typically specific to the over-parameterized
regime. We shall mention here a recent subset. The highly non-convex
nature of the optimization landscape enjoys attention from a major
body of works \cite{safran2016quality,soudry2016no,freeman2016topology,nguyen2017loss,mei2018landscape,nguyen2018optimization,venturi2018spurious,cooper2018loss,du2018power,soltanolkotabi2018theoretical,yun2018small}.
This is yet far from a complete picture without a study of the trajectory
of the learning dynamics, which has witnessed recent progresses. Several
works \cite{li2018learning,du2018gradient,du2018gradientDeep,allen2018learning,allen2018convergence,zou2018stochastic}
concurrently show that gradient-based learning dynamics can find the
global optimum in multilayer networks, provided an extremely large
number of neurons. The work \cite{jacot2018neural} develops a complementary
viewpoint on the dynamics, the so-called neural tangent kernel, also
in the limit of infinitely many neurons. A common feature of these
works is that throughout the considered training period, certain properties
of the network remain close to the randomized initialization, and
the network behaves like kernel regression. Further discussions in
this regard can be found in the recent note \cite{chizat2018note}.
Complementing these mathematical approaches, \cite{geiger2018jamming,spigler2018jamming,geiger2019scaling}
utilize the physics of jamming to make a quantitative prediction of
the boundary between the over-parameterized and under-parameterized
regions, as well as the generalization behavior of over-parameterized
networks, under a specific choice of the loss function. In another
development, several works \cite{poole2016exponential,schoenholz2016deep,pennington2017resurrecting,yang2017mean,chen2018dynamical,xiao2018dynamical,hanin2018start,hanin2018neural,li2018on,yang2018a}
obtain good initialization strategies by studying networks with infinitely
many neurons and random weights (which hence disregard the learning
dynamics). These works form a basis for a Gaussian process perspective
\cite{lee2017deep,matthews2018gaussian,garriga2018deep,novak2018bayesian}.
All these directions are not directly comparable with ours. Furthermore
we note that the settings in these works assume different scalings
from ours and thus do not exhibit the same MF limit behavior that
is to be presented here.

\section{Mean field limit in three-layers fully-connected networks\label{sec:MF-three-layers}}

In this section, we develop a formalism in which the MF limit is derived
for a three-layers neural network under suitable scalings. The focus
on this specific case is made for simplicity of the presentation and
illustration of the key ideas. While certain elements have already
been seen in the two-layers case, there are important and substantial
differences that shall be highlighted.

\subsection{Setting: A three-layers network\label{subsec:three-layers-setting}}

\subsubsection*{Forward pass.}

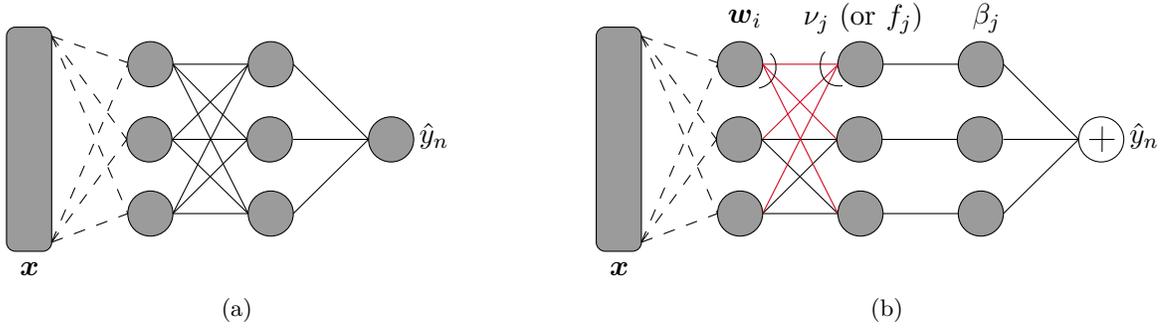
\begin{figure}
\subfloat[]{\begin{centering}
\begin{minipage}[t]{0.5\columnwidth}%
\begin{center}
\begin{center}
\begin{tikzpicture}[yscale=-0.02,xscale=0.02] 
\draw  [fill={rgb, 255:red, 155; green, 155; blue, 155 }  ,fill opacity=1 ] (50.08,107) .. controls (50.08,103.69) and (52.77,101) .. (56.08,101) -- (74.08,101) .. controls (77.4,101) and (80.08,103.69) .. (80.08,107) -- (80.08,244.08) .. controls (80.08,247.4) and (77.4,250.08) .. (74.08,250.08) -- (56.08,250.08) .. controls (52.77,250.08) and (50.08,247.4) .. (50.08,244.08) -- cycle ; 

\draw  [fill={rgb, 255:red, 155; green, 155; blue, 155 }  ,fill opacity=1 ] (130.67,125.67) .. controls (130.67,117.38) and (137.38,110.67) .. (145.67,110.67) .. controls (153.95,110.67) and (160.67,117.38) .. (160.67,125.67) .. controls (160.67,133.95) and (153.95,140.67) .. (145.67,140.67) .. controls (137.38,140.67) and (130.67,133.95) .. (130.67,125.67) -- cycle ; 

\draw  [fill={rgb, 255:red, 155; green, 155; blue, 155 }  ,fill opacity=1 ] (130,175.67) .. controls (130,167.38) and (136.72,160.67) .. (145,160.67) .. controls (153.28,160.67) and (160,167.38) .. (160,175.67) .. controls (160,183.95) and (153.28,190.67) .. (145,190.67) .. controls (136.72,190.67) and (130,183.95) .. (130,175.67) -- cycle ; 

\draw  [fill={rgb, 255:red, 155; green, 155; blue, 155 }  ,fill opacity=1 ] (130.67,225) .. controls (130.67,216.72) and (137.38,210) .. (145.67,210) .. controls (153.95,210) and (160.67,216.72) .. (160.67,225) .. controls (160.67,233.28) and (153.95,240) .. (145.67,240) .. controls (137.38,240) and (130.67,233.28) .. (130.67,225) -- cycle ; 

\draw  [fill={rgb, 255:red, 155; green, 155; blue, 155 }  ,fill opacity=1 ] (210.67,125.67) .. controls (210.67,117.38) and (217.38,110.67) .. (225.67,110.67) .. controls (233.95,110.67) and (240.67,117.38) .. (240.67,125.67) .. controls (240.67,133.95) and (233.95,140.67) .. (225.67,140.67) .. controls (217.38,140.67) and (210.67,133.95) .. (210.67,125.67) -- cycle ; 

\draw  [fill={rgb, 255:red, 155; green, 155; blue, 155 }  ,fill opacity=1 ] (210,175.67) .. controls (210,167.38) and (216.72,160.67) .. (225,160.67) .. controls (233.28,160.67) and (240,167.38) .. (240,175.67) .. controls (240,183.95) and (233.28,190.67) .. (225,190.67) .. controls (216.72,190.67) and (210,183.95) .. (210,175.67) -- cycle ; 

\draw  [fill={rgb, 255:red, 155; green, 155; blue, 155 }  ,fill opacity=1 ] (210.67,225) .. controls (210.67,216.72) and (217.38,210) .. (225.67,210) .. controls (233.95,210) and (240.67,216.72) .. (240.67,225) .. controls (240.67,233.28) and (233.95,240) .. (225.67,240) .. controls (217.38,240) and (210.67,233.28) .. (210.67,225) -- cycle ;

\draw  [fill={rgb, 255:red, 155; green, 155; blue, 155 }  ,fill opacity=1 ] (290.67,175.67) .. controls (290.67,167.38) and (297.38,160.67) .. (305.67,160.67) .. controls (313.95,160.67) and (320.67,167.38) .. (320.67,175.67) .. controls (320.67,183.95) and (313.95,190.67) .. (305.67,190.67) .. controls (297.38,190.67) and (290.67,183.95) .. (290.67,175.67) -- cycle ; 

\draw    (240.67,175.67) -- (290.67,175.67) ;

\draw    (240.67,125.67) -- (290.67,175.67) ;

\draw    (290.67,175.67) -- (240.67,225) ;

\draw (160.67,125.67) -- (210.67,125.67) ;

\draw    (160,175.67) -- (210,175.67) ;

\draw    (160.67,225) -- (210.67,225) ;

\draw (160.67,125.67) -- (210,175.67) ;

\draw    (160,175.67) -- (210.67,225) ;

\draw    (210,176) -- (160.67,225) ;

\draw (210.67,126) -- (160.67,175.33) ;

\draw  [dash pattern={on 4.5pt off 4.5pt}]  (80.08,107) -- (130.67,124.67) ;

\draw  [dash pattern={on 4.5pt off 4.5pt}]  (80.08,244.08) -- (130.67,125.67) ;

\draw  [dash pattern={on 4.5pt off 4.5pt}]  (80.08,107) -- (130,175.67) ;

\draw  [dash pattern={on 4.5pt off 4.5pt}]  (80.08,244.08) -- (130,175.67) ;

\draw  [dash pattern={on 4.5pt off 4.5pt}]  (80.08,244.08) -- (130.67,225) ;

\draw  [dash pattern={on 4.5pt off 4.5pt}]  (80.08,107) -- (130.67,225) ;

\draw (160.67,225) -- (210.67,126) ;

\draw (160.67,125.67) -- (210.67,225) ;

\draw (65.5,262) node   {$\boldsymbol{x}$}; 

\draw (334.17,173.33) node   {$\hat{y}_n$}; 

\end{tikzpicture}
\par\end{center}
\par\end{center}%
\end{minipage}
\par\end{centering}
}\subfloat[]{\begin{centering}
\begin{minipage}[t]{0.5\columnwidth}%
\begin{center}
\begin{center}
\begin{tikzpicture}[xscale=0.02, yscale=-0.02] 

\draw  [fill={rgb, 255:red, 155; green, 155; blue, 155 }  ,fill opacity=1 ] (50.08,107) .. controls (50.08,103.69) and (52.77,101) .. (56.08,101) -- (74.08,101) .. controls (77.4,101) and (80.08,103.69) .. (80.08,107) -- (80.08,244.08) .. controls (80.08,247.4) and (77.4,250.08) .. (74.08,250.08) -- (56.08,250.08) .. controls (52.77,250.08) and (50.08,247.4) .. (50.08,244.08) -- cycle ; 

\draw  [fill={rgb, 255:red, 155; green, 155; blue, 155 }  ,fill opacity=1 ] (130.67,125.67) .. controls (130.67,117.38) and (137.38,110.67) .. (145.67,110.67) .. controls (153.95,110.67) and (160.67,117.38) .. (160.67,125.67) .. controls (160.67,133.95) and (153.95,140.67) .. (145.67,140.67) .. controls (137.38,140.67) and (130.67,133.95) .. (130.67,125.67) -- cycle ; 

\draw  [fill={rgb, 255:red, 155; green, 155; blue, 155 }  ,fill opacity=1 ] (130,175.67) .. controls (130,167.38) and (136.72,160.67) .. (145,160.67) .. controls (153.28,160.67) and (160,167.38) .. (160,175.67) .. controls (160,183.95) and (153.28,190.67) .. (145,190.67) .. controls (136.72,190.67) and (130,183.95) .. (130,175.67) -- cycle ; 

\draw  [fill={rgb, 255:red, 155; green, 155; blue, 155 }  ,fill opacity=1 ] (130.67,225) .. controls (130.67,216.72) and (137.38,210) .. (145.67,210) .. controls (153.95,210) and (160.67,216.72) .. (160.67,225) .. controls (160.67,233.28) and (153.95,240) .. (145.67,240) .. controls (137.38,240) and (130.67,233.28) .. (130.67,225) -- cycle ; 

\draw  [fill={rgb, 255:red, 155; green, 155; blue, 155 }  ,fill opacity=1 ] (210.67,125.67) .. controls (210.67,117.38) and (217.38,110.67) .. (225.67,110.67) .. controls (233.95,110.67) and (240.67,117.38) .. (240.67,125.67) .. controls (240.67,133.95) and (233.95,140.67) .. (225.67,140.67) .. controls (217.38,140.67) and (210.67,133.95) .. (210.67,125.67) -- cycle ; 

\draw  [fill={rgb, 255:red, 155; green, 155; blue, 155 }  ,fill opacity=1 ] (210,175.67) .. controls (210,167.38) and (216.72,160.67) .. (225,160.67) .. controls (233.28,160.67) and (240,167.38) .. (240,175.67) .. controls (240,183.95) and (233.28,190.67) .. (225,190.67) .. controls (216.72,190.67) and (210,183.95) .. (210,175.67) -- cycle ; 

\draw  [fill={rgb, 255:red, 155; green, 155; blue, 155 }  ,fill opacity=1 ] (210.67,225) .. controls (210.67,216.72) and (217.38,210) .. (225.67,210) .. controls (233.95,210) and (240.67,216.72) .. (240.67,225) .. controls (240.67,233.28) and (233.95,240) .. (225.67,240) .. controls (217.38,240) and (210.67,233.28) .. (210.67,225) -- cycle ;

\draw  [fill={rgb, 255:red, 255; green, 255; blue, 255 }  ,fill opacity=1 ] (370.67,175.67) .. controls (370.67,167.38) and (377.38,160.67) .. (385.67,160.67) .. controls (393.95,160.67) and (400.67,167.38) .. (400.67,175.67) .. controls (400.67,183.95) and (393.95,190.67) .. (385.67,190.67) .. controls (377.38,190.67) and (370.67,183.95) .. (370.67,175.67) -- cycle ; 

\draw    (320.67,175.67) -- (370.67,175.67) ;

\draw    (320.67,125.67) -- (370.67,175.67) ;

\draw    (370.67,175.67) -- (320.67,225) ;

\draw [color={rgb, 255:red, 208; green, 2; blue, 27 }  ,draw opacity=1 ]   (160.67,125.67) -- (210.67,125.67) ;

\draw    (160,175.67) -- (210,175.67) ;

\draw    (160.67,225) -- (210.67,225) ;

\draw [color={rgb, 255:red, 208; green, 2; blue, 27 }  ,draw opacity=1 ]   (160.67,125.67) -- (210,175.67) ;

\draw    (160,175.67) -- (210.67,225) ;

\draw    (210,176) -- (160.67,225) ;

\draw [color={rgb, 255:red, 208; green, 2; blue, 27 }  ,draw opacity=1 ]   (210.67,126) -- (160.67,175.33) ;

\draw  [dash pattern={on 4.5pt off 4.5pt}]  (80.08,107) -- (130.67,124.67) ;

\draw  [dash pattern={on 4.5pt off 4.5pt}]  (80.08,244.08) -- (130.67,125.67) ;

\draw  [dash pattern={on 4.5pt off 4.5pt}]  (80.08,107) -- (130,175.67) ;

\draw  [dash pattern={on 4.5pt off 4.5pt}]  (80.08,244.08) -- (130,175.67) ;

\draw  [dash pattern={on 4.5pt off 4.5pt}]  (80.08,244.08) -- (130.67,225) ;

\draw  [dash pattern={on 4.5pt off 4.5pt}]  (80.08,107) -- (130.67,225) ;

\draw [color={rgb, 255:red, 208; green, 2; blue, 27 }  ,draw opacity=1 ]   (160.67,225) -- (210.67,126) ;

\draw [color={rgb, 255:red, 208; green, 2; blue, 27 }  ,draw opacity=1 ]   (160.67,125.67) -- (210.67,225) ;

\draw [color={rgb, 255:red, 0; green, 0; blue, 0 }  ,draw opacity=1 ]   (158.25,141.5) .. controls (170.75,139) and (172.25,124) .. (165.75,118.5) ;

\draw [color={rgb, 255:red, 0; green, 0; blue, 0 }  ,draw opacity=1 ]   (213.75,140) .. controls (195.75,142) and (197.25,120) .. (202.25,117.5) ;

\draw  [fill={rgb, 255:red, 155; green, 155; blue, 155 }  ,fill opacity=1 ] (290.67,125.67) .. controls (290.67,117.38) and (297.38,110.67) .. (305.67,110.67) .. controls (313.95,110.67) and (320.67,117.38) .. (320.67,125.67) .. controls (320.67,133.95) and (313.95,140.67) .. (305.67,140.67) .. controls (297.38,140.67) and (290.67,133.95) .. (290.67,125.67) -- cycle ; 

\draw  [fill={rgb, 255:red, 155; green, 155; blue, 155 }  ,fill opacity=1 ] (290,175.67) .. controls (290,167.38) and (296.72,160.67) .. (305,160.67) .. controls (313.28,160.67) and (320,167.38) .. (320,175.67) .. controls (320,183.95) and (313.28,190.67) .. (305,190.67) .. controls (296.72,190.67) and (290,183.95) .. (290,175.67) -- cycle ; 

\draw  [fill={rgb, 255:red, 155; green, 155; blue, 155 }  ,fill opacity=1 ] (290.67,225) .. controls (290.67,216.72) and (297.38,210) .. (305.67,210) .. controls (313.95,210) and (320.67,216.72) .. (320.67,225) .. controls (320.67,233.28) and (313.95,240) .. (305.67,240) .. controls (297.38,240) and (290.67,233.28) .. (290.67,225) -- cycle ;

\draw    (377.67,175.75) -- (394.5,175.67) ;

\draw    (386.21,184) -- (386.33,167.03) ;

\draw    (240.67,125.67) -- (290.67,125.67) ;

\draw    (240,175.67) -- (290,175.67) ;

\draw    (240.67,225) -- (290.67,225) ;

\draw (65.5,262) node   {$\boldsymbol{x}$}; 

\draw (414.17,173.33) node   {$\hat{y}_n$}; 

\draw (148.83,95.33) node   {$\boldsymbol{w}_{i}$}; 

\draw (227.5,95) node   {$\nu_j$ (or $f_{j}$)}; 

\draw (309.5,95) node   {$\beta_{j}$};
\end{tikzpicture} 
\par\end{center}
\par\end{center}%
\end{minipage}
\par\end{centering}
}

\caption{(a): A graphical representation of a three-layers network. (b): An
equivalent representation, as proposed in Section \ref{subsec:three-layers-derivation}.
Neuron $j$ of the second layer is represented by $\nu_{j}$, and
$f_{j}={\rm CE}\left\{ \nu_{j}\right\} $. Notice that neuron $j$
of the second layer receives the forward pass information averaged
over all neurons of the first layer. Likewise, neuron $i$ of the
first layer receives the backward pass information averaged over all
neurons of the second layer. However neuron $j$ of the third layer
does not average its received forward pass information over all neurons
of the second layer, due to its connectivity. Likewise, neuron $j$
of the second layer does not average its received backward pass information
over all neurons of the third layer.}
\label{fig:Three-layers-net}
\end{figure}

We consider the following three-layers neural network with fully-connected
layers and no biases:
\begin{equation}
\hat{y}_{n}\left(\boldsymbol{x};{\cal W}\right)=\frac{1}{n_{2}}\left\langle \boldsymbol{\beta},\sigma\left(\boldsymbol{h}_{2}\right)\right\rangle ,\quad\boldsymbol{h}_{2}=\frac{1}{n_{1}}\boldsymbol{W}_{2}\sigma\left(\boldsymbol{h}_{1}\right),\quad\boldsymbol{h}_{1}=\boldsymbol{W}_{1}\boldsymbol{x},\label{eq:three-layers-net}
\end{equation}
in which $\boldsymbol{x}\in\mathbb{R}^{d}$ is the input to the network,
$\hat{y}_{n}\left(\boldsymbol{x};{\cal W}\right)\in\mathbb{R}$ is
the output, ${\cal W}=\left\{ \boldsymbol{W}_{1},\boldsymbol{W}_{2},\boldsymbol{\beta}\right\} $
is the collection of weights, $\boldsymbol{W}_{1}\in\mathbb{R}^{n_{1}\times d}$,
$\boldsymbol{W}_{2}\in\mathbb{R}^{n_{2}\times n_{1}}$, $\boldsymbol{\beta}\in\mathbb{R}^{n_{2}}$,
and $\sigma:\;\mathbb{R}\mapsto\mathbb{R}$ is a nonlinear activation.
$\boldsymbol{h}_{1}$ and $\boldsymbol{h}_{2}$ are commonly called
the pre-activations. Here $n_{1}=n_{1}\left(n\right)$ and $n_{2}=n_{2}\left(n\right)$,
both of which shall be taken to $\infty$ as $n\to\infty$. An illustration
is given in Fig. \ref{fig:Three-layers-net}.(a).

\subsubsection*{Backward pass.}

The backward pass computes several derivative quantities to be used
for learning. Let us define:
\begin{align}
\tilde{\nabla}_{\boldsymbol{\beta}}\hat{y}_{n}\left(\boldsymbol{x};{\cal W}\right) & =n_{2}\nabla_{\boldsymbol{\beta}}\hat{y}_{n}\left(\boldsymbol{x};{\cal W}\right)=\sigma\left(\boldsymbol{h}_{2}\right),\label{eq:three-layers-gradtheta}\\
\tilde{\nabla}_{\boldsymbol{h}_{2}}\hat{y}_{n}\left(\boldsymbol{x};{\cal W}\right) & =n_{2}\nabla_{\boldsymbol{h}_{2}}\hat{y}_{n}\left(\boldsymbol{x};{\cal W}\right)=\boldsymbol{\beta}\odot\sigma'\left(\boldsymbol{h}_{2}\right),\label{eq:three-layers-gradh2}\\
\tilde{\nabla}_{\boldsymbol{W}_{2}}\hat{y}_{n}\left(\boldsymbol{x};{\cal W}\right) & =n_{1}n_{2}\nabla_{\boldsymbol{W}_{2}}\hat{y}_{n}\left(\boldsymbol{x};{\cal W}\right)=\tilde{\nabla}_{\boldsymbol{h}_{2}}\hat{y}_{n}\left(\boldsymbol{x};{\cal W}\right)\sigma\left(\boldsymbol{h}_{1}\right)^{\top},\label{eq:three-layers-gradW2}\\
\tilde{\nabla}_{\boldsymbol{h}_{1}}\hat{y}_{n}\left(\boldsymbol{x};{\cal W}\right) & =n_{1}\nabla_{\boldsymbol{h}_{1}}\hat{y}_{n}\left(\boldsymbol{x};{\cal W}\right)=\frac{1}{n_{2}}\left(\boldsymbol{W}_{2}^{\top}\tilde{\nabla}_{\boldsymbol{h}_{2}}\hat{y}_{n}\left(\boldsymbol{x};{\cal W}\right)\right)\odot\sigma'\left(\boldsymbol{h}_{1}\right),\label{eq:three-layers-gradh1}\\
\tilde{\nabla}_{\boldsymbol{W}_{1}}\hat{y}_{n}\left(\boldsymbol{x};{\cal W}\right) & =n_{1}\nabla_{\boldsymbol{W}_{1}}\hat{y}_{n}\left(\boldsymbol{x};{\cal W}\right)=\tilde{\nabla}_{\boldsymbol{h}_{1}}\hat{y}_{n}\left(\boldsymbol{x};{\cal W}\right)\boldsymbol{x}^{\top}.\label{eq:three-layers-gradW1}
\end{align}

\subsubsection*{Learning dynamics.}

Similar to Section \ref{subsec:two-layers}, we assume that at each
time $k\in\mathbb{N}$, the data $\left(\boldsymbol{x}^{k},y^{k}\right)\in\mathbb{R}^{d}\times\mathbb{R}$
is drawn independently from a probabilistic source ${\cal P}$. We
train the network with the loss ${\cal L}\left(y,\hat{y}_{n}\left(\boldsymbol{x};{\cal W}\right)\right)$
for a loss function ${\cal L}:\;\mathbb{R}\times\mathbb{R}\mapsto\mathbb{R}$,
using SGD with an initialization ${\cal W}^{0}=\left\{ \boldsymbol{W}_{1}^{0},\boldsymbol{W}_{2}^{0},\boldsymbol{\beta}^{0}\right\} $
and a learning rate $\alpha>0$:
\begin{align}
\boldsymbol{\beta}^{k+1} & =\boldsymbol{\beta}^{k}-\alpha\partial_{2}{\cal L}\left(y^{k},\hat{y}_{n}\left(\boldsymbol{x}^{k};{\cal W}^{k}\right)\right)\tilde{\nabla}_{\boldsymbol{\beta}}\hat{y}_{n}\left(\boldsymbol{x}^{k};{\cal W}^{k}\right),\label{eq:three-layers-SGDtheta}\\
\boldsymbol{W}_{2}^{k+1} & =\boldsymbol{W}_{2}^{k}-\alpha\partial_{2}{\cal L}\left(y^{k},\hat{y}_{n}\left(\boldsymbol{x}^{k};{\cal W}^{k}\right)\right)\tilde{\nabla}_{\boldsymbol{W}_{2}}\hat{y}_{n}\left(\boldsymbol{x}^{k};{\cal W}^{k}\right),\label{eq:three-layers-SGDW2}\\
\boldsymbol{W}_{1}^{k+1} & =\boldsymbol{W}_{1}^{k}-\alpha\partial_{2}{\cal L}\left(y^{k},\hat{y}_{n}\left(\boldsymbol{x}^{k};{\cal W}^{k}\right)\right)\tilde{\nabla}_{\boldsymbol{W}_{1}}\hat{y}_{n}\left(\boldsymbol{x}^{k};{\cal W}^{k}\right).\label{eq:three-layers-SGDW1}
\end{align}
This yields the learning dynamics of ${\cal W}^{k}=\left\{ \boldsymbol{W}_{1}^{k},\boldsymbol{W}_{2}^{k},\boldsymbol{\beta}^{k}\right\} $.
Notice the scaling by $n_{1}$, $n_{2}$ and $n_{1}n_{2}$ at various
places in Eq. (\ref{eq:three-layers-net}) and Eq. (\ref{eq:three-layers-gradtheta})-(\ref{eq:three-layers-gradW1}).

\subsection{Mean field limit\label{subsec:three-layers-formalism}}

In the following, we describe a time-evolving system which resembles
the three-layers network but does not involve the numbers of neurons
$n_{1}$ and $n_{2}$. We then state a prediction that connects this
formal system with the three-layers network. This, in particular,
specifies the MF limit of the three-layers network.

\subsubsection*{Forward pass.}

Let us define
\begin{equation}
\hat{y}\left(\boldsymbol{x};\rho_{1},\rho_{2}\right)=\int\beta\sigma\left(H_{2}\left(f;\boldsymbol{x},\rho_{1}\right)\right)\rho_{2}\left({\rm d}f,{\rm d}\beta\right),\label{eq:three-layers-formal-net}
\end{equation}
where $\rho_{1}\in\mathscr{P}\left(\mathbb{R}^{d}\right)$, $\rho_{2}\in\mathscr{P}\left({\cal F}\times\mathbb{R}\right)$
for ${\cal F}=\left\{ f:\;\mathbb{R}^{d}\mapsto\mathbb{R}\right\} $,
and
\[
H_{1}\left(\boldsymbol{w};\boldsymbol{x}\right)=\left\langle \boldsymbol{w},\boldsymbol{x}\right\rangle ,\qquad H_{2}\left(f;\boldsymbol{x},\rho_{1}\right)=\int f\left(\boldsymbol{w}\right)\sigma\left(H_{1}\left(\boldsymbol{w};\boldsymbol{x}\right)\right)\rho_{1}\left({\rm d}\boldsymbol{w}\right).
\]
This describes a system defined via $\rho_{1}$ and $\rho_{2}$. More
specifically, $\rho_{1}$ and $\rho_{2}$ are the state of the system,
and the system takes $\boldsymbol{x}\in\mathbb{R}^{d}$ as input and
outputs $\hat{y}\left(\boldsymbol{x};\rho_{1},\rho_{2}\right)\in\mathbb{R}$.
One should compare $\hat{y}\left(\boldsymbol{x};\rho_{1},\rho_{2}\right)$,
$H_{1}\left(\boldsymbol{w};\boldsymbol{x}\right)$ and $H_{2}\left(f;\boldsymbol{x},\rho_{1}\right)$
with respectively $\hat{y}_{n}\left(\boldsymbol{x};{\cal W}\right)$,
$\boldsymbol{h}_{1}$ and $\boldsymbol{h}_{2}$ of the three-layers
network.

\subsubsection*{Backward pass.}

Let us define the following quantities:
\begin{align}
\Delta_{\beta}\left(f;\boldsymbol{x},\rho_{1}\right) & =\sigma\left(H_{2}\left(f;\boldsymbol{x},\rho_{1}\right)\right),\label{eq:three-layers-formal-gradtheta}\\
\Delta_{H_{2}}\left(\beta,f;\boldsymbol{x},\rho_{1}\right) & =\beta\sigma'\left(H_{2}\left(f;\boldsymbol{x},\rho_{1}\right)\right),\label{eq:three-layers-formal-gradh2}\\
\Delta_{w_{2}}\left(\beta,f,\boldsymbol{w};\boldsymbol{x},\rho_{1}\right) & =\Delta_{H_{2}}\left(\beta,f;\boldsymbol{x},\rho_{1}\right)\sigma\left(H_{1}\left(\boldsymbol{w};\boldsymbol{x}\right)\right),\label{eq:three-layers-formal-gradW2}\\
\Delta_{H_{1}}\left(\boldsymbol{w};\boldsymbol{x},\rho_{1},\rho_{2}\right) & =\sigma'\left(H_{1}\left(\boldsymbol{w};\boldsymbol{x}\right)\right)\int f\left(\boldsymbol{w}\right)\Delta_{H_{2}}\left(\beta,f;\boldsymbol{x},\rho_{1}\right)\rho_{2}\left({\rm d}f,{\rm d}\beta\right),\label{eq:three-layers-formal-gradh1}\\
\Delta_{\boldsymbol{w}_{1}}\left(\boldsymbol{w};\boldsymbol{x},\rho_{1},\rho_{2}\right) & =\Delta_{H_{1}}\left(\boldsymbol{w};\boldsymbol{x},\rho_{1},\rho_{2}\right)\boldsymbol{x}.\label{eq:three-layers-formal-gradW1}
\end{align}
One should compare Eq. (\ref{eq:three-layers-formal-gradtheta})-(\ref{eq:three-layers-formal-gradW1})
with Eq. (\ref{eq:three-layers-gradtheta})-(\ref{eq:three-layers-gradW1})
respectively.

\subsubsection*{Evolution dynamics.}

Now we describe a continuous-time evolution dynamics of the system,
defined at each time $t$ via $\rho_{1}^{t}\in\mathscr{P}\left(\mathbb{R}^{d}\right)$
and $\rho_{2}^{t}\in\mathscr{P}\left({\cal F}\times\mathbb{R}\right)$.
Specifically, given $\rho_{1}^{0}\in\mathscr{P}\left(\mathbb{R}^{d}\right)$
and $\rho_{2}^{0}\in\mathscr{P}\left({\cal F}\times\mathbb{R}\right)$,
we generate $\boldsymbol{w}^{0}\sim\rho_{1}^{0}$ and $\left(f^{0},\beta^{0}\right)\sim\rho_{2}^{0}$.
Taking them as the initialization, we then let $\boldsymbol{w}^{t}$,
$f^{t}$ and $\beta^{t}$ evolve according to
\begin{align}
\frac{{\rm d}}{{\rm d}t}\boldsymbol{w}^{t} & =G_{\boldsymbol{w}}\left(\boldsymbol{w}^{t};\rho_{1}^{t},\rho_{2}^{t}\right),\label{eq:three-layers-formal-dynamicsW}\\
\partial_{t}f^{t}\left(\boldsymbol{w}\right)+\left\langle \nabla f^{t}\left(\boldsymbol{w}\right),G_{\boldsymbol{w}}\left(\boldsymbol{w};\rho_{1}^{t},\rho_{2}^{t}\right)\right\rangle  & =G_{f}\left(\beta^{t},f^{t},\boldsymbol{w};\rho_{1}^{t},\rho_{2}^{t}\right)\qquad\forall\boldsymbol{w}\in\mathbb{R}^{d},\label{eq:three-layers-formal-dynamicsf}\\
\frac{{\rm d}}{{\rm d}t}\beta^{t} & =G_{\beta}\left(f^{t};\rho_{1}^{t},\rho_{2}^{t}\right),\label{eq:three-layers-formal-dynamicstheta}
\end{align}
with $\rho_{1}^{t}={\rm Law}\left(\boldsymbol{w}^{t}\right)$ and
$\rho_{2}^{t}={\rm Law}\left(f^{t},\beta^{t}\right)$, where we define
\begin{align*}
G_{\boldsymbol{w}}\left(\boldsymbol{w};\rho_{1},\rho_{2}\right) & =-\mathbb{E}_{{\cal P}}\left\{ \partial_{2}{\cal L}\left(y,\hat{y}\left(\boldsymbol{x};\rho_{1},\rho_{2}\right)\right)\Delta_{\boldsymbol{w}_{1}}\left(\boldsymbol{w};\boldsymbol{x},\rho_{1},\rho_{2}\right)\right\} ,\\
G_{f}\left(\beta,f,\boldsymbol{w};\rho_{1},\rho_{2}\right) & =-\mathbb{E}_{{\cal P}}\left\{ \partial_{2}{\cal L}\left(y,\hat{y}\left(\boldsymbol{x};\rho_{1},\rho_{2}\right)\right)\Delta_{w_{2}}\left(\beta,f,\boldsymbol{w};\boldsymbol{x},\rho_{1}\right)\right\} ,\\
G_{\beta}\left(f;\rho_{1},\rho_{2}\right) & =-\mathbb{E}_{{\cal P}}\left\{ \partial_{2}{\cal L}\left(y,\hat{y}\left(\boldsymbol{x};\rho_{1},\rho_{2}\right)\right)\Delta_{\beta}\left(f;\boldsymbol{x},\rho_{1}\right)\right\} .
\end{align*}
The evolution is thus described by a system of partial differential
equations with a random initialization.

\subsubsection*{The prediction.}

We state our prediction on the connection between this system and
the three-layers neural network. Given two measures $\rho_{1}^{0}\in\mathscr{P}\left(\mathbb{R}^{d}\right)$
and $\rho_{2}^{0}\in\mathscr{P}\left({\cal F}\times\mathbb{R}\right)$,
we generate $\boldsymbol{W}_{1}^{0}$, $\boldsymbol{W}_{2}^{0}$ and
$\boldsymbol{\beta}^{0}$ as follows. We draw the rows $\left\{ \boldsymbol{w}_{1,i}^{0}\right\} _{i\in\left[n_{1}\right]}$
of $\boldsymbol{W}_{1}^{0}$ i.i.d. from $\rho_{1}^{0}$. We also
draw $n_{2}$ i.i.d. samples $\left\{ f_{j}^{0},\beta_{j}^{0}\right\} _{j\in\left[n_{2}\right]}$
from $\rho_{2}^{0}$ independently. We then form $\boldsymbol{W}_{2}^{0}$
by making $f_{j}^{0}\left(\boldsymbol{w}_{1,i}^{0}\right)$ its $\left(j,i\right)$-th
entry. Finally we form $\boldsymbol{\beta}^{0}=\left(\beta_{1}^{0},...,\beta_{n_{2}}^{0}\right)^{\top}$.
We then run the system initialized at $\rho_{1}^{0}$ and $\rho_{2}^{0}$
to obtain $\rho_{1}^{t}$ and $\rho_{2}^{t}$ for any $t$. We also
train the neural network initialized at $\boldsymbol{W}_{1}^{0}$,
$\boldsymbol{W}_{2}^{0}$ and $\boldsymbol{\beta}^{0}$ to obtain
${\cal W}^{k}$ for any $k$. Our formalism states that for any $t\geq0$,
with $n\to\infty$ (and hence $n_{1},n_{2}\to\infty$) and $\alpha\downarrow0$,
for sufficiently regular (e.g. smooth and bounded) $\phi:\;\mathbb{R}\times\mathbb{R}\mapsto\mathbb{R}$,
\[
\mathbb{E}_{{\cal P}}\left\{ \phi\left(y,\hat{y}_{n}\left(\boldsymbol{x};{\cal W}^{\left\lfloor t/\alpha\right\rfloor }\right)\right)\right\} \to\mathbb{E}_{{\cal P}}\left\{ \phi\left(y,\hat{y}\left(\boldsymbol{x};\rho_{1}^{t},\rho_{2}^{t}\right)\right)\right\} 
\]
in probability over the randomness of initialization and data generation
throughout SGD learning.

In fact, it is our expectation that a more general behavior could
be observed. For example, we expect that for any $t\geq0$, with $n\to\infty$
and $\alpha\downarrow0$,
\[
\mathbb{E}_{{\cal P}_{{\rm test}}}\left\{ \phi\left(y,\hat{y}_{n}\left(\boldsymbol{x};{\cal W}^{\left\lfloor t/\alpha\right\rfloor }\right)\right)\right\} \to\mathbb{E}_{{\cal P}_{{\rm test}}}\left\{ \phi\left(y,\hat{y}\left(\boldsymbol{x};\rho_{1}^{t},\rho_{2}^{t}\right)\right)\right\} 
\]
in probability, where ${\cal P}_{{\rm test}}$ is an out-of-sample
distribution.

\subsection{From three-layers network to the mean field limit: a heuristic derivation\label{subsec:three-layers-derivation}}

To heuristically derive a connection between the three-layers network
and its corresponding formal system, we first state our postulates.

\subsubsection*{The postulates.}

We observe that there is symmetry in the role among the neurons of
the same layer. This symmetry, once attained by proper initialization,
is expected to hold at all subsequent time. This, in particular, suggests
the following two properties:
\begin{enumerate}
\item \textbf{Marginal uniformity:} If a law that governs neuron $i$ of
layer $\ell$ depends on other neurons of layer $\ell$ only through
global statistics of layer $\ell$, then this law applies to all neurons
of layer $\ell$.
\item \textbf{Self-averaging:} One can replace a sum of sufficiently many
terms, which display symmetry in their roles and each of which corresponds
to one neuron from the same layer, with an appropriate integral. More
explicitly, if we associate neuron $i$ among the $n$ neurons of
the same layer with $g\left(x_{i},A_{i}\right)$, where $A_{i}$ is
a random quantity sampled independently from a measure $\mu_{i}$
of neuron $i$, then for sufficiently large $n$,
\begin{equation}
\frac{1}{n}\sum_{i=1}^{n}g\left(x_{i},A_{i}\right)\approx\int g\left(x,a\right)\mu\left({\rm d}a\right)\rho\left({\rm d}x,{\rm d}\mu\right)\label{eq:self-averaging}
\end{equation}
for an appropriate probability measure $\rho$ which plays a surrogate
role for the ensemble of neurons of this layer.
\end{enumerate}
These properties suggest that one can obtain a non-trivial description,
independent of the number of neurons, of the network as $n$ grows
large.

To quantify the above properties, it is necessary to represent each
neuron with a quantity. We propose the following representation, which
accords with the graphical model in Fig. \ref{fig:Three-layers-net}.(b):
\begin{itemize}
\item At the first layer, neuron $i$ is represented by the weight vector
$\boldsymbol{w}_{1,i}\in\mathbb{R}^{d}$ (the $i$-th row of $\boldsymbol{W}_{1}$).
\item At the second layer, neuron $j$ is represented by a stochastic kernel
$\nu_{j}\in\mathscr{K}\left(\mathbb{R}^{d},\mathbb{R}\right)$. Neuron
$j$ generates the weight $w_{2,ji}$ (the $\left(j,i\right)$-th
entry of $\boldsymbol{W}_{2}$) according to $\nu_{j}\left(\cdot\middle|\boldsymbol{w}_{1,i}\right)$.
\item At the third layer, neuron $j$ is represented by the weight $\beta_{j}\in\mathbb{R}$.
\end{itemize}
Representation by the weights $\left\{ \boldsymbol{w}_{1,i}\right\} _{i\in\left[n_{1}\right]}$
and $\left\{ \beta_{j}\right\} _{j\in\left[n_{2}\right]}$ is natural.
The crucial role of stochastic kernels $\left\{ \nu_{j}\right\} _{j\in\left[n_{2}\right]}$
will be clearer later (cf. Section \ref{subsec:Discussions-three-layers}),
even though they do not appear in the description of the formal system
in Section \ref{subsec:three-layers-formalism}.

We are now ready to give a heuristic derivation of the connection
between the three-layers network and the formal system.

\subsubsection*{Forward pass.}

Let us derive Eq. (\ref{eq:three-layers-formal-net}) from Eq. (\ref{eq:three-layers-net}).
Similar to the two-layers case, we have for large $n_{1}$, at each
neuron $j$ of the second layer for $j\in\left[n_{2}\right]$:
\begin{equation}
h_{2,j}=\frac{1}{n_{1}}\left\langle \boldsymbol{w}_{2,j},\sigma\left(\boldsymbol{W}_{1}\boldsymbol{x}\right)\right\rangle \approx\int w_{2}\sigma\left(\left\langle \boldsymbol{w},\boldsymbol{x}\right\rangle \right)\nu_{j}\left({\rm d}w_{2}\middle|\boldsymbol{w}\right)\rho_{1}\left({\rm d}\boldsymbol{w}\right).\label{eq:three-layers-connection-forward-0}
\end{equation}
Here in the approximation, we have replaced the empirical measure
${\rm Emp}\left(\left\{ \boldsymbol{w}_{1,i},w_{2,ji}\right\} _{i\in\left[n_{1}\right]}\right)$
with $\nu_{j}\rho_{1}$, by the self-averaging property. We also note
that unlike the two-layers case, $h_{2,j}$'s for different neuron
$j$'s involve the same $\boldsymbol{W}_{1}$ of the first layer.
This is reflected in the use of $\rho_{1}$ independent of $j$. Now
by setting
\begin{equation}
f_{j}\left(\boldsymbol{w}\right)={\rm CE}\left\{ \nu_{j}\right\} \left(\boldsymbol{w}\right)=\int w_{2}\nu_{j}\left({\rm d}w_{2}\middle|\boldsymbol{w}\right),\label{eq:three-layers-connection-forward-f}
\end{equation}
we obtain
\begin{equation}
h_{2,j}\approx H_{2}\left(f_{j};\boldsymbol{x},\rho_{1}\right).\label{eq:three-layers-connection-forward-1}
\end{equation}
This results in
\[
\hat{y}_{n}\left(\boldsymbol{x};{\cal W}\right)=\frac{1}{n_{2}}\sum_{j=1}^{n_{2}}\beta_{j}\sigma\left(h_{2,j}\right)\approx\frac{1}{n_{2}}\sum_{j=1}^{n_{2}}\beta_{j}\sigma\left(H_{2}\left(f_{j};\boldsymbol{x},\rho_{1}\right)\right).
\]
Applying self-averaging again, we finally obtain, for large $n_{2}$,
\begin{equation}
\hat{y}_{n}\left(\boldsymbol{x};{\cal W}\right)\approx\int\beta\sigma\left(H_{2}\left(f;\boldsymbol{x},\rho_{1}\right)\right)\rho_{2}\left({\rm d}f,{\rm d}\beta\right)=\hat{y}\left(\boldsymbol{x};\rho_{1},\rho_{2}\right).\label{eq:three-layers-connection-forward-2}
\end{equation}
From this derivation, we see that:
\begin{itemize}
\item $\rho_{1}$ is a surrogate measure for ${\rm Emp}\left(\left\{ \boldsymbol{w}_{1,i}\right\} _{i\in\left[n_{1}\right]}\right)$
for the first layer's neurons;
\item $\rho_{2}$ is a surrogate measure for ${\rm Emp}\left(\left\{ f_{j},\beta_{j}\right\} _{j\in\left[n_{2}\right]}\right)$
for the second and third layers' neurons;
\item as per Eq. (\ref{eq:three-layers-connection-forward-f}), the only
information about $\nu_{j}$ that is used to compute the forward pass
is $f_{j}$.
\end{itemize}

\subsubsection*{Backward pass.}

We derive the respective connection between Eq. (\ref{eq:three-layers-formal-gradtheta})-(\ref{eq:three-layers-formal-gradW1})
and Eq. (\ref{eq:three-layers-gradtheta})-(\ref{eq:three-layers-gradW1})
. From Eq. (\ref{eq:three-layers-connection-forward-1}), we have
immediately:
\begin{align}
\left(\tilde{\nabla}_{\boldsymbol{\beta}}\hat{y}_{n}\left(\boldsymbol{x};{\cal W}\right)\right)_{j} & =\sigma\left(h_{2,j}\right)\approx\Delta_{\beta}\left(f_{j};\boldsymbol{x},\rho_{1}\right),\label{eq:three-layers-connection-backward-theta}\\
\left(\tilde{\nabla}_{\boldsymbol{h}_{2}}\hat{y}_{n}\left(\boldsymbol{x};{\cal W}\right)\right)_{j} & =\beta_{j}\sigma'\left(h_{2,j}\right)\approx\Delta_{H_{2}}\left(\beta_{j},f_{j};\boldsymbol{x},\rho_{1}\right),\nonumber 
\end{align}
which gives
\begin{align}
\left(\tilde{\nabla}_{\boldsymbol{W}_{2}}\hat{y}_{n}\left(\boldsymbol{x};{\cal W}\right)\right)_{ji} & =\left(\tilde{\nabla}_{\boldsymbol{h}_{2}}\hat{y}_{n}\left(\boldsymbol{x};{\cal W}\right)\right)_{j}\sigma\left(h_{1,i}\right)\nonumber \\
 & \approx\Delta_{H_{2}}\left(\beta_{j},f_{j};\boldsymbol{x},\rho_{1}\right)\sigma\left(\left\langle \boldsymbol{w}_{1,i},\boldsymbol{x}\right\rangle \right)=\Delta_{w_{2}}\left(\beta_{j},f_{j},\boldsymbol{w}_{1,i};\boldsymbol{x},\rho_{1}\right).\label{eq:three-layers-connection-backward-W2}
\end{align}
Let us consider $\tilde{\nabla}_{\boldsymbol{h}_{1}}\hat{y}_{n}\left(\boldsymbol{x};{\cal W}\right)$:
\begin{align*}
\left(\tilde{\nabla}_{\boldsymbol{h}_{1}}\hat{y}_{n}\left(\boldsymbol{x};{\cal W}\right)\right)_{i} & =\left(\frac{1}{n_{2}}\sum_{j=1}^{n_{2}}w_{2,ji}\left(\tilde{\nabla}_{\boldsymbol{h}_{2}}\hat{y}_{n}\left(\boldsymbol{x};{\cal W}\right)\right)_{j}\right)\sigma'\left(h_{1,i}\right)\\
 & \approx\left(\frac{1}{n_{2}}\sum_{j=1}^{n_{2}}w_{2,ji}\Delta_{H_{2}}\left(\beta_{j},f_{j};\boldsymbol{x},\rho_{1}\right)\right)\sigma'\left(\left\langle \boldsymbol{w}_{1,i},\boldsymbol{x}\right\rangle \right).
\end{align*}
Recall that $f_{j}={\rm CE}\left\{ \nu_{j}\right\} $. By our proposed
representation, given a fixed $\boldsymbol{w}_{1,i}$ we have $w_{2,ji}\sim\nu_{j}\left(\cdot\middle|\boldsymbol{w}_{1,i}\right)$.
Hence we can again apply self-averaging in the following way:
\begin{align}
\frac{1}{n_{2}}\sum_{j=1}^{n_{2}}w_{2,ji}\Delta_{H_{2}}\left(\beta_{j},f_{j};\boldsymbol{x},\rho_{1}\right) & \approx\int w_{2}\Delta_{H_{2}}\left(\beta,{\rm CE}\left\{ \nu\right\} ;\boldsymbol{x},\rho_{1}\right)\nu\left({\rm d}w_{2}\middle|\boldsymbol{w}_{1,i}\right)\mu\left({\rm d}\nu,{\rm d}\beta\right)\label{eq:three-layers-connection-backward-1}\\
 & =\int{\rm CE}\left\{ \nu\right\} \left(\boldsymbol{w}_{1,i}\right)\Delta_{H_{2}}\left(\beta,{\rm CE}\left\{ \nu\right\} ;\boldsymbol{x},\rho_{1}\right)\mu\left({\rm d}\nu,{\rm d}\beta\right),\nonumber 
\end{align}
for a probability measure $\mu$ surrogate for ${\rm Emp}\left(\left\{ \nu_{j},\beta_{j}\right\} _{j\in\left[n_{2}\right]}\right)$.
We make further simplification by observing that the integrand depends
on $\nu$ only through ${\rm CE}\left\{ \nu\right\} $ and recalling
that $\rho_{2}$ is the surrogate measure for ${\rm Emp}\left(\left\{ f_{j},\beta_{j}\right\} _{j\in\left[n_{2}\right]}\right)$:
\[
\left(\tilde{\nabla}_{\boldsymbol{h}_{1}}\hat{y}_{n}\left(\boldsymbol{x};{\cal W}\right)\right)_{i}\approx\left(\int f\left(\boldsymbol{w}_{1,i}\right)\Delta_{H_{2}}\left(\beta,f;\boldsymbol{x},\rho_{1}\right)\rho_{2}\left({\rm d}f,{\rm d}\beta\right)\right)\sigma'\left(\left\langle \boldsymbol{w}_{1,i},\boldsymbol{x}\right\rangle \right)=\Delta_{H_{1}}\left(\boldsymbol{w}_{1,i};\boldsymbol{x},\rho_{1},\rho_{2}\right).
\]
Finally we consider the $i$-th row of $\tilde{\nabla}_{\boldsymbol{W}_{1}}\hat{y}_{n}\left(\boldsymbol{x};{\cal W}\right)$:
\begin{equation}
\left(\tilde{\nabla}_{\boldsymbol{W}_{1}}\hat{y}_{n}\left(\boldsymbol{x};{\cal W}\right)\right)_{i}=\left(\tilde{\nabla}_{\boldsymbol{h}_{1}}\hat{y}_{n}\left(\boldsymbol{x};{\cal W}\right)\right)_{i}\boldsymbol{x}\approx\Delta_{H_{1}}\left(\boldsymbol{w}_{1,i};\boldsymbol{x},\rho_{1},\rho_{2}\right)\boldsymbol{x}=\Delta_{\boldsymbol{w}_{1}}\left(\boldsymbol{w}_{1,i};\boldsymbol{x},\rho_{1},\rho_{2}\right).\label{eq:three-layers-connection-backward-W1}
\end{equation}
We observe that like the forward pass, the only information about
$\nu_{j}$ that is used to compute the backward pass is $f_{j}$.

\subsubsection*{Learning dynamics.}

We derive the evolution dynamics (\ref{eq:three-layers-formal-dynamicsW})-(\ref{eq:three-layers-formal-dynamicstheta})
of the formal system from the SGD dynamics (\ref{eq:three-layers-SGDtheta})-(\ref{eq:three-layers-SGDW1})
of the neural network. First, by identifying $t=k\alpha$ and taking
$\alpha\downarrow0$, one obtains time continuum and in-expectation
property w.r.t. ${\cal P}$ from the SGD dynamics:
\begin{align*}
\frac{{\rm d}}{{\rm d}t}\boldsymbol{\beta}^{t} & =-\mathbb{E}_{{\cal P}}\left\{ \partial_{2}{\cal L}\left(y,\hat{y}_{n}\left(\boldsymbol{x};{\cal W}^{t}\right)\right)\tilde{\nabla}_{\boldsymbol{\beta}}\hat{y}_{n}\left(\boldsymbol{x};{\cal W}^{t}\right)\right\} ,\\
\frac{{\rm d}}{{\rm d}t}\boldsymbol{W}_{2}^{t} & =-\mathbb{E}_{{\cal P}}\left\{ \partial_{2}{\cal L}\left(y,\hat{y}_{n}\left(\boldsymbol{x};{\cal W}^{t}\right)\right)\tilde{\nabla}_{\boldsymbol{W}_{2}}\hat{y}_{n}\left(\boldsymbol{x};{\cal W}^{t}\right)\right\} ,\\
\frac{{\rm d}}{{\rm d}t}\boldsymbol{W}_{1}^{t} & =-\mathbb{E}_{{\cal P}}\left\{ \partial_{2}{\cal L}\left(y,\hat{y}_{n}\left(\boldsymbol{x};{\cal W}^{t}\right)\right)\tilde{\nabla}_{\boldsymbol{W}_{1}}\hat{y}_{n}\left(\boldsymbol{x};{\cal W}^{t}\right)\right\} ,
\end{align*}
in which ${\cal W}^{t}=\left\{ \boldsymbol{W}_{1}^{t},\boldsymbol{W}_{2}^{t},\boldsymbol{\beta}^{t}\right\} $.
We represent the neurons, at time $t$, by $\left\{ \boldsymbol{w}_{1,i}^{t}\right\} _{i\in\left[n_{1}\right]}$,
$\left\{ \nu_{j}^{t}\right\} _{j\in\left[n_{2}\right]}$ and $\left\{ \beta_{j}^{t}\right\} _{j\in\left[n_{2}\right]}$,
in which $\boldsymbol{w}_{1,i}^{t}$ is the $i$-th row of $\boldsymbol{W}_{1}^{t}$,
$\beta_{j}^{t}$ is the $j$-th entry of $\boldsymbol{\beta}^{t}$,
$\nu_{j}^{t}\in\mathscr{K}\left(\mathbb{R}^{d},\mathbb{R}\right)$
and the $\left(j,i\right)$-th entry of $\boldsymbol{W}_{2}^{t}$
being $w_{2,ji}^{t}\sim\nu_{j}^{t}\left(\cdot\middle|\boldsymbol{w}_{1,i}^{t}\right)$.
We also define $f_{j}^{t}={\rm CE}\left\{ \nu_{j}^{t}\right\} $.

From Eq. (\ref{eq:three-layers-connection-forward-2}), (\ref{eq:three-layers-connection-backward-theta}),
(\ref{eq:three-layers-connection-backward-W2}) and (\ref{eq:three-layers-connection-backward-W1}),
letting $\tilde{\rho}_{1}^{t}$ and $\tilde{\rho}_{2}^{t}$ be surrogate
measures for respectively ${\rm Emp}\left(\left\{ \boldsymbol{w}_{1,i}^{t}\right\} _{i\in\left[n_{1}\right]}\right)$
and ${\rm Emp}\left(\left\{ f_{j}^{t},\beta_{j}^{t}\right\} _{j\in\left[n_{2}\right]}\right)$,
it is easy to see that at any time $t\geq0$, for $j\in\left[n_{2}\right]$
and $i\in\left[n_{1}\right]$,
\begin{align}
\frac{{\rm d}}{{\rm d}t}\beta_{j}^{t} & \approx G_{\beta}\left(f_{j}^{t};\tilde{\rho}_{1}^{t},\tilde{\rho}_{2}^{t}\right),\label{eq:three-layers-connection-dynamicstheta}\\
\frac{{\rm d}}{{\rm d}t}w_{2,ji}^{t} & \approx G_{f}\left(\beta_{j}^{t},f_{j}^{t},\boldsymbol{w}_{1,i}^{t};\tilde{\rho}_{1}^{t},\tilde{\rho}_{2}^{t}\right),\label{eq:three-layers-connection-dynamicsW2}\\
\frac{{\rm d}}{{\rm d}t}\boldsymbol{w}_{1,i}^{t} & \approx G_{\boldsymbol{w}}\left(\boldsymbol{w}_{1,i}^{t};\tilde{\rho}_{1}^{t},\tilde{\rho}_{2}^{t}\right).\label{eq:three-layers-connection-dynamicsW1}
\end{align}
Recall that $w_{2,ji}^{t}\sim\nu_{j}^{t}\left(\cdot\middle|\boldsymbol{w}_{1,i}^{t}\right)$.
A key observation is that the right-hand side of Eq. (\ref{eq:three-layers-connection-dynamicsW2})
does not depend on $w_{2,ji}^{t}$. As such, for $\Delta t\to0$ and
any event $E\subseteq\mathbb{R}$,
\[
\nu_{j}^{t+\Delta t}\left(E+G_{f}\left(\beta_{j}^{t},f_{j}^{t},\boldsymbol{w}_{1,i}^{t};\tilde{\rho}_{1}^{t},\tilde{\rho}_{2}^{t}\right)\Delta t\middle|\boldsymbol{w}_{1,i}^{t+\Delta t}\right)\approx\nu_{j}^{t}\left(E\middle|\boldsymbol{w}_{1,i}^{t}\right).
\]
We then get:
\begin{align*}
\frac{{\rm d}}{{\rm d}t}\left(f_{j}^{t}\left(\boldsymbol{w}_{1,i}^{t}\right)\right) & =\lim_{\Delta t\to0}\frac{1}{\Delta t}\left(\int w_{2}\nu_{j}^{t+\Delta t}\left({\rm d}w_{2}\middle|\boldsymbol{w}_{1,i}^{t+\Delta t}\right)-\int w_{2}\nu_{j}^{t}\left({\rm d}w_{2}\middle|\boldsymbol{w}_{1,i}^{t}\right)\right)\\
 & \approx\lim_{\Delta t\to0}\frac{1}{\Delta t}\left(\int\left(w_{2}+G_{f}\left(\beta_{j}^{t},f_{j}^{t},\boldsymbol{w}_{1,i}^{t};\tilde{\rho}_{1}^{t},\tilde{\rho}_{2}^{t}\right)\Delta t\right)\nu_{j}^{t}\left({\rm d}w_{2}\middle|\boldsymbol{w}_{1,i}^{t}\right)-\int w_{2}\nu_{j}^{t}\left({\rm d}w_{2}\middle|\boldsymbol{w}_{1,i}^{t}\right)\right)\\
 & =G_{f}\left(\beta_{j}^{t},f_{j}^{t},\boldsymbol{w}_{1,i}^{t};\tilde{\rho}_{1}^{t},\tilde{\rho}_{2}^{t}\right).
\end{align*}
On the other hand,
\begin{align*}
\frac{{\rm d}}{{\rm d}t}\left(f_{j}^{t}\left(\boldsymbol{w}_{1,i}^{t}\right)\right) & =\left(\partial_{t}f_{j}^{t}\right)\left(\boldsymbol{w}_{1,i}^{t}\right)+\left\langle \nabla f_{j}^{t}\left(\boldsymbol{w}_{1,i}^{t}\right),\frac{{\rm d}}{{\rm d}t}\boldsymbol{w}_{1,i}^{t}\right\rangle \\
 & \approx\left(\partial_{t}f_{j}^{t}\right)\left(\boldsymbol{w}_{1,i}^{t}\right)+\left\langle \nabla f_{j}^{t}\left(\boldsymbol{w}_{1,i}^{t}\right),G_{\boldsymbol{w}}\left(\boldsymbol{w}_{1,i}^{t};\tilde{\rho}_{1}^{t},\tilde{\rho}_{2}^{t}\right)\right\rangle ,
\end{align*}
by Eq. (\ref{eq:three-layers-connection-dynamicsW1}). Hence,
\[
\left(\partial_{t}f_{j}^{t}\right)\left(\boldsymbol{w}_{1,i}^{t}\right)+\left\langle \nabla f_{j}^{t}\left(\boldsymbol{w}_{1,i}^{t}\right),G_{\boldsymbol{w}}\left(\boldsymbol{w}_{1,i}^{t};\tilde{\rho}_{1}^{t},\tilde{\rho}_{2}^{t}\right)\right\rangle \approx G_{f}\left(\beta_{j}^{t},f_{j}^{t},\boldsymbol{w}_{1,i}^{t};\tilde{\rho}_{1}^{t},\tilde{\rho}_{2}^{t}\right).
\]
The marginal uniformity property applied to Eq. (\ref{eq:three-layers-connection-dynamicsW1})
posits that ${\rm Law}\left(\boldsymbol{w}_{1,i}^{t}\right)$ is independent
of $i$, and the self-averaging property then suggests ${\rm Law}\left(\boldsymbol{w}_{1,i}^{t}\right)\approx\tilde{\rho}_{1}^{t}$
in the limit $n\to\infty$. In this case we also have
\begin{equation}
\left(\partial_{t}f_{j}^{t}\right)\left(\boldsymbol{w}\right)+\left\langle \nabla f_{j}^{t}\left(\boldsymbol{w}\right),G_{\boldsymbol{w}}\left(\boldsymbol{w};\tilde{\rho}_{1}^{t},\tilde{\rho}_{2}^{t}\right)\right\rangle \approx G_{f}\left(\beta_{j}^{t},f_{j}^{t},\boldsymbol{w};\tilde{\rho}_{1}^{t},\tilde{\rho}_{2}^{t}\right)\label{eq:three-layers-connection-dynamicsf}
\end{equation}
for $\tilde{\rho}_{1}^{t}$-a.e. $\boldsymbol{w}$. Observe that at
any time $t$, values of $f_{j}^{t}$ at $\boldsymbol{w}\notin{\rm supp}\left(\tilde{\rho}_{1}^{t}\right)$
are used in the computation of neither the forward pass nor the backward
pass. As such, we can extend the dynamic (\ref{eq:three-layers-connection-dynamicsf})
to all $\boldsymbol{w}\in\mathbb{R}^{d}$ without affecting the prediction
stated in Section \ref{subsec:three-layers-formalism}. Applying again
marginal uniformity and self-averaging to Eq. (\ref{eq:three-layers-connection-dynamicstheta})
and (\ref{eq:three-layers-connection-dynamicsf}), we have ${\rm Law}\left(f_{j}^{t},\beta_{j}^{t}\right)\approx\tilde{\rho}_{2}^{t}$
independent of $j$ in the limit $n\to\infty$. If $\tilde{\rho}_{1}^{0}=\rho_{1}^{0}$
and $\tilde{\rho}_{2}^{0}=\rho_{2}^{0}$ then one identifies $\tilde{\rho}_{1}^{t}\approx\rho_{1}^{t}$
and $\tilde{\rho}_{2}^{t}\approx\rho_{2}^{t}$. This completes the
derivation.

Finally we note that the initialization in the prediction statement
in Section \ref{subsec:three-layers-formalism} is sufficient to ensure
firstly that symmetry among the neurons is attained at initialization
and hence at all subsequent time, and secondly $\tilde{\rho}_{1}^{0}=\rho_{1}^{0}$
and $\tilde{\rho}_{2}^{0}=\rho_{2}^{0}$.

\subsection{Discussions\label{subsec:Discussions-three-layers}}

Having established the MF limit and its derivation, we now make several
discussions. These discussions extend in a similar spirit to the case
of general multilayer networks.

\paragraph{Comparison with the two-layers case.}

We remark on two differences, which are not apparent from the last
sections, in the formulations between the two-layers case and the
three-layers (or multilayer) case:
\begin{itemize}
\item In the two-layers case, the population loss $\mathbb{E}_{{\cal P}}\left\{ {\cal L}\left(y,\hat{y}\left(\boldsymbol{x};\rho\right)\right)\right\} $
is convex in $\rho$, if ${\cal L}$ is convex in the second argument.
In the three-layers case, the many-neurons description $\hat{y}\left(\boldsymbol{x};\rho_{1},\rho_{2}\right)$
as per Eq. (\ref{eq:three-layers-formal-net}) is no longer linear
in $\left(\rho_{1},\rho_{2}\right)$ and hence $\mathbb{E}_{{\cal P}}\left\{ {\cal L}\left(y,\hat{y}\left(\boldsymbol{x};\rho_{1},\rho_{2}\right)\right)\right\} $
is generally non-convex in $\left(\rho_{1},\rho_{2}\right)$ (although
it is convex in $\rho_{2}$). This highlights the complexity of multilayer
structures.
\item In the two-layers case, in both the forward and backward passes, self-averaging
is used only at the output $\hat{y}_{n}\left(\boldsymbol{x};{\cal W}\right)\approx\hat{y}\left(\boldsymbol{x};\rho\right)$.
In the three-layers case, self-averaging occurs not only at the output
but also at certain neurons, which is evident from Eq. (\ref{eq:three-layers-connection-forward-0})
and (\ref{eq:three-layers-connection-backward-1}).

Occurrences of self-averaging can be spotted visually from the connectivity
among layers. Compare Fig. \ref{fig:Three-layers-net}.(b) (or Fig.
\ref{fig:Multilayer-net}) against Fig. \ref{fig:Two-layers-net}.(b)
for visualization. Except for the connection between the last layer
and the second last one, all other pairs of adjacent layers are densely
connected. Also notice that the last layer's connectivity is similar
to that of a two-layers network. For a neuron at layer $\ell$, self-averaging
occurs in the forward pass information it receives from layer $\ell-1$,
if layer $\ell$ is not the last layer, provided that this piece of
information assumes the form (\ref{eq:self-averaging}). Likewise,
self-averaging occurs in the backward pass information it receives
from layer $\ell+1$, if layer $\ell+1$ is not the last layer, provided
that this piece of information assumes the form (\ref{eq:self-averaging}).
\end{itemize}

\paragraph{Stochastic kernel representation.}

We remark on the use of stochastic kernels $\left\{ \nu_{j}\right\} _{j\in\left[n_{2}\right]}$.
On one hand, given any $\boldsymbol{W}_{1}$, any $\boldsymbol{W}_{2}$
can be realized by means of the random generation $w_{2,ji}\sim\nu_{j}\left(\cdot\middle|\boldsymbol{w}_{1,i}\right)$
for suitable $\left\{ \nu_{j}\right\} _{j\in\left[n_{2}\right]}$.
On the other hand, this random generation enables the application
of self-averaging to make the approximation (\ref{eq:three-layers-connection-backward-1}),
which is a crucial step in the backward pass analysis.

\paragraph{The scalings.}

We comment on the rationale behind the scalings by the numbers of
neurons $n_{1}$ and $n_{2}$ in the three-layers network (\ref{eq:three-layers-net})
and its gradient update quantities (\ref{eq:three-layers-gradtheta})-(\ref{eq:three-layers-gradW1}).
The principle is, roughly speaking, to maintain each entry in the
pre-activations $\boldsymbol{h}_{1}$ and $\boldsymbol{h}_{2}$, the
output $\hat{y}_{n}\left(\boldsymbol{x};{\cal W}\right)$, as well
as their iteration-to-iteration changes, to be $O\left(1\right)$.
It is done by a simple practice: we normalize any quantity which is
a sum over neurons of the same layer by its number of neurons in both
the forward and backward passes. This effectively enables self-averaging
in light of Eq. (\ref{eq:self-averaging}).

A consequence of this principle is that while the gradient update
of $\boldsymbol{W}_{1}$ has a factor of $n_{1}$ similar to the two-layers
case (recalling Eq. (\ref{eq:three-layers-gradW1}) and Eq. (\ref{eq:two-layers-SGD})),
that of $\boldsymbol{W}_{2}$ is $n_{1}n_{2}$ as per Eq. (\ref{eq:three-layers-gradW2}).
In general, for multilayer networks, the gradient update of a weight
matrix that is not of the first layer is scaled by its total number
of entries. Furthermore, each entry of $\boldsymbol{W}_{2}$ (or any
weight matrices, not of the first layer, in the multilayer case) remains
$O\left(1\right)$, distinct from $\boldsymbol{W}_{1}$ whose rows
must adapt to $\boldsymbol{x}$ and whose entries are therefore not
necessarily $O\left(1\right)$.

\paragraph{On the learning rule.}

While we have used a fixed learning rate for simplicity of presentation,
the formalism can easily incorporate a non-uniform varying learning
rate schedule. Specifically if we modify the SGD updates (\ref{eq:three-layers-SGDtheta})-(\ref{eq:three-layers-SGDW1})
as follows:
\begin{align*}
\boldsymbol{\beta}^{k+1} & =\boldsymbol{\beta}^{k}-\alpha\xi_{0}\left(k\alpha\right)\partial_{2}{\cal L}\left(y^{k},\hat{y}_{n}\left(\boldsymbol{x}^{k};{\cal W}^{k}\right)\right)\tilde{\nabla}_{\boldsymbol{\beta}}\hat{y}_{n}\left(\boldsymbol{x}^{k};{\cal W}^{k}\right),\\
\boldsymbol{W}_{2}^{k+1} & =\boldsymbol{W}_{2}^{k}-\alpha\xi_{2}\left(k\alpha\right)\partial_{2}{\cal L}\left(y^{k},\hat{y}_{n}\left(\boldsymbol{x}^{k};{\cal W}^{k}\right)\right)\tilde{\nabla}_{\boldsymbol{W}_{2}}\hat{y}_{n}\left(\boldsymbol{x}^{k};{\cal W}^{k}\right),\\
\boldsymbol{W}_{1}^{k+1} & =\boldsymbol{W}_{1}^{k}-\alpha\xi_{1}\left(k\alpha\right)\partial_{2}{\cal L}\left(y^{k},\hat{y}_{n}\left(\boldsymbol{x}^{k};{\cal W}^{k}\right)\right)\tilde{\nabla}_{\boldsymbol{W}_{1}}\hat{y}_{n}\left(\boldsymbol{x}^{k};{\cal W}^{k}\right),
\end{align*}
for sufficiently regular functions $\xi_{0}$, $\xi_{1}$ and $\xi_{2}$,
then the evolution dynamics of the formal system (\ref{eq:three-layers-formal-dynamicsW})-(\ref{eq:three-layers-formal-dynamicstheta})
should be adjusted to:
\begin{align*}
\frac{{\rm d}}{{\rm d}t}\boldsymbol{w}^{t} & =\xi_{1}\left(t\right)G_{\boldsymbol{w}}\left(\boldsymbol{w}^{t};\rho_{1}^{t},\rho_{2}^{t}\right),\\
\partial_{t}f^{t}\left(\boldsymbol{w}\right)+\left\langle \nabla f^{t}\left(\boldsymbol{w}\right),G_{\boldsymbol{w}}\left(\boldsymbol{w};\rho_{1}^{t},\rho_{2}^{t}\right)\right\rangle  & =\xi_{2}\left(t\right)G_{f}\left(\beta^{t},f^{t},\boldsymbol{w};\rho_{1}^{t},\rho_{2}^{t}\right)\qquad\forall\boldsymbol{w}\in\mathbb{R}^{d},\\
\frac{{\rm d}}{{\rm d}t}\beta^{t} & =\xi_{0}\left(t\right)G_{\beta}\left(f^{t};\rho_{1}^{t},\rho_{2}^{t}\right).
\end{align*}
The same prediction holds as we take $\alpha\downarrow0$.

\paragraph{Non-fully-connected structures.}

While we have focused entirely on fully-connected networks, we expect
the same principle is applicable to other types of structure that
maintain the same key features. In Appendix \ref{sec:LLN-CNNs}, we
give a brief argument, as well as an experiment, to justify that this
is indeed the case for one example of interest: multilayer convolutional
neural networks (CNNs).

\paragraph{On the local operation.}

One key structure that is exploited here is the summation of the form
(\ref{eq:self-averaging}), which does not explicitly requires a specific
form of local interaction between a weight entry and the pre-activation
of a neuron. Here we are interested in more general local interactions.
For example, recalling the three-layers network (\ref{eq:three-layers-net}),
we consider the following form of the pre-activation $h_{2,j}$ of
neuron $j$ of the second layer:
\[
h_{2,j}=\frac{1}{n_{1}}\sum_{i=1}^{n_{1}}\sigma^{*}\left(w_{2,ji},h_{1,i}\right),\qquad\sigma^{*}:\;\mathbb{R}\times\mathbb{R}\mapsto\mathbb{R}.
\]
The local operation $\sigma^{*}$ reduces to the considered three-layers
case if we set $\sigma^{*}\left(w,h\right)=w\sigma\left(h\right)$.
Since the summation structure is retained, we expect that the choice
of $\sigma^{*}$ does not play a very critical role: for a general
$\sigma^{*}$, under the introduced scalings, the MF limit behavior
can still be observed. This is demonstrated for the case of CNNs in
Appendix \ref{sec:LLN-CNNs}.

\section{Mean field limit in multilayer fully-connected networks\label{sec:MF-Multilayer}}

The development in this section is parallel to Section \ref{subsec:three-layers-formalism}.
We describe the multilayer neural network, as well as its corresponding
formal system and the prediction. This, in particular, specifies the
MF limit of the network. We defer to Appendix \ref{sec:multilayer-derivation}
to give a heuristic derivation.

\subsection{Setting: Multilayer fully-connected networks\label{subsec:multilayer-setting}}

\subsubsection*{Forward pass.}

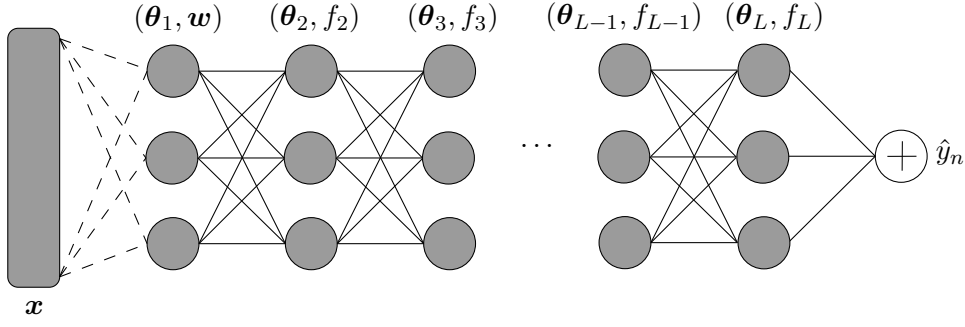
\begin{figure}
\begin{centering}
\begin{center}
\begin{tikzpicture}[yscale=-0.03,xscale=0.03] \draw  [fill={rgb, 255:red, 155; green, 155; blue, 155 }  ,fill opacity=1 ] (92.08,105.1) .. controls (92.08,102.56) and (94.14,100.5) .. (96.68,100.5) -- (110.47,100.5) .. controls (113.01,100.5) and (115.07,102.56) .. (115.07,105.1) -- (115.07,210.76) .. controls (115.07,213.3) and (113.01,215.35) .. (110.47,215.35) -- (96.68,215.35) .. controls (94.14,215.35) and (92.08,213.3) .. (92.08,210.76) -- cycle ; \draw  [fill={rgb, 255:red, 155; green, 155; blue, 155 }  ,fill opacity=1 ] (153.81,119.51) .. controls (153.81,113.12) and (158.96,107.95) .. (165.31,107.95) .. controls (171.65,107.95) and (176.8,113.12) .. (176.8,119.51) .. controls (176.8,125.89) and (171.65,131.06) .. (165.31,131.06) .. controls (158.96,131.06) and (153.81,125.89) .. (153.81,119.51) -- cycle ; \draw  [fill={rgb, 255:red, 155; green, 155; blue, 155 }  ,fill opacity=1 ] (153.3,158.02) .. controls (153.3,151.64) and (158.45,146.47) .. (164.8,146.47) .. controls (171.14,146.47) and (176.29,151.64) .. (176.29,158.02) .. controls (176.29,164.41) and (171.14,169.58) .. (164.8,169.58) .. controls (158.45,169.58) and (153.3,164.41) .. (153.3,158.02) -- cycle ; \draw  [fill={rgb, 255:red, 155; green, 155; blue, 155 }  ,fill opacity=1 ] (153.81,196.03) .. controls (153.81,189.65) and (158.96,184.47) .. (165.31,184.47) .. controls (171.65,184.47) and (176.8,189.65) .. (176.8,196.03) .. controls (176.8,202.41) and (171.65,207.58) .. (165.31,207.58) .. controls (158.96,207.58) and (153.81,202.41) .. (153.81,196.03) -- cycle ; \draw  [fill={rgb, 255:red, 155; green, 155; blue, 155 }  ,fill opacity=1 ] (215.1,119.51) .. controls (215.1,113.12) and (220.24,107.95) .. (226.59,107.95) .. controls (232.94,107.95) and (238.08,113.12) .. (238.08,119.51) .. controls (238.08,125.89) and (232.94,131.06) .. (226.59,131.06) .. controls (220.24,131.06) and (215.1,125.89) .. (215.1,119.51) -- cycle ; \draw  [fill={rgb, 255:red, 155; green, 155; blue, 155 }  ,fill opacity=1 ] (214.59,158.02) .. controls (214.59,151.64) and (219.73,146.47) .. (226.08,146.47) .. controls (232.43,146.47) and (237.57,151.64) .. (237.57,158.02) .. controls (237.57,164.41) and (232.43,169.58) .. (226.08,169.58) .. controls (219.73,169.58) and (214.59,164.41) .. (214.59,158.02) -- cycle ; \draw  [fill={rgb, 255:red, 155; green, 155; blue, 155 }  ,fill opacity=1 ] (215.1,196.03) .. controls (215.1,189.65) and (220.24,184.47) .. (226.59,184.47) .. controls (232.94,184.47) and (238.08,189.65) .. (238.08,196.03) .. controls (238.08,202.41) and (232.94,207.58) .. (226.59,207.58) .. controls (220.24,207.58) and (215.1,202.41) .. (215.1,196.03) -- cycle ; \draw  [fill={rgb, 255:red, 255; green, 255; blue, 255 }  ,fill opacity=1 ] (476.52,157.25) .. controls (476.52,150.87) and (481.66,145.7) .. (488.01,145.7) .. controls (494.35,145.7) and (499.5,150.87) .. (499.5,157.25) .. controls (499.5,163.64) and (494.35,168.81) .. (488.01,168.81) .. controls (481.66,168.81) and (476.52,163.64) .. (476.52,157.25) -- cycle ;
\draw    (438.21,157.25) -- (476.52,157.25) ; \draw    (438.21,118.73) -- (476.52,157.25) ; \draw    (476.52,157.25) -- (438.21,195.26) ; \draw [color={rgb, 255:red, 0; green, 0; blue, 0 }  ,draw opacity=1 ]   (176.8,119.51) -- (215.1,119.51) ; \draw    (176.29,158.02) -- (214.59,158.02) ; \draw    (176.8,196.03) -- (215.1,196.03) ; \draw [color={rgb, 255:red, 0; green, 0; blue, 0 }  ,draw opacity=1 ]   (176.8,119.51) -- (214.59,158.02) ; \draw    (176.29,158.02) -- (215.1,196.03) ; \draw    (214.59,158.28) -- (176.8,196.03) ; \draw [color={rgb, 255:red, 0; green, 0; blue, 0 }  ,draw opacity=1 ]   (215.1,119.76) -- (176.8,157.77) ; \draw  [dash pattern={on 4.5pt off 4.5pt}]  (115.07,105.12) -- (153.81,118.73) ; \draw  [dash pattern={on 4.5pt off 4.5pt}]  (115.07,210.73) -- (153.81,119.51) ; \draw  [dash pattern={on 4.5pt off 4.5pt}]  (115.07,105.12) -- (153.3,158.02) ; \draw  [dash pattern={on 4.5pt off 4.5pt}]  (115.07,210.73) -- (153.3,158.02) ; \draw  [dash pattern={on 4.5pt off 4.5pt}]  (115.07,210.73) -- (153.81,196.03) ; \draw  [dash pattern={on 4.5pt off 4.5pt}]  (115.07,105.12) -- (153.81,196.03) ; \draw [color={rgb, 255:red, 0; green, 0; blue, 0 }  ,draw opacity=1 ]   (176.8,196.03) -- (215.1,119.76) ; \draw [color={rgb, 255:red, 0; green, 0; blue, 0 }  ,draw opacity=1 ]   (176.8,119.51) -- (215.1,196.03) ; \draw    (481.88,157.31) -- (494.77,157.26) ; \draw    (488.42,163.67) -- (488.52,150.6) ;
\draw  [fill={rgb, 255:red, 155; green, 155; blue, 155 }  ,fill opacity=1 ] (276.38,119.51) .. controls (276.38,113.12) and (281.53,107.95) .. (287.88,107.95) .. controls (294.22,107.95) and (299.37,113.12) .. (299.37,119.51) .. controls (299.37,125.89) and (294.22,131.06) .. (287.88,131.06) .. controls (281.53,131.06) and (276.38,125.89) .. (276.38,119.51) -- cycle ; \draw  [fill={rgb, 255:red, 155; green, 155; blue, 155 }  ,fill opacity=1 ] (275.87,158.02) .. controls (275.87,151.64) and (281.02,146.47) .. (287.36,146.47) .. controls (293.71,146.47) and (298.86,151.64) .. (298.86,158.02) .. controls (298.86,164.41) and (293.71,169.58) .. (287.36,169.58) .. controls (281.02,169.58) and (275.87,164.41) .. (275.87,158.02) -- cycle ; \draw  [fill={rgb, 255:red, 155; green, 155; blue, 155 }  ,fill opacity=1 ] (276.38,196.03) .. controls (276.38,189.65) and (281.53,184.47) .. (287.88,184.47) .. controls (294.22,184.47) and (299.37,189.65) .. (299.37,196.03) .. controls (299.37,202.41) and (294.22,207.58) .. (287.88,207.58) .. controls (281.53,207.58) and (276.38,202.41) .. (276.38,196.03) -- cycle ;
\draw [color={rgb, 255:red, 0; green, 0; blue, 0 }  ,draw opacity=1 ]   (238.08,119.51) -- (276.38,119.51) ; \draw    (237.57,158.02) -- (275.87,158.02) ; \draw    (238.08,196.03) -- (276.38,196.03) ; \draw [color={rgb, 255:red, 0; green, 0; blue, 0 }  ,draw opacity=1 ]   (238.08,119.51) -- (275.87,158.02) ; \draw    (237.57,158.02) -- (276.38,196.03) ; \draw    (275.87,158.28) -- (238.08,196.03) ; \draw [color={rgb, 255:red, 0; green, 0; blue, 0 }  ,draw opacity=1 ]   (276.38,119.76) -- (238.08,157.77) ; \draw [color={rgb, 255:red, 0; green, 0; blue, 0 }  ,draw opacity=1 ]   (238.08,196.03) -- (276.38,119.76) ; \draw [color={rgb, 255:red, 0; green, 0; blue, 0 }  ,draw opacity=1 ]   (238.08,119.51) -- (276.38,196.03) ;
\draw  [fill={rgb, 255:red, 155; green, 155; blue, 155 }  ,fill opacity=1 ] (415.7,119.12) .. controls (415.7,112.74) and (420.85,107.56) .. (427.19,107.56) .. controls (433.54,107.56) and (438.68,112.74) .. (438.68,119.12) .. controls (438.68,125.5) and (433.54,130.68) .. (427.19,130.68) .. controls (420.85,130.68) and (415.7,125.5) .. (415.7,119.12) -- cycle ; \draw  [fill={rgb, 255:red, 155; green, 155; blue, 155 }  ,fill opacity=1 ] (415.19,157.64) .. controls (415.19,151.26) and (420.33,146.08) .. (426.68,146.08) .. controls (433.03,146.08) and (438.17,151.26) .. (438.17,157.64) .. controls (438.17,164.02) and (433.03,169.19) .. (426.68,169.19) .. controls (420.33,169.19) and (415.19,164.02) .. (415.19,157.64) -- cycle ; \draw  [fill={rgb, 255:red, 155; green, 155; blue, 155 }  ,fill opacity=1 ] (415.7,195.64) .. controls (415.7,189.26) and (420.85,184.09) .. (427.19,184.09) .. controls (433.54,184.09) and (438.68,189.26) .. (438.68,195.64) .. controls (438.68,202.03) and (433.54,207.2) .. (427.19,207.2) .. controls (420.85,207.2) and (415.7,202.03) .. (415.7,195.64) -- cycle ; \draw  [fill={rgb, 255:red, 155; green, 155; blue, 155 }  ,fill opacity=1 ] (354.11,119.04) .. controls (354.11,112.66) and (359.25,107.49) .. (365.6,107.49) .. controls (371.95,107.49) and (377.09,112.66) .. (377.09,119.04) .. controls (377.09,125.42) and (371.95,130.6) .. (365.6,130.6) .. controls (359.25,130.6) and (354.11,125.42) .. (354.11,119.04) -- cycle ; \draw  [fill={rgb, 255:red, 155; green, 155; blue, 155 }  ,fill opacity=1 ] (353.6,157.56) .. controls (353.6,151.18) and (358.74,146.01) .. (365.09,146.01) .. controls (371.44,146.01) and (376.58,151.18) .. (376.58,157.56) .. controls (376.58,163.94) and (371.44,169.12) .. (365.09,169.12) .. controls (358.74,169.12) and (353.6,163.94) .. (353.6,157.56) -- cycle ; \draw  [fill={rgb, 255:red, 155; green, 155; blue, 155 }  ,fill opacity=1 ] (354.11,195.57) .. controls (354.11,189.19) and (359.25,184.01) .. (365.6,184.01) .. controls (371.95,184.01) and (377.09,189.19) .. (377.09,195.57) .. controls (377.09,201.95) and (371.95,207.12) .. (365.6,207.12) .. controls (359.25,207.12) and (354.11,201.95) .. (354.11,195.57) -- cycle ;
\draw [color={rgb, 255:red, 0; green, 0; blue, 0 }  ,draw opacity=1 ]   (377.09,119.04) -- (415.39,119.04) ; \draw    (376.58,157.56) -- (414.88,157.56) ; \draw    (377.09,195.57) -- (415.39,195.57) ; \draw [color={rgb, 255:red, 0; green, 0; blue, 0 }  ,draw opacity=1 ]   (377.09,119.04) -- (414.88,157.56) ; \draw    (376.58,157.56) -- (415.39,195.57) ; \draw    (414.88,157.82) -- (377.09,195.57) ; \draw [color={rgb, 255:red, 0; green, 0; blue, 0 }  ,draw opacity=1 ]   (415.39,119.3) -- (377.09,157.31) ; \draw [color={rgb, 255:red, 0; green, 0; blue, 0 }  ,draw opacity=1 ]   (377.09,195.57) -- (415.39,119.3) ; \draw [color={rgb, 255:red, 0; green, 0; blue, 0 }  ,draw opacity=1 ]   (377.09,119.04) -- (415.39,195.57) ;
\draw (103.89,224.53) node   {$\boldsymbol{x}$}; \draw (509.84,155.46) node   {$\hat{y}_n$}; \draw (167.73,96.14) node   {$(\boldsymbol{\theta}_{1} ,\boldsymbol{w})$}; \draw (228,95.88) node   {$(\boldsymbol{\theta}_{2} ,f_{2})$}; \draw (289.28,95.88) node   {$(\boldsymbol{\theta}_{3} ,f_{3})$}; \draw (431.74,95.5) node   {$(\boldsymbol{\theta}_{L} ,f_{L})$}; \draw (327.63,152.5) node   {$\dotsc $}; \draw (365.22,95.5) node   {$(\boldsymbol{\theta}_{L-1} ,f_{L-1})$};
\end{tikzpicture} 
\par\end{center}
\par\end{centering}
\caption{A graphical representation of a multilayer neural network, with $L+1$
fully-connected layers. Here neuron $j$ at layer $\ell>1$ is represented
by $\left(\boldsymbol{\theta}_{\ell,j},f_{\ell,j}\right)$ to be consistent
with the information presented in Section \ref{sec:MF-Multilayer},
while we note the actual representation is $\left(\boldsymbol{\theta}_{\ell,j},\nu_{\ell,j}\right)$
for some stochastic kernel $\nu_{\ell,j}$ and $f_{\ell,j}={\rm CE}\left\{ \nu_{\ell,j}\right\} $,
as per the derivation in Appendix \ref{sec:multilayer-derivation}.}
\label{fig:Multilayer-net}
\end{figure}

We describe a neural network with $L$ hidden layers, for a given
$L\geq1$, a collection of integers $\left\{ d,n_{1},n_{2},...,n_{L}\right\} $
and an integer $q\geq1$:
\begin{align}
\hat{y}_{n}\left(\boldsymbol{x};{\cal W}\right) & =\frac{1}{n_{L}}\sum_{i=1}^{n_{L}}\left(\sigma_{L}\left(\boldsymbol{\Theta}_{L},\boldsymbol{h}_{L}\right)\right)_{i},\label{eq:multilayer-net}\\
\boldsymbol{h}_{\ell} & =\frac{1}{n_{\ell-1}}\boldsymbol{W}_{\ell}\sigma_{\ell-1}\left(\boldsymbol{\Theta}_{\ell-1},\boldsymbol{h}_{\ell-1}\right),\qquad\ell=2,...,L,\nonumber \\
\boldsymbol{h}_{1} & =\boldsymbol{W}_{1}\boldsymbol{x},\nonumber 
\end{align}
in which $\boldsymbol{x}\in\mathbb{R}^{d}$ is the input to the network,
$\hat{y}_{n}\left(\boldsymbol{x};{\cal W}\right)\in\mathbb{R}$ is
the output, ${\cal W}=\left\{ \boldsymbol{W}_{1},...,\boldsymbol{W}_{L},\boldsymbol{\Theta}_{1},...,\boldsymbol{\Theta}_{L}\right\} $
is the collection of weights, $\boldsymbol{W}_{1}\in\mathbb{\mathbb{R}}^{n_{1}\times d}$,
$\boldsymbol{W}_{\ell}\in\mathbb{R}^{n_{\ell}\times n_{\ell-1}}$,
$\boldsymbol{\Theta}_{\ell}\in\left(\mathbb{\mathbb{R}}^{q}\right)^{n_{\ell}}$,
and $\sigma_{\ell}:\;\mathbb{\mathbb{R}}^{q}\times\mathbb{R}\mapsto\mathbb{R}$
is a nonlinear activation. (We treat $\boldsymbol{\Theta}_{\ell}$
as a vector of length $n_{\ell}$ with each entry being an element
in $\mathbb{R}^{q}$.) Here $n_{\ell}=n_{\ell}\left(n\right)\to\infty$
as $n\to\infty$. It is a common practice to use $q=3$, $\sigma_{L}\left(\boldsymbol{\theta},h\right)=\theta_{1}\sigma\left(h+\theta_{2}\right)+\theta_{3}$
and $\sigma_{\ell}\left(\boldsymbol{\theta},h\right)=\sigma\left(h+\theta_{1}\right)$
for $\ell<L$ and some scalar nonlinearity $\sigma$, in which case
we obtain the usual $\left(L+1\right)$-layers fully-connected network
with biases. An illustration is given in Fig. \ref{fig:Multilayer-net}.

\subsubsection*{Backward pass.}

Let us define the following derivative quantities:
\begin{align*}
\tilde{\nabla}_{\boldsymbol{\boldsymbol{\Theta}}_{L}}\hat{y}_{n}\left(\boldsymbol{x};{\cal W}\right) & =n_{L}\nabla_{\boldsymbol{\Theta}_{L}}\hat{y}_{n}\left(\boldsymbol{x};{\cal W}\right)=\nabla_{1}\sigma{}_{L}\left(\boldsymbol{\Theta}_{L},\boldsymbol{h}_{L}\right),\\
\tilde{\nabla}_{\boldsymbol{h}_{L}}\hat{y}_{n}\left(\boldsymbol{x};{\cal W}\right) & =n_{L}\nabla_{\boldsymbol{h}_{L}}\hat{y}_{n}\left(\boldsymbol{x};{\cal W}\right)=\partial_{2}\sigma{}_{L}\left(\boldsymbol{\Theta}_{L},\boldsymbol{h}_{L}\right),\\
\tilde{\nabla}_{\boldsymbol{\Theta}_{\ell}}\hat{y}_{n}\left(\boldsymbol{x};{\cal W}\right) & =n_{\ell}\nabla_{\boldsymbol{\Theta}_{\ell}}\hat{y}_{n}\left(\boldsymbol{x};{\cal W}\right)=\left(\frac{1}{n_{\ell+1}}\boldsymbol{W}_{\ell+1}^{\top}\tilde{\nabla}_{\boldsymbol{h}_{\ell+1}}\hat{y}_{n}\left(\boldsymbol{x};{\cal W}\right)\right)\odot\nabla_{1}\sigma_{\ell}\left(\boldsymbol{\Theta}_{\ell},\boldsymbol{h}_{\ell}\right),\\
\tilde{\nabla}_{\boldsymbol{h}_{\ell}}\hat{y}_{n}\left(\boldsymbol{x};{\cal W}\right) & =n_{\ell}\nabla_{\boldsymbol{h}_{\ell}}\hat{y}_{n}\left(\boldsymbol{x};{\cal W}\right)=\left(\frac{1}{n_{\ell+1}}\boldsymbol{W}_{\ell+1}^{\top}\tilde{\nabla}_{\boldsymbol{h}_{\ell+1}}\hat{y}_{n}\left(\boldsymbol{x};{\cal W}\right)\right)\odot\partial_{2}\sigma_{\ell}\left(\boldsymbol{\Theta}_{\ell},\boldsymbol{h}_{\ell}\right),\qquad\ell=L-1,...,1,\\
\tilde{\nabla}_{\boldsymbol{W}_{\ell}}\hat{y}_{n}\left(\boldsymbol{x};{\cal W}\right) & =n_{\ell}n_{\ell-1}\nabla_{\boldsymbol{W}_{\ell}}\hat{y}_{n}\left(\boldsymbol{x};{\cal W}\right)=\left(\tilde{\nabla}_{\boldsymbol{h}_{\ell}}\hat{y}_{n}\left(\boldsymbol{x};{\cal W}\right)\right)\sigma_{\ell-1}\left(\boldsymbol{\Theta}_{\ell-1},\boldsymbol{h}_{\ell-1}\right)^{\top},\qquad\ell=L,...,2,\\
\tilde{\nabla}_{\boldsymbol{W}_{1}}\hat{y}_{n}\left(\boldsymbol{x};{\cal W}\right) & =n_{1}\nabla_{\boldsymbol{W}_{1}}\hat{y}_{n}\left(\boldsymbol{x};{\cal W}\right)=\left(\tilde{\nabla}_{\boldsymbol{h}_{1}}\hat{y}_{n}\left(\boldsymbol{x};{\cal W}\right)\right)\boldsymbol{x}^{\top}.
\end{align*}
Notice the scalings by $n_{\ell}$ and $n_{\ell}n_{\ell-1}$.

\subsubsection*{Learning dynamics.}

We assume that at each time $k\in\mathbb{N}$, the data $\left(\boldsymbol{x}^{k},y^{k}\right)\in\mathbb{R}^{d}\times\mathbb{R}$
is drawn independently from a probabilistic source ${\cal P}$. We
train the network with the loss ${\cal L}\left(y,\hat{y}_{n}\left(\boldsymbol{x};{\cal W}\right)\right)$
for a loss function ${\cal L}:\;\mathbb{R}\times\mathbb{R}\mapsto\mathbb{R}$,
using SGD with an initialization ${\cal W}^{0}=\left\{ \boldsymbol{W}_{1}^{0},...,\boldsymbol{W}_{L}^{0},\boldsymbol{\Theta}_{1}^{0},...,\boldsymbol{\Theta}_{L}^{0}\right\} $
and a learning rate $\alpha>0$:
\begin{align*}
\boldsymbol{W}_{\ell}^{k+1} & =\boldsymbol{W}_{\ell}^{k}-\alpha\partial_{2}{\cal L}\left(y^{k},\hat{y}_{n}\left(\boldsymbol{x}^{k};{\cal W}^{k}\right)\right)\tilde{\nabla}_{\boldsymbol{W}_{\ell}}\hat{y}_{n}\left(\boldsymbol{x}^{k};{\cal W}^{k}\right),\qquad\ell=1,...,L,\\
\boldsymbol{\Theta}_{\ell}^{k+1} & =\boldsymbol{\Theta}_{\ell}^{k}-\alpha\partial_{2}{\cal L}\left(y^{k},\hat{y}_{n}\left(\boldsymbol{x}^{k};{\cal W}^{k}\right)\right)\tilde{\nabla}_{\boldsymbol{\Theta}_{\ell}}\hat{y}_{n}\left(\boldsymbol{x}^{k};{\cal W}^{k}\right),\qquad\ell=1,...,L.
\end{align*}
This yields the dynamics of ${\cal W}^{k}=\left\{ \boldsymbol{W}_{1}^{k},...,\boldsymbol{W}_{L}^{k},\boldsymbol{\Theta}_{1}^{k},...,\boldsymbol{\Theta}_{L}^{k}\right\} $.

\subsection{Mean field limit\label{subsec:multilayer-formalism}}

Similar to Section \ref{subsec:three-layers-formalism}, we describe
a time-evolving system which does not involve the numbers of neurons
$n_{1},...,n_{L}$. This leads to a MF limit which characterizes the
behavior of the multilayer network (\ref{eq:multilayer-net}) during
learning in the limit $n\to\infty$ via this formal system. Before
all, let ${\cal F}_{1}=\mathbb{R}^{d}$ and ${\cal F}_{\ell}=\left\{ f\middle|f:\;\mathbb{R}^{q}\times{\cal F}_{\ell-1}\mapsto\mathbb{R}\right\} $
a vector space for $\ell=2,...,L$.

\subsubsection*{Forward pass.}

The forward pass of the formal system is defined by the following:
\begin{align}
\hat{y}\left(\boldsymbol{x};\underbar{\ensuremath{\rho}}\right) & =\int\sigma_{L}\left(\boldsymbol{\theta},H_{L}\left(f;\boldsymbol{x},\left\{ \rho_{i}\right\} _{i=1}^{L-1}\right)\right)\rho_{L}\left({\rm d}\boldsymbol{\theta},{\rm d}f\right),\label{eq:multilayer-formal-net}
\end{align}
where we define inductively:
\begin{align*}
H_{1}\left(\boldsymbol{w};\boldsymbol{x}\right) & =\left\langle \boldsymbol{w},\boldsymbol{x}\right\rangle ,\\
H_{2}\left(f;\boldsymbol{x},\rho_{1}\right) & =\int f\left(\boldsymbol{\theta},\boldsymbol{w}\right)\sigma_{1}\left(\boldsymbol{\theta},H_{1}\left(\boldsymbol{w};\boldsymbol{x}\right)\right)\rho_{1}\left({\rm d}\boldsymbol{\theta},{\rm d}\boldsymbol{w}\right),\\
H_{\ell}\left(f;\boldsymbol{x},\left\{ \rho_{i}\right\} _{i=1}^{\ell-1}\right) & =\int f\left(\boldsymbol{\theta},g\right)\sigma_{\ell-1}\left(\boldsymbol{\theta},H_{\ell-1}\left(g;\boldsymbol{x},\left\{ \rho_{i}\right\} _{i=1}^{\ell-2}\right)\right)\rho_{\ell-1}\left({\rm d}\boldsymbol{\theta},{\rm d}g\right),\qquad\ell=3,...,L,
\end{align*}
for which $\underbar{\ensuremath{\rho}}=\left\{ \rho_{i}\right\} _{i=1}^{L}$,
$\rho_{\ell}\in\mathscr{P}\left(\mathbb{R}^{q}\times{\cal F}_{\ell}\right)$
for $\ell=1,...,L-1$. Specifically, $\text{\ensuremath{\underbar{\ensuremath{\rho}}}}$
is the state of the system, and the system takes $\boldsymbol{x}\in\mathbb{R}^{d}$
as input and outputs $\hat{y}\left(\boldsymbol{x};\underbar{\ensuremath{\rho}}\right)\in\mathbb{R}$.
One should compare Eq. (\ref{eq:multilayer-formal-net}) with the
multilayer network (\ref{eq:multilayer-net}).

\subsubsection*{Backward pass.}

We define the quantities $\Delta_{\boldsymbol{\theta},\ell}$ and
$\Delta_{H,\ell}$ for $\ell=1,...,L$ as follows. First we define
the quantities for $\ell=L$:
\begin{align*}
\Delta_{\boldsymbol{\theta},L}\left(\boldsymbol{\theta},f;\boldsymbol{x},\left\{ \rho_{i}\right\} _{i=1}^{L-1}\right) & =\nabla_{1}\sigma_{L}\left(\boldsymbol{\theta},H_{L}\left(f;\boldsymbol{x},\left\{ \rho_{i}\right\} _{i=1}^{L-1}\right)\right),\\
\Delta_{H,L}\left(\boldsymbol{\theta},f;\boldsymbol{x},\left\{ \rho_{i}\right\} _{i=1}^{L-1}\right) & =\partial_{2}\sigma_{L}\left(\boldsymbol{\theta},H_{L}\left(f;\boldsymbol{x},\left\{ \rho_{i}\right\} _{i=1}^{L-1}\right)\right).
\end{align*}
Then we define the rest inductively:
\begin{align*}
\Delta_{\boldsymbol{\theta},L-1}\left(\boldsymbol{\theta},f;\boldsymbol{x},\underbar{\ensuremath{\rho}}\right) & =\left(\int g\left(\boldsymbol{\theta},f\right)\Delta_{H,L}\left(\boldsymbol{\theta}',g;\boldsymbol{x},\left\{ \rho_{i}\right\} _{i=1}^{L-1}\right)\rho_{L}\left({\rm d}\boldsymbol{\theta}',{\rm d}g\right)\right)\nabla_{1}\sigma_{L-1}\left(\boldsymbol{\theta},H_{L-1}\left(f;\boldsymbol{x},\left\{ \rho_{i}\right\} _{i=1}^{L-2}\right)\right),\\
\Delta_{H,L-1}\left(\boldsymbol{\theta},f;\boldsymbol{x},\underbar{\ensuremath{\rho}}\right) & =\left(\int g\left(\boldsymbol{\theta},f\right)\Delta_{H,L}\left(\boldsymbol{\theta}',g;\boldsymbol{x},\left\{ \rho_{i}\right\} _{i=1}^{L-1}\right)\rho_{L}\left({\rm d}\boldsymbol{\theta}',{\rm d}g\right)\right)\partial_{2}\sigma_{L-1}\left(\boldsymbol{\theta},H_{L-1}\left(f;\boldsymbol{x},\left\{ \rho_{i}\right\} _{i=1}^{L-2}\right)\right),\\
\Delta_{\boldsymbol{\theta},\ell}\left(\boldsymbol{\theta},f;\boldsymbol{x},\underbar{\ensuremath{\rho}}\right) & =\left(\int g\left(\boldsymbol{\theta},f\right)\Delta_{H,\ell+1}\left(\boldsymbol{\theta}',g;\boldsymbol{x},\underbar{\ensuremath{\rho}}\right)\rho_{\ell+1}\left({\rm d}\boldsymbol{\theta}',{\rm d}g\right)\right)\nabla_{1}\sigma_{\ell}\left(\boldsymbol{\theta},H_{\ell}\left(f;\boldsymbol{x},\left\{ \rho_{i}\right\} _{i=1}^{\ell-1}\right)\right),\\
\Delta_{H,\ell}\left(\boldsymbol{\theta},f;\boldsymbol{x},\underbar{\ensuremath{\rho}}\right) & =\left(\int g\left(\boldsymbol{\theta},f\right)\Delta_{H,\ell+1}\left(\boldsymbol{\theta}',g;\boldsymbol{x},\underbar{\ensuremath{\rho}}\right)\rho_{\ell+1}\left({\rm d}\boldsymbol{\theta}',{\rm d}g\right)\right)\partial_{2}\sigma_{\ell}\left(\boldsymbol{\theta},H_{\ell}\left(f;\boldsymbol{x},\left\{ \rho_{i}\right\} _{i=1}^{\ell-1}\right)\right),\\
 & \qquad\ell=L-2,...,2,\\
\Delta_{\boldsymbol{\theta},1}\left(\boldsymbol{\theta},\boldsymbol{w};\boldsymbol{x},\underbar{\ensuremath{\rho}}\right) & =\left(\int g\left(\boldsymbol{\theta},\boldsymbol{w}\right)\Delta_{H,2}\left(\boldsymbol{\theta}',g;\boldsymbol{x},\underbar{\ensuremath{\rho}}\right)\rho_{2}\left({\rm d}\boldsymbol{\theta}',{\rm d}g\right)\right)\nabla_{1}\sigma_{1}\left(\boldsymbol{\theta},H_{1}\left(\boldsymbol{w};\boldsymbol{x}\right)\right),\\
\Delta_{H,1}\left(\boldsymbol{\theta},f;\boldsymbol{x},\underbar{\ensuremath{\rho}}\right) & =\left(\int g\left(\boldsymbol{\theta},\boldsymbol{w}\right)\Delta_{H,2}\left(\boldsymbol{\theta}',g;\boldsymbol{x},\underbar{\ensuremath{\rho}}\right)\rho_{2}\left({\rm d}\boldsymbol{\theta}',{\rm d}g\right)\right)\partial_{2}\sigma_{1}\left(\boldsymbol{\theta},H_{1}\left(\boldsymbol{w};\boldsymbol{x}\right)\right).
\end{align*}
From these quantities, we define $\Delta_{W,\ell}$ for $\ell=1,...,L$:
\begin{align*}
\Delta_{W,L}\left(\boldsymbol{\theta},f,\boldsymbol{\theta}',g;\boldsymbol{x},\left\{ \rho_{i}\right\} _{i=1}^{L-1}\right) & =\Delta_{H,L}\left(\boldsymbol{\theta},f;\boldsymbol{x},\left\{ \rho_{i}\right\} _{i=1}^{L-1}\right)\sigma_{L-1}\left(\boldsymbol{\theta}',H_{L-1}\left(g;\boldsymbol{x},\left\{ \rho_{i}\right\} _{i=1}^{L-2}\right)\right),\\
\Delta_{W,\ell}\left(\boldsymbol{\theta},f,\boldsymbol{\theta}',g;\boldsymbol{x},\underbar{\ensuremath{\rho}}\right) & =\Delta_{H,\ell}\left(\boldsymbol{\theta},f;\boldsymbol{x},\underbar{\ensuremath{\rho}}\right)\sigma_{\ell-1}\left(\boldsymbol{\theta}',H_{\ell-1}\left(g;\boldsymbol{x},\left\{ \rho_{i}\right\} _{i=1}^{\ell-2}\right)\right),\qquad\ell=L-1,...,3,\\
\Delta_{W,2}\left(\boldsymbol{\theta},f,\boldsymbol{\theta}',\boldsymbol{w};\boldsymbol{x},\underbar{\ensuremath{\rho}}\right) & =\Delta_{H,2}\left(\boldsymbol{\theta},f;\boldsymbol{x},\underbar{\ensuremath{\rho}}\right)\sigma_{1}\left(\boldsymbol{\theta}',H_{1}\left(\boldsymbol{w};\boldsymbol{x}\right)\right),\\
\Delta_{W,1}\left(\boldsymbol{\theta},\boldsymbol{w};\boldsymbol{x},\underbar{\ensuremath{\rho}}\right) & =\Delta_{H,1}\left(\boldsymbol{\theta},\boldsymbol{w};\boldsymbol{x},\underbar{\ensuremath{\rho}}\right)\boldsymbol{x}.
\end{align*}
As a note, except for $\Delta_{W,1}$ whose range is $\mathbb{R}^{d}$
and $\Delta_{\boldsymbol{\theta},\ell}$ whose range is $\mathbb{R}^{q}$
for $\ell=1,...,L$, all other derivative quantities map to $\mathbb{R}$.

\subsubsection*{Evolution dynamics.}

We describe a continuous-time evolution dynamics of the system, defined
at each time $t$ via $\underbar{\ensuremath{\rho}}^{t}=\left\{ \rho_{\ell}^{t}\right\} _{\ell=1}^{L}$
where $\rho_{\ell}^{t}\in\mathscr{P}\left(\mathbb{R}^{q}\times{\cal F}_{\ell}\right)$
for $\ell=1,...,L$. First we define
\begin{align*}
G_{\boldsymbol{\theta},1}\left(\boldsymbol{\theta},\boldsymbol{w};\underbar{\ensuremath{\rho}}\right) & =-\mathbb{E}_{{\cal P}}\left\{ \partial_{2}{\cal L}\left(y,\hat{y}\left(\boldsymbol{x};\underbar{\ensuremath{\rho}}\right)\right)\Delta_{\boldsymbol{\theta},1}\left(\boldsymbol{\theta},\boldsymbol{w};\boldsymbol{x},\underbar{\ensuremath{\rho}}\right)\right\} ,\\
G_{\boldsymbol{\theta},\ell}\left(\boldsymbol{\theta},f;\underbar{\ensuremath{\rho}}\right) & =-\mathbb{E}_{{\cal P}}\left\{ \partial_{2}{\cal L}\left(y,\hat{y}\left(\boldsymbol{x};\underbar{\ensuremath{\rho}}\right)\right)\Delta_{\boldsymbol{\theta},\ell}\left(\boldsymbol{\theta},f;\boldsymbol{x},\underbar{\ensuremath{\rho}}\right)\right\} ,\qquad\ell=2,...,L-1,\\
G_{\boldsymbol{\theta},L}\left(\boldsymbol{\theta},f;\underbar{\ensuremath{\rho}}\right) & =-\mathbb{E}_{{\cal P}}\left\{ \partial_{2}{\cal L}\left(y,\hat{y}\left(\boldsymbol{x};\underbar{\ensuremath{\rho}}\right)\right)\Delta_{\boldsymbol{\theta},L}\left(\boldsymbol{\theta},f;\boldsymbol{x},\left\{ \rho_{i}\right\} _{i=1}^{L-1}\right)\right\} ,\\
G_{W,1}\left(\boldsymbol{\theta},\boldsymbol{w};\underbar{\ensuremath{\rho}}\right) & =-\mathbb{E}_{{\cal P}}\left\{ \partial_{2}{\cal L}\left(y,\hat{y}\left(\boldsymbol{x};\underbar{\ensuremath{\rho}}\right)\right)\Delta_{W,1}\left(\boldsymbol{\theta},\boldsymbol{w};\boldsymbol{x},\underbar{\ensuremath{\rho}}\right)\right\} ,\\
G_{f,2}\left(\boldsymbol{\theta},f,\boldsymbol{\theta}',\boldsymbol{w};\underbar{\ensuremath{\rho}}\right) & =-\mathbb{E}_{{\cal P}}\left\{ \partial_{2}{\cal L}\left(y,\hat{y}\left(\boldsymbol{x};\underbar{\ensuremath{\rho}}\right)\right)\Delta_{W,2}\left(\boldsymbol{\theta},f,\boldsymbol{\theta}',\boldsymbol{w};\boldsymbol{x},\underbar{\ensuremath{\rho}}\right)\right\} ,\\
G_{f,\ell}\left(\boldsymbol{\theta},f,\boldsymbol{\theta}',g;\underbar{\ensuremath{\rho}}\right) & =-\mathbb{E}_{{\cal P}}\left\{ \partial_{2}{\cal L}\left(y,\hat{y}\left(\boldsymbol{x};\underbar{\ensuremath{\rho}}\right)\right)\Delta_{W,\ell}\left(\boldsymbol{\theta},f,\boldsymbol{\theta}',g;\boldsymbol{x},\underbar{\ensuremath{\rho}}\right)\right\} ,\qquad\ell=3,...,L-1,\\
G_{f,L}\left(\boldsymbol{\theta},f,\boldsymbol{\theta}',g;\underbar{\ensuremath{\rho}}\right) & =-\mathbb{E}_{{\cal P}}\left\{ \partial_{2}{\cal L}\left(y,\hat{y}\left(\boldsymbol{x};\underbar{\ensuremath{\rho}}\right)\right)\Delta_{W,L}\left(\boldsymbol{\theta},f,\boldsymbol{\theta}',g;\boldsymbol{x},\left\{ \rho_{i}\right\} _{i=1}^{L-1}\right)\right\} .
\end{align*}
In addition, for each $\ell=2,...,L-1$, we define ${\cal G}_{\ell}:\;\mathbb{R}^{q}\times{\cal F}_{\ell}\times\mathbb{R}^{q}\times{\cal F}_{\ell-1}\mapsto\mathbb{R}$
such that, inductively,
\begin{align*}
{\cal G}_{2}\left(\boldsymbol{\theta},f,\boldsymbol{\theta}',\boldsymbol{w};\underbar{\ensuremath{\rho}}\right) & =G_{f,2}\left(\boldsymbol{\theta},f,\boldsymbol{\theta}',\boldsymbol{w};\underbar{\ensuremath{\rho}}\right)-\left\langle \nabla_{1}f\left(\boldsymbol{\theta}',\boldsymbol{w}\right),G_{\boldsymbol{\theta},1}\left(\boldsymbol{\theta}',\boldsymbol{w};\underbar{\ensuremath{\rho}}\right)\right\rangle -\left\langle \nabla_{2}f\left(\boldsymbol{\theta}',\boldsymbol{w}\right),G_{W,1}\left(\boldsymbol{\theta}',\boldsymbol{w};\underbar{\ensuremath{\rho}}\right)\right\rangle ,\\
{\cal G}_{\ell}\left(\boldsymbol{\theta},f,\boldsymbol{\theta}',g;\underbar{\ensuremath{\rho}}\right) & =G_{f,\ell}\left(\boldsymbol{\theta},f,\boldsymbol{\theta}',g;\underbar{\ensuremath{\rho}}\right)-\left\langle \nabla_{1}f\left(\boldsymbol{\theta}',g\right),G_{\boldsymbol{\theta},\ell-1}\left(\boldsymbol{\theta}',g;\underbar{\ensuremath{\rho}}\right)\right\rangle -\mathscr{D}_{2}f\left\{ \boldsymbol{\theta}',g\right\} \left({\cal G}_{f,\ell-1}\left(\boldsymbol{\theta}',g,\cdot,\cdot;\underbar{\ensuremath{\rho}}\right)\right),\\
 & \qquad\ell=3,...,L.
\end{align*}
The evolution dynamics is then defined by the following differential
equations:
\begin{align*}
\frac{{\rm d}}{{\rm d}t}\boldsymbol{w}^{t} & =G_{W,1}\left(\boldsymbol{\theta}_{1}^{t},\boldsymbol{w}^{t};\underbar{\ensuremath{\rho}}^{t}\right),\\
\frac{{\rm d}}{{\rm d}t}\boldsymbol{\theta}_{1}^{t} & =G_{\boldsymbol{\theta},1}\left(\boldsymbol{\theta}_{1}^{t},\boldsymbol{w}^{t};\underbar{\ensuremath{\rho}}^{t}\right),\\
\partial_{t}f_{2}^{t}\left(\boldsymbol{\theta},\boldsymbol{w}\right) & ={\cal G}_{2}\left(\boldsymbol{\theta}_{2}^{t},f_{2}^{t},\boldsymbol{\theta},\boldsymbol{w};\underbar{\ensuremath{\rho}}^{t}\right)\qquad\forall\left(\boldsymbol{\theta},\boldsymbol{w}\right)\in\mathbb{R}^{q}\times{\cal F}_{1},\\
\frac{{\rm d}}{{\rm d}t}\boldsymbol{\theta}_{\ell}^{t} & =G_{\boldsymbol{\theta},\ell}\left(\boldsymbol{\theta}_{\ell}^{t},f_{\ell}^{t};\underbar{\ensuremath{\rho}}^{t}\right),\qquad\ell=2,...,L,\\
\partial_{t}f_{\ell}^{t}\left(\boldsymbol{\theta},g\right) & ={\cal G}_{f,\ell}\left(\boldsymbol{\theta}_{\ell}^{t},f_{\ell}^{t},\boldsymbol{\theta},g;\underbar{\ensuremath{\rho}}^{t}\right)\qquad\forall\left(\boldsymbol{\theta},g\right)\in\mathbb{R}^{q}\times{\cal F}_{\ell-1},\;\ell=3,...,L,
\end{align*}
for $\left(\boldsymbol{\theta}_{1}^{t},\boldsymbol{w}^{t}\right)\sim\rho_{1}^{t}$,
$\left(\boldsymbol{\theta}_{\ell}^{t},f_{\ell}^{t}\right)\sim\rho_{\ell}^{t}$
for $\ell=2,...,L$, and $\underbar{\ensuremath{\rho}}^{t}=\left\{ \rho_{\ell}^{t}\right\} _{\ell=1}^{L}$.
More specifically, given $\text{\ensuremath{\underbar{\ensuremath{\rho}}}}^{0}=\left\{ \rho_{\ell}^{0}\right\} _{\ell=1}^{L}$
where $\rho_{\ell}^{0}\in\mathscr{P}\left(\mathbb{R}^{q}\times{\cal F}_{\ell}\right)$,
we generate $\left(\boldsymbol{\theta}_{1}^{0},\boldsymbol{w}^{0}\right)\sim\rho_{1}^{0}$
and $\left(\boldsymbol{\theta}_{\ell}^{0},f_{\ell}^{0}\right)\sim\rho_{\ell}^{0}$
for $\ell=2,...,L$. Taking them as the initialization, we let $\boldsymbol{w}^{t}$,
$\left\{ \boldsymbol{\theta}_{\ell}^{t}\right\} _{\ell=1}^{L}$ and
$\left\{ f_{\ell}^{t}\right\} _{\ell=2}^{L}$ evolve according to
the aforementioned differential equations, with $\rho_{1}^{t}={\rm Law}\left(\boldsymbol{\theta}_{1}^{t},\boldsymbol{w}^{t}\right)$
and $\rho_{\ell}^{t}={\rm Law}\left(\boldsymbol{\theta}_{\ell}^{t},f_{\ell}^{t}\right)$
for $\ell=2,...,L$.

\subsubsection*{The prediction.}

We state our prediction on the connection between the formal system
and the multilayer neural network. First, given $\underbar{\ensuremath{\rho}}^{0}=\left\{ \rho_{\ell}^{0}:\;\rho_{\ell}^{0}\in\mathscr{P}\left(\mathbb{R}^{q}\times{\cal F}_{\ell}\right)\right\} _{\ell=1}^{L}$,
we generate ${\cal W}^{0}=\left\{ \boldsymbol{W}_{1}^{0},...,\boldsymbol{W}_{L}^{0},\boldsymbol{\Theta}_{1}^{0},...,\boldsymbol{\Theta}_{L}^{0}\right\} $
for the neural network as follows. We draw $\left\{ \boldsymbol{\theta}_{1,i}^{0},\boldsymbol{w}_{1,i}^{0}\right\} _{i\in\left[n_{1}\right]}$
i.i.d. from $\rho_{1}^{0}$, where $\boldsymbol{\theta}_{1,i}^{0}$
is the $i$-th element of $\boldsymbol{\Theta}_{1}^{0}$ and $\boldsymbol{w}_{1,i}^{0}$
is the $i$-th row of $\boldsymbol{W}_{1}^{0}$. We also independently
draw $n_{\ell}$ i.i.d. samples $\left\{ \boldsymbol{\theta}_{\ell,j}^{0},f_{\ell,j}^{0}\right\} _{j\in\left[n_{\ell}\right]}$
from $\rho_{\ell}^{0}$, for $\ell=2,...,L$. We then form $\boldsymbol{\Theta}_{\ell}^{0}$
by using $\boldsymbol{\theta}_{\ell,i}^{0}$ as its $i$-th element.
We also form $\boldsymbol{W}_{\ell}^{0}$ by letting its $\left(j,i\right)$-th
entry equal $f_{\ell,j}^{0}\left(\boldsymbol{\theta}_{\ell-1,i}^{0},f_{\ell-1,i}^{0}\right)$
if $\ell\geq3$ and $f_{2,j}^{0}\left(\boldsymbol{\theta}_{1,i}^{0},\boldsymbol{w}_{1,i}^{0}\right)$
if $\ell=2$.

Given the above initialization, we run the formal system initialized
at $\underbar{\ensuremath{\rho}}^{0}$ to obtain $\underbar{\ensuremath{\rho}}^{t}$
for any $t$. We also train the neural network initialized at ${\cal W}^{0}$
to obtain ${\cal W}^{k}$ for any $k$. Our formalism states that
for any $t\geq0$, with $n\to\infty$ (and hence $n_{1},...,n_{L}\to\infty$)
and $\alpha\downarrow0$, for sufficiently regular (e.g. smooth and
bounded) $\phi:\;\mathbb{R}\times\mathbb{R}\mapsto\mathbb{R}$,
\[
\mathbb{E}_{{\cal P}}\left\{ \phi\left(y,\hat{y}_{n}\left(\boldsymbol{x};{\cal W}^{\left\lfloor t/\alpha\right\rfloor }\right)\right)\right\} \to\mathbb{E}_{{\cal P}}\left\{ \phi\left(y,\hat{y}\left(\boldsymbol{x};\underbar{\ensuremath{\rho}}^{t}\right)\right)\right\} 
\]
in probability over the randomness of initialization and data generation
throughout SGD learning.

Similar to the three-layers case in Section \ref{subsec:three-layers-formalism},
we expect to observe a more general behavior, for example, that for
any $t\geq0$, with $n\to\infty$ and $\alpha\downarrow0$,
\[
\mathbb{E}_{{\cal P}_{{\rm test}}}\left\{ \phi\left(y,\hat{y}_{n}\left(\boldsymbol{x};{\cal W}^{\left\lfloor t/\alpha\right\rfloor }\right)\right)\right\} \to\mathbb{E}_{{\cal P}_{{\rm test}}}\left\{ \phi\left(y,\hat{y}\left(\boldsymbol{x};\underbar{\ensuremath{\rho}}^{t}\right)\right)\right\} 
\]
in probability, where ${\cal P}_{{\rm test}}$ is an out-of-sample
distribution.

\section{Validation of the formalism\label{sec:Validation}}

In this section, we perform validation tests on the MF limit behavior
as predicted by the formalism.

\subsection{Statics: a theoretical justification}

As a first test, we ask whether the forward pass description of the
formalism is meaningful, in particular, whether one can obtain a result
similar to Eq. (\ref{eq:two-layers-statics}) of the two-layers case.
We shall argue that it is indeed the case, in particular,
\[
\lim_{n\to\infty}\inf_{{\cal W}}\mathbb{E}_{{\cal P}}\left\{ {\cal L}\left(y,\hat{y}_{n}\left(\boldsymbol{x};{\cal W}\right)\right)\right\} =\inf_{\underbar{\ensuremath{\rho}}}\mathbb{E}_{{\cal P}}\left\{ {\cal L}\left(y,\hat{y}\left(\boldsymbol{x};\underbar{\ensuremath{\rho}}\right)\right)\right\} 
\]
under suitable conditions. Here we recall the multilayer network (\ref{eq:multilayer-net})
and its formal system (\ref{eq:multilayer-formal-net}) from Section
\ref{sec:MF-Multilayer}, and also that $n_{\ell}=n_{\ell}\left(n\right)\to\infty$
as $n\to\infty$. This establishes an asymptotic equivalence between
the global optimum of the network and that of the formal system. We
defer this task to Appendix \ref{sec:multilayer-statics}.

\subsection{Dynamics: an experimental justification\label{subsec:Dynamics-validation}}

We present a second test, which aims to validate the predictions as
stated in Section \ref{subsec:three-layers-formalism} and Section
\ref{subsec:multilayer-formalism}, via experiments. In particular,
we would like to verify whether the evolution curve of a multilayer
network approaches some non-trivial limiting curve as its number of
neurons grows large. As a reminder, our experimental settings are
not tuned to attain competitive performances since it is not our goal.

\subsubsection{Experimental setting}

We shall mainly consider the following three supervised learning tasks:
\begin{itemize}
\item Isotropic Gaussians classification: this is an artificial 2-classes
dataset, considered in \cite{mei2018mean}. The data is generated
as follow: $y\sim{\rm Unif}\left(\left\{ -1,+1\right\} \right)$,
and $\boldsymbol{x}|y\sim\mathsf{N}\left(0,\left(1+y\Delta\right)^{2}\boldsymbol{I}_{d}\right)$,
for some $\Delta\in\left(0,1\right)$. Here we choose $d=32$ and
$\Delta=0.4$. Note that no linear classifiers can attain non-trivial
performance on this problem. We use the squared loss ${\cal L}\left(y_{1},y_{2}\right)=\left(y_{1}-y_{2}\right)^{2}$.
We measure the population loss $\mathbb{E}_{{\cal P}}\left\{ {\cal L}\left(y,\hat{y}_{n}\left(\boldsymbol{x};{\cal W}\right)\right)\right\} $
and the classification error ${\cal P}\left(y\hat{y}_{n}\left(\boldsymbol{x};{\cal W}\right)<0\right)$,
estimated by Monte-Carlo averaging over $10^{4}$ samples.
\item MNIST classification: this is the popular MNIST 10-classes dataset.
We normalize all gray-scale pixels in the image to the range $\left[-1,+1\right]$.
We use the cross-entropy loss ${\cal L}$, and use the whole training
set of size $60\times10^{3}$. Here $d=28\times28=784$. We measure
the loss and the classification error, averaged over $10^{4}$ samples
drawn from the training set or the test set.
\item CIFAR-10 classification: this is the popular CIFAR-10 dataset with
10 classes. We normalize each RGB value in the image to the range
$\left[-1,+1\right]$. We use the cross-entropy loss ${\cal L}$,
and use the whole training set of size $50\times10^{3}$. Here $d=3\times32\times32=3072$.
We measure the loss and the classification error, averaged over $10^{4}$
samples drawn from the training set or the test set.
\end{itemize}
To further the validation, the following task is also considered:
\begin{itemize}
\item CIFAR-10 classification with VGG16 features: the setting is almost
the same as the above CIFAR-10 task, except that instead of the raw
CIFAR-10 images, we use the features which are computed by the convolutional
layers of the VGG16 network \cite{simonyan2014very} pre-trained on
the ImageNet dataset \cite{russakovsky2015imagenet}. We first upscale
the images to the size $128\times128\times3$, then feed them into
the VGG16 network to extract the features of dimension $d=4\times4\times512=8192$.
Note that the VGG16 network is not under our scalings; only the networks
that we train on the VGG16 features employ the scalings.
\end{itemize}
We use the usual structure (with scalings) for an $\left(L+1\right)$-layers
network:
\[
\hat{y}_{n}\left(\boldsymbol{x};{\cal W}\right)=\frac{1}{n_{L}}\boldsymbol{\beta}\sigma\left(\boldsymbol{b}_{L}+\frac{1}{n_{L-1}}\boldsymbol{W}_{L}\sigma\left(...\frac{1}{n_{1}}\boldsymbol{W}_{2}\sigma\left(\boldsymbol{b}_{1}+\boldsymbol{W}_{1}\boldsymbol{x}\right)\right)\right)+\boldsymbol{b}_{L+1},
\]
where the weight matrices are $\boldsymbol{W}_{1}\in\mathbb{\mathbb{R}}^{n_{1}\times d}$
and $\boldsymbol{W}_{\ell}\in\mathbb{R}^{n_{\ell}\times n_{\ell-1}}$
for $\ell=2,...,L$, the output weight is $\boldsymbol{\beta}\in\mathbb{R}^{n_{{\rm out}}\times n_{L}}$,
the biases are $\boldsymbol{b}_{\ell}\in\mathbb{R}^{n_{\ell}}$ for
$\ell=1,...,L$ and $\boldsymbol{b}_{L+1}\in\mathbb{R}^{n_{{\rm out}}}$,
and $\sigma$ is the nonlinearity. We observe that this fits into
the general description (\ref{eq:multilayer-net}) of multilayer networks,
except for that $n_{{\rm out}}$ can be larger than $1$. Here the
isotropic Gaussians classification task has $n_{{\rm out}}=1$, whereas
the other two tasks have $n_{{\rm out}}=10$. While strictly speaking
this is not covered by our theory, it can be extended easily to $n_{{\rm out}}>1$,
provided that $n_{{\rm out}}$ remains a finite constant as $n\to\infty$.
We shall also restrict our experiments to uniform widths $n_{1}=...=n_{L}=n$.
The scalings for gradient updates can also be easily deduced from
Section \ref{subsec:multilayer-setting}.

In the experiments, unless otherwise stated, we use mini-batch SGD
to perform training. Note that while we develop our theory for SGD
learning dynamics in Sections \ref{sec:MF-three-layers} and \ref{sec:MF-Multilayer},
an inspection of their derivations reveals that the batch size does
not play a role in the asymptotic $n\to\infty$ and hence does not
affect the MF limit. In finite-size simulations, the use of batch
size larger than $1$ has the advantage of smoother evolution. We
shall use a batch size of $100$ in our experiments.

We now make a remark on the initialization practice. As the formalism
suggests, and also as discussed in Section \ref{subsec:Discussions-three-layers},
the entries of the weight matrices $\boldsymbol{W}_{\ell}$ at layer
$\ell>1$ and the output weight $\boldsymbol{\beta}$ should be of
the order $O\left(1\right)$. Therefore, we do the following initialization:
\begin{gather*}
\left(\boldsymbol{W}_{1}^{0}\right)_{ij}\sim\mathsf{N}\left(0,\frac{\tau_{1}^{2}}{d}\right),\quad\left(\boldsymbol{W}_{\ell}^{0}\right)_{ij}\sim\mathsf{N}\left(\mu_{2},\tau_{2}^{2}\right)\text{ for }\ell=2...,L,\quad\left(\boldsymbol{\beta}^{0}\right)_{ij}\sim\mathsf{N}\left(\mu_{2},\tau_{2}^{2}\right),\\
b_{\ell,i}^{0}\sim\mathsf{N}\left(\mu_{3},\tau_{3}^{2}\right)\text{ for }\ell=1...,L+1.
\end{gather*}
This initialization is different from the usual practice, due to the
introduced scalings. A quick calculation shows that this initialization
may appear degenerate in the following sense: for each $\boldsymbol{x}$,
in the limit $n\to\infty$, the output $\hat{y}_{n}\left(\boldsymbol{x};{\cal W}^{0}\right)$
at initialization becomes independent of $\tau_{2}$, and if $\tau_{3}=0$,
it converges to a non-random quantity, unlike the usual practice.
This is not an issue, since it corresponds to a non-trivial initialization
$\underbar{\ensuremath{\rho}}^{0}$ of the formal system. In general,
we use $\mu_{2}\neq0$.

\subsubsection{Experimental results}

\begin{figure}
\begin{centering}
\includegraphics[width=0.5\columnwidth]{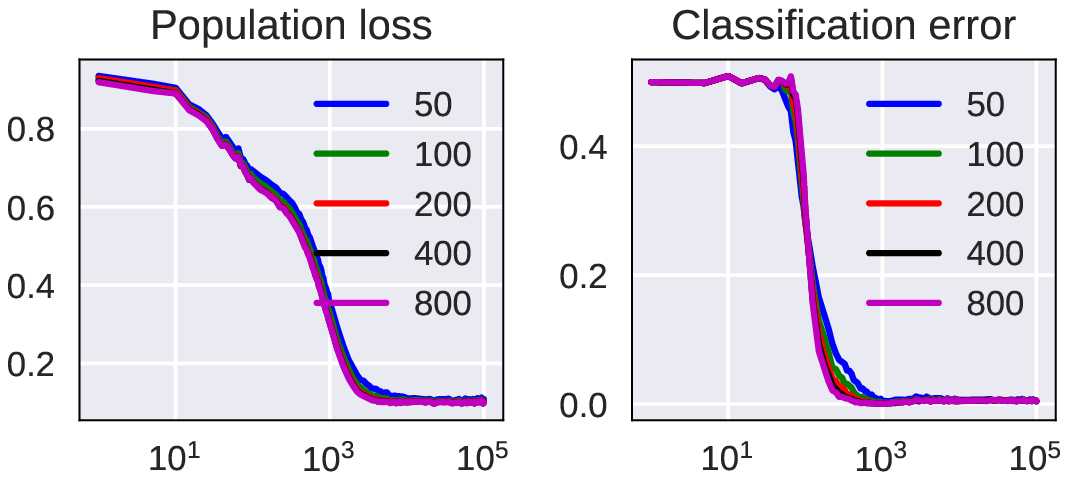}
\par\end{centering}
\begin{centering}
\includegraphics[width=0.5\columnwidth]{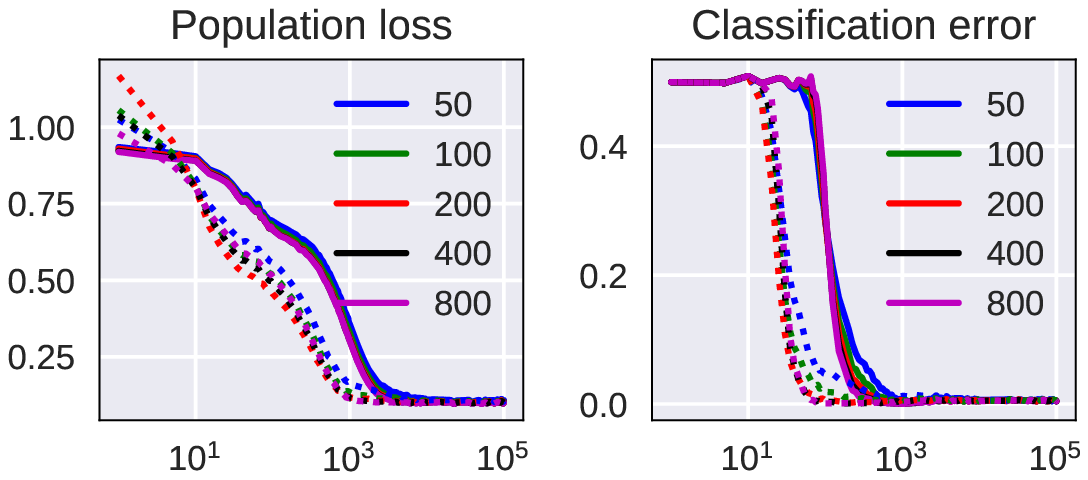}
\par\end{centering}
\caption{The performance of five $5$-layers fully-connected networks on isotropic
Gaussians classification, plotted against training iteration. Here
for each network, $n=50,100,200,400,800$ respectively, $\sigma$
is the ReLU, and the learning rate $\alpha=0.001.$ Top row: we initialize
with $\tau_{1}=\sqrt{2}$, $\mu_{2}=1$, $\tau_{2}=0.1$, $\mu_{3}=0$
and $\tau_{3}=0$. Bottom row: aside from the same initialization
(solid lines), we perform another initialization that differs by $\tau_{2}=3$
(dotted lines).}
\label{fig:isotropic_vary_n}
\end{figure}

\begin{figure}
\begin{centering}
\includegraphics[width=1\columnwidth]{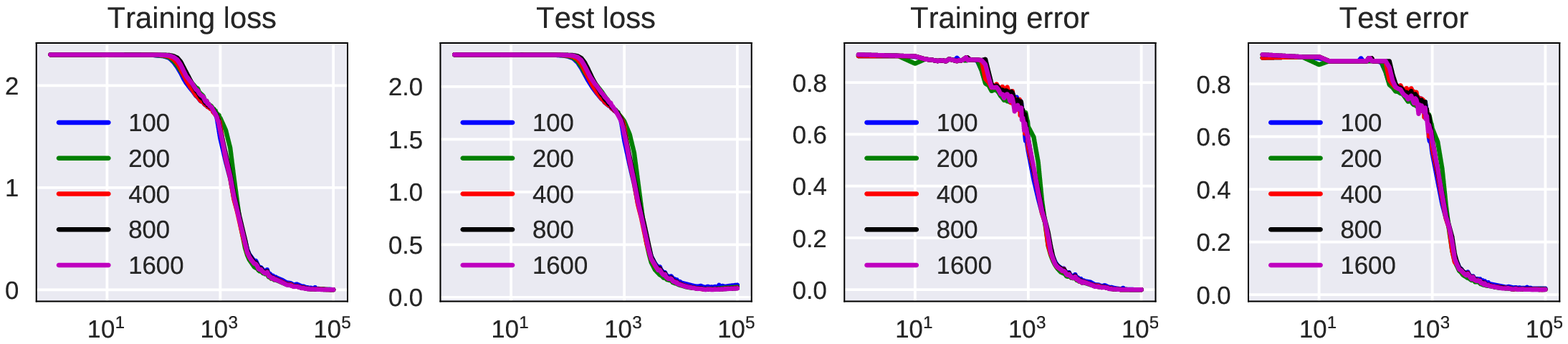}
\par\end{centering}
\begin{centering}
\includegraphics[width=1\columnwidth]{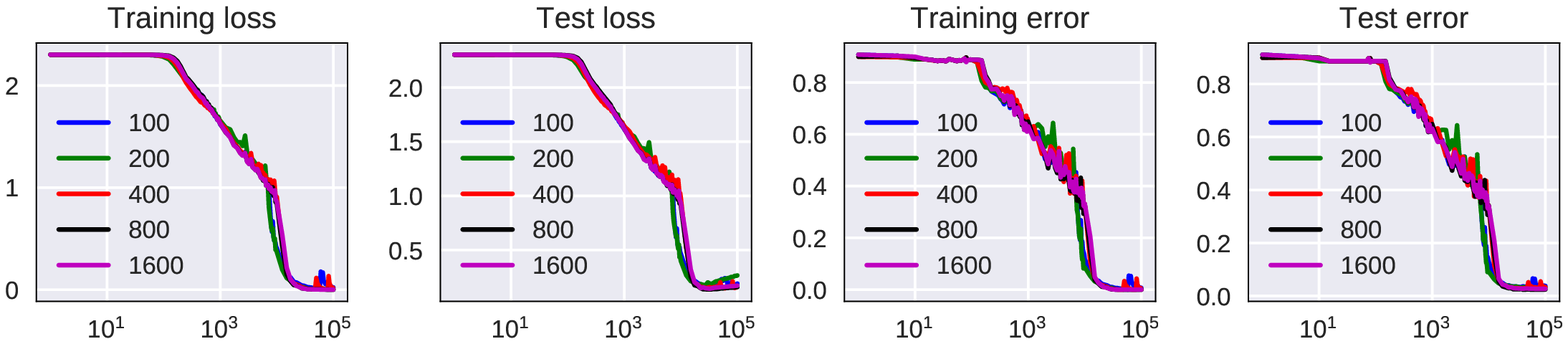}
\par\end{centering}
\begin{centering}
\includegraphics[width=1\columnwidth]{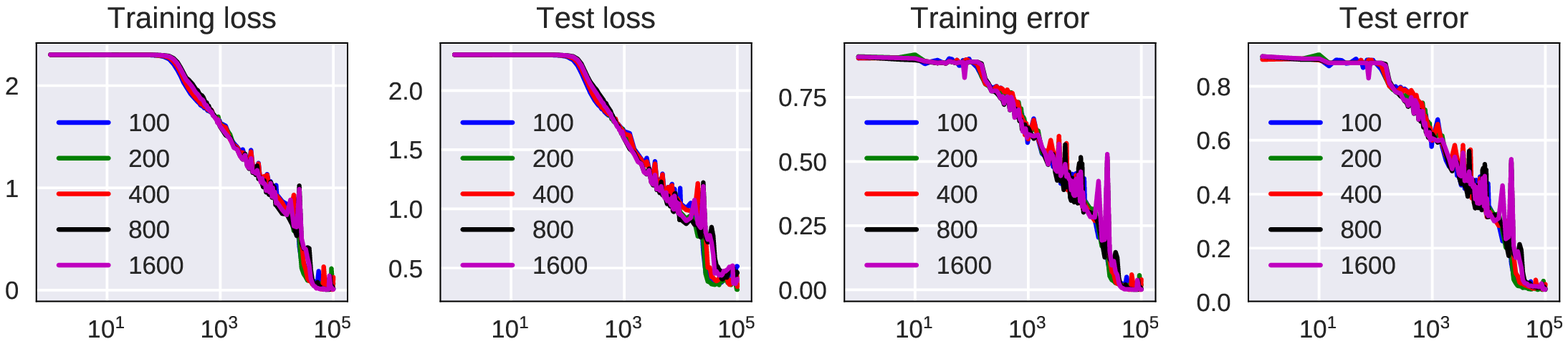}
\par\end{centering}
\caption{The performance of five fully-connected networks on MNIST classification,
plotted against training iteration, in each plot. Here for each network,
$n=100,200,400,800,1600$ respectively, $\sigma$ is the ReLU, and
the learning rate $\alpha=0.01$. We initialize with $\tau_{1}=\sqrt{2}$,
$\mu_{2}=1$, $\tau_{2}=0.1$, $\mu_{3}=0$ and $\tau_{3}=0$. From
the top row: 3-layers networks, 4-layers networks and 5-layers networks.}
\label{fig:mnist_vary_n}
\end{figure}

\begin{figure}
\begin{centering}
\includegraphics[width=1\columnwidth]{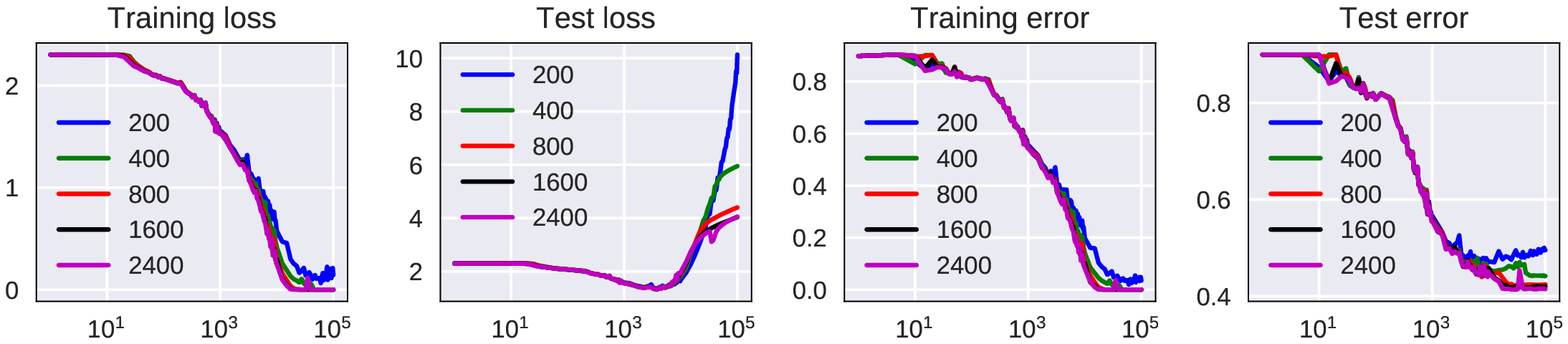}
\par\end{centering}
\begin{centering}
\includegraphics[width=1\columnwidth]{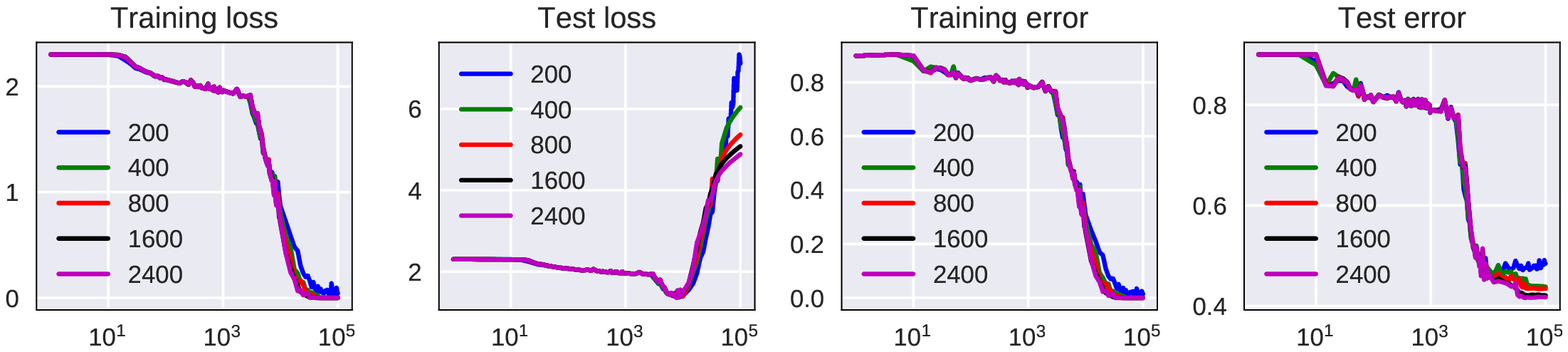}
\par\end{centering}
\begin{centering}
\includegraphics[width=1\columnwidth]{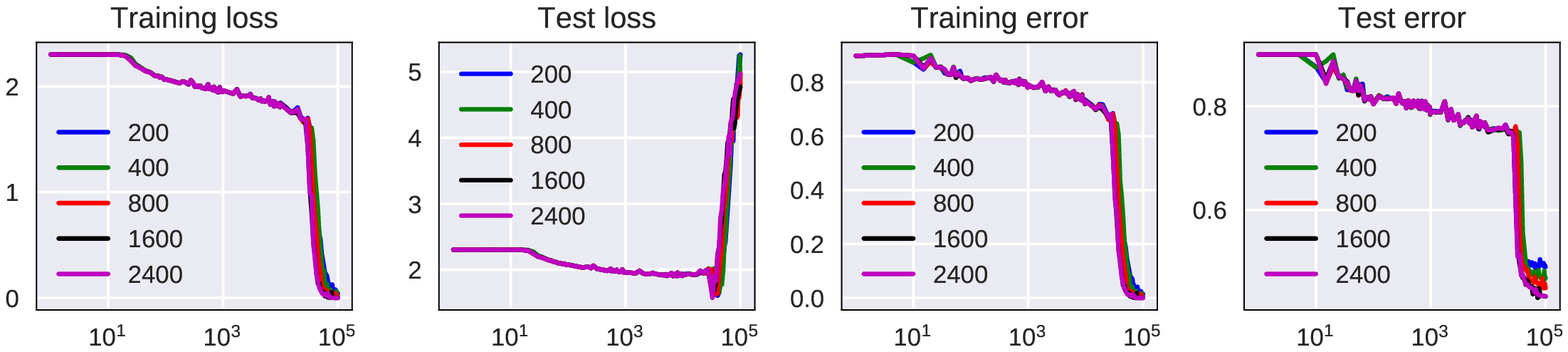}
\par\end{centering}
\caption{The performance of five fully-connected networks on CIFAR-10 classification,
plotted against training iteration, in each plot. For each network,
$n=200,400,800,1600,2400$ respectively, $\sigma$ is the ReLU, and
the learning rate $\alpha=0.07.$ We initialize with $\tau_{1}=\sqrt{2}$,
$\mu_{2}=1$, $\tau_{2}=0.1$, $\mu_{3}=0$ and $\tau_{3}=0$. From
the top row: 3-layers networks, 4-layers networks and 5-layers networks.}
\label{fig:cifar_vary_n}
\end{figure}

In Fig. \ref{fig:isotropic_vary_n}, \ref{fig:mnist_vary_n} and \ref{fig:cifar_vary_n},
we present the results for several fully-connected networks of different
$n$ and $L$ on the three classification tasks, with $\sigma$ being
the rectifier linear unit (ReLU) $\sigma\left(u\right)=\max\left\{ 0,u\right\} $.
We observe that in each plot, the curves become increasingly matching
on the whole training period as $n$ increases, even when the networks
overfit in the CIFAR-10 task. We also see that the networks display
highly nonlinear dynamics and attain non-trivial performances by the
end of training. In particular, it is evident from the isotropic Gaussians
task that the trained networks must exploit the nonlinearity of the
ReLU \textendash{} even though they are initialized to be completely
in the linear region of the ReLU of layer $\ell>1$. This is because
otherwise the resultant classifiers would be linear and cannot attain
close to zero classification error. For the other two tasks, at each
layer, we count the number of non-positive pre-activation entries
and average it over the test set. Our simulations similarly indicate
that by the end of training, this number amounts to a significant
fraction, and therefore, the networks also exploit the nonlinearity
of the ReLU.

The performance on the real datasets is realistic and not trivialized
by the introduced scalings. To illustrate the point, we note that
the work \cite{pennington2017resurrecting} reports of a 200-layers
vanilla fully-connected network, which attains a test error rate of
more than $45\%$ on the CIFAR-10 dataset. This network is initialized
with i.i.d. Gaussian weights of zero mean and carefully selected variance,
is trained without regularization, but is not under our scalings.
We contrast this with the 4-layers network with $n=800$ in Fig. \ref{fig:cifar_vary_n},
which achieves a similar test error rate of about $43\%$.

\begin{figure}
\begin{centering}
\includegraphics[width=1\columnwidth]{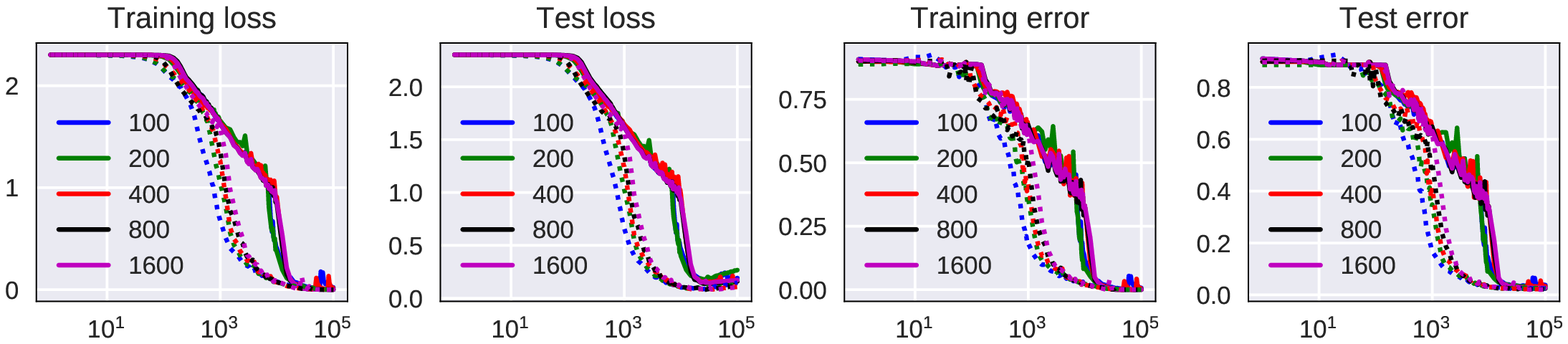}
\par\end{centering}
\caption{The performance of 4-layers fully-connected networks on MNIST classification,
plotted against training iteration, in each plot. For each network,
$n=100,200,400,800,1600$ respectively, $\sigma$ is the ReLU, and
the learning rate $\alpha=0.01.$ We initialize with $\tau_{1}=\sqrt{2}$,
$\mu_{2}=1$, $\mu_{3}=0$, $\tau_{3}=0$, with $\tau_{2}=0.1$ for
the solid lines and $\tau_{2}=2$ for the dotted lines.}
\label{fig:mnist_vary_init}
\end{figure}

\begin{figure}
\begin{centering}
\includegraphics[width=1\columnwidth]{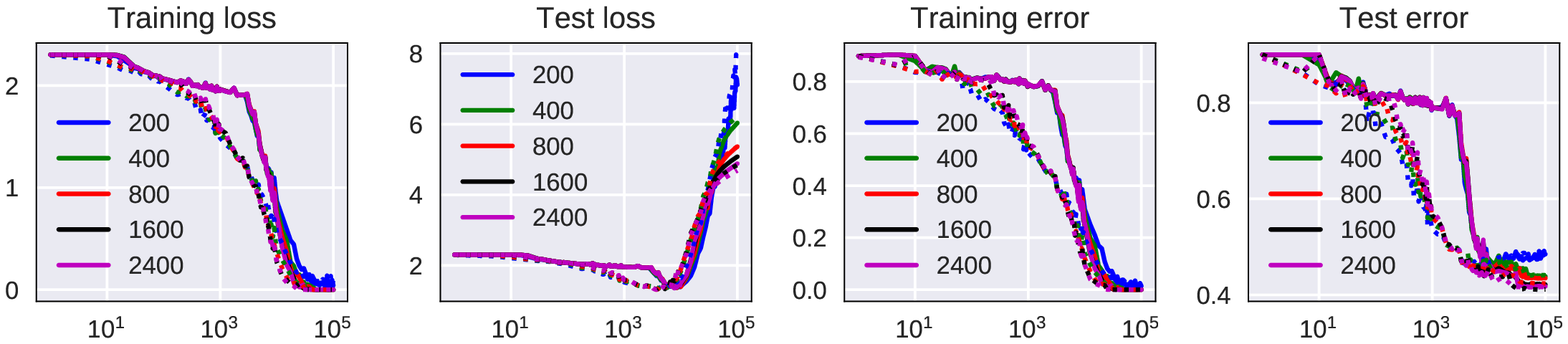}
\par\end{centering}
\caption{The performance of 4-layers fully-connected networks on CIFAR-10 classification,
plotted against training iteration, in each plot. For each network,
$n=200,400,800,1600,2400$ respectively, $\sigma$ is the ReLU, and
the learning rate $\alpha=0.07.$ We initialize with $\tau_{1}=\sqrt{2}$,
$\mu_{2}=1$, $\mu_{3}=0$, $\tau_{3}=0$, with $\tau_{2}=0.1$ for
the solid lines and $\tau_{2}=2$ for the dotted lines.}
\label{fig:cifar_vary_init}
\end{figure}

In Fig. \ref{fig:isotropic_vary_n}, \ref{fig:mnist_vary_init} and
\ref{fig:cifar_vary_init}, we compare the evolutions of two initializations
which differ only by the choice of $\tau_{2}$, but share the same
$\tau_{3}=0$. Recall that due to the way we initialize the networks,
the value of $\tau_{2}$ does not affect the initial values of the
network output or the pre-activations in the limit $n\to\infty$.
Despite this fact, we observe in Fig. \ref{fig:isotropic_vary_n},
\ref{fig:mnist_vary_init} and \ref{fig:cifar_vary_init} that the
two initializations yield two different trajectories. This is consistent
with our formalism: each $\tau_{2}$ gives rise to a different initialization
$\text{\ensuremath{\underbar{\ensuremath{\rho}}}}^{0}$ of the formal
system and hence a different evolution trajectory.

\begin{figure}
\begin{centering}
\vspace*{-0.5cm}
\par\end{centering}
\begin{centering}
\includegraphics[width=0.5\columnwidth]{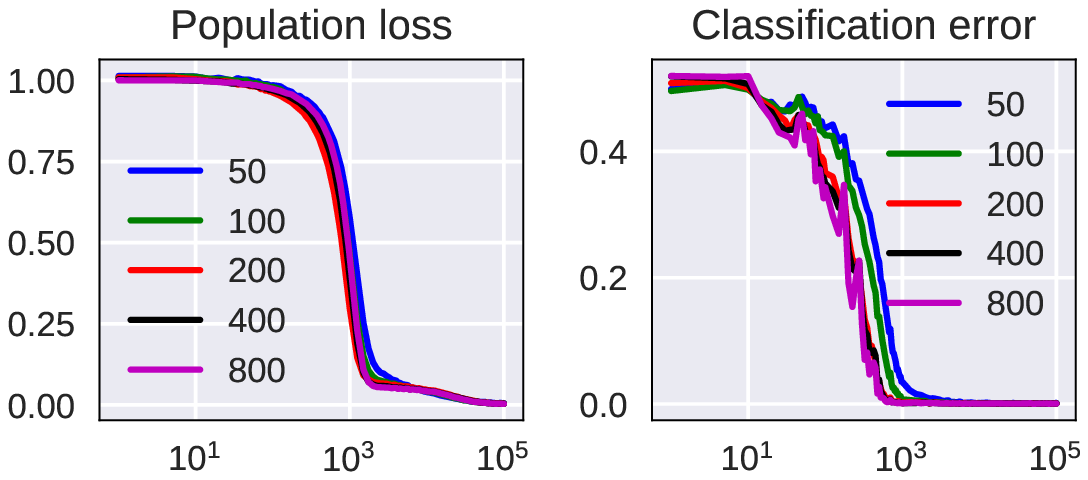}
\par\end{centering}
\begin{centering}
\includegraphics[width=1\columnwidth]{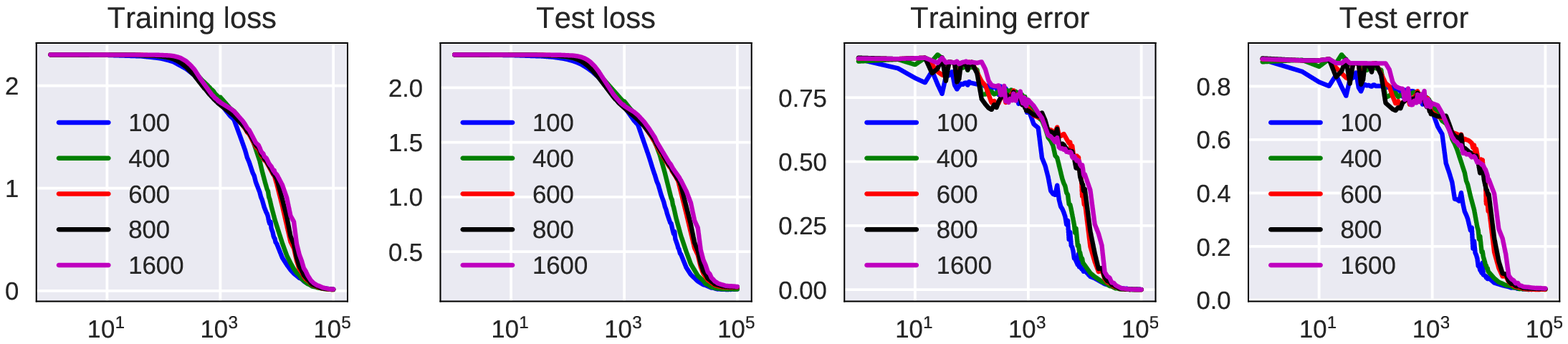}
\par\end{centering}
\begin{centering}
\includegraphics[width=1\columnwidth]{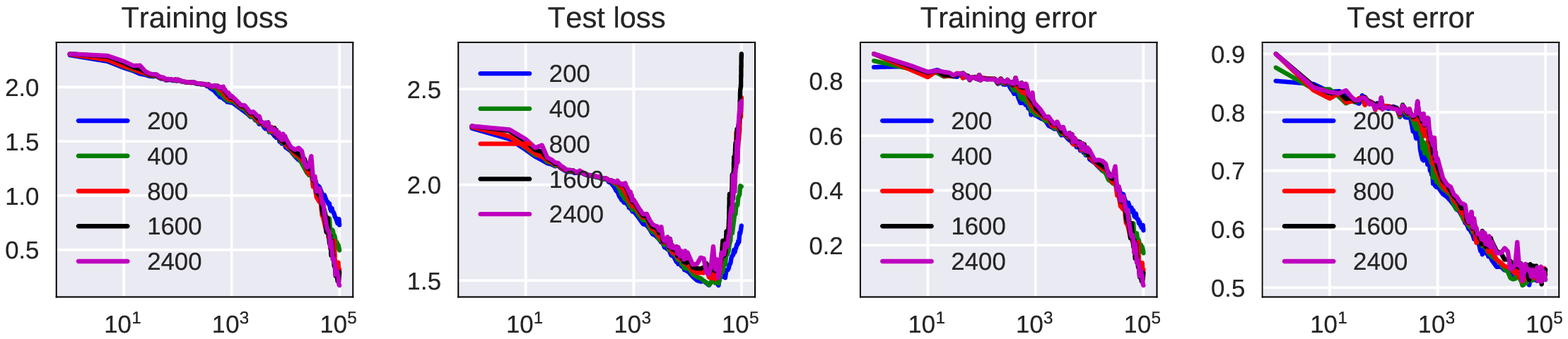}
\par\end{centering}
\caption{The performance of several fully-connected networks, plotted against
training iteration, in each plot. Here the activation $\sigma=\tanh$.
We choose $\tau_{1}=\sqrt{2}$, $\mu_{2}=1$, $\mu_{3}=0$ and $\tau_{3}=0$
in all plots. First row: isotropic Gaussians task, 5 layers, $n=50,100,200,400,800$,
$\tau_{2}=3$, and annealed learning rate $\alpha_{k}=0.003k^{-0.1}$
for $k\protect\geq1$ the SGD iteration. Second row: MNIST task, 4
layers, $n=100,400,600,800,1600$, $\tau_{2}=2$, and $\alpha_{k}=0.01k^{-0.1}$.
Third row: CIFAR-10 task, 4 layers, $n=200,400,800,1600,2400$, $\tau_{2}=2$,
and $\alpha_{k}=0.2k^{-0.1}$.}
\label{fig:tanh_activation}
\end{figure}

\begin{figure}
\begin{centering}
\includegraphics[width=1\columnwidth]{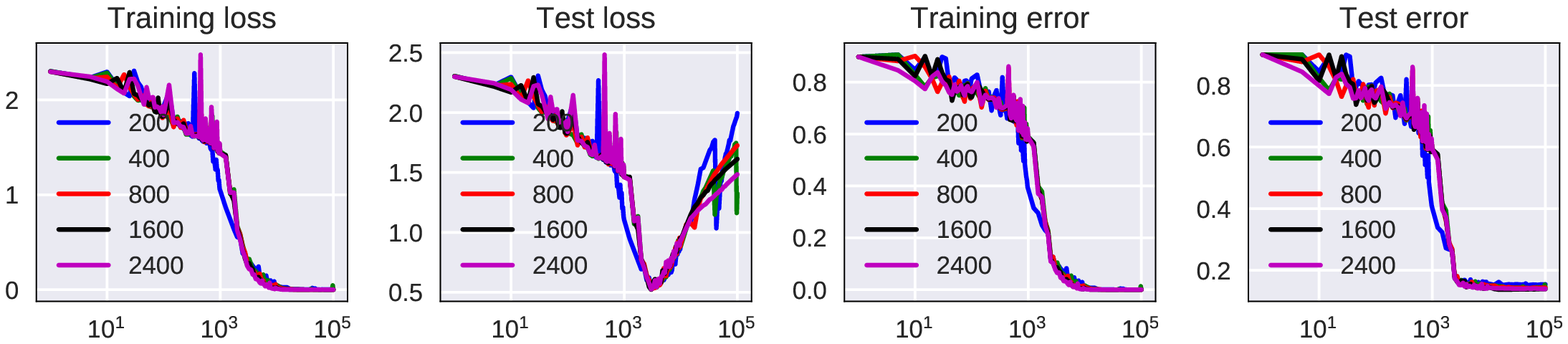}
\par\end{centering}
\caption{The performance of five 4-layers fully-connected networks on CIFAR-10
classification with VGG16 features, plotted against training iteration,
in each plot. Here $n=200,400,800,1600,2400$, $\sigma$ is the ReLU,
$\alpha=0.02$, $\tau_{1}=\sqrt{2}$, $\mu_{2}=1$, $\tau_{2}=0.1$,
$\mu_{3}=0$, $\tau_{3}=0$.}
\label{fig:cifar_vgg16}
\end{figure}

In Fig. \ref{fig:tanh_activation}, we plot the evolution for a different
choice of $\sigma$: the $\tanh$ activation. In Fig. \ref{fig:cifar_vgg16},
we plot the evolution of 4-layers networks for the specific task of
CIFAR-10 classification with VGG16 features. We still observe that
the larger $n$, the better the curves match. The performance in Fig.
\ref{fig:cifar_vgg16} is also reasonable and shows an expectedly
marked improvement over networks trained on raw CIFAR-10 images; for
instance, by the end of training, the network with $n=800$ achieves
a test error rate of about 14\%.

\section*{Acknowledgement}

The work was partially supported by NSF through grants CCF-1714305
and IIS-1741162, and the William R. Hewlett Stanford Graduate Fellowship.
The author would like to thank Andrea Montanari and Huy Tuan Pham
for many helpful and inspiring discussions and their encouragement.

\clearpage{}

\appendix

\section{A heuristic derivation for the mean field limit of multilayer networks\label{sec:multilayer-derivation}}

We first recall the setting, as well as the formalism, in Section
\ref{sec:MF-Multilayer}. We also recall the two properties, as discussed
in Section \ref{subsec:three-layers-derivation}: marginal uniformity
and self-averaging. These properties still hold in the multilayer
case, since symmetry among the neurons of the same layer holds. To
make use of these properties, we represent each neuron with the following
(see also Fig. \ref{fig:Multilayer-net}):
\begin{itemize}
\item At the first layer, neuron $i$ is represented by $\left(\boldsymbol{\theta}_{1,i},\boldsymbol{w}_{1,i}\right)\in\mathbb{R}^{q}\times{\cal F}_{1}$,
where $\boldsymbol{w}_{1,i}$ is the $i$-th row of $\boldsymbol{W}_{1}$
and $\boldsymbol{\theta}_{1,i}$ is the $i$-th element of $\boldsymbol{\Theta}_{1}$.
\item At layer $\ell\geq2$, neuron $j$ is represented by $\left(\boldsymbol{\theta}_{\ell,j},\nu_{\ell,j}\right)$
where $\nu_{\ell,j}\in\mathscr{K}\left(\mathbb{R}^{q}\times{\cal F}_{\ell},\mathbb{R}\right)$
is a stochastic kernel and $\boldsymbol{\theta}_{\ell,j}$ is the
$j$-th element of $\boldsymbol{\Theta}_{\ell}$. Neuron $j$ generates
the weight $w_{\ell,ji}$ (the $\left(j,i\right)$-th entry of $\boldsymbol{W}_{\ell}$)
according to $\nu_{2,j}\left(\cdot\middle|\boldsymbol{\theta}_{1,i},\boldsymbol{w}_{1,i}\right)$
if $\ell=2$ and $\nu_{\ell,j}\left(\cdot\middle|\boldsymbol{\theta}_{\ell-1,i},{\rm CE}\left\{ \nu_{\ell-1,i}\right\} \right)$
otherwise.
\end{itemize}
We are now ready to give a heuristic derivation of the connection
between the three-layers network and the formal system.

\subsubsection*{Forward pass.}

Let us derive Eq. (\ref{eq:multilayer-formal-net}) from Eq. (\ref{eq:multilayer-net}).
We have for large $n_{1}$, at each neuron $j$ of the second layer
for $j\in\left[n_{2}\right]$:
\[
h_{2,j}=\frac{1}{n_{1}}\left\langle \boldsymbol{w}_{2,j},\sigma_{1}\left(\boldsymbol{\Theta}_{1},\boldsymbol{W}_{1}\boldsymbol{x}\right)\right\rangle \approx\int w\sigma_{1}\left(\boldsymbol{\theta},H_{1}\left(\boldsymbol{w};\boldsymbol{x}\right)\right)\nu_{2,j}\left({\rm d}w\middle|\boldsymbol{\theta},\boldsymbol{w}\right)\rho_{1}\left({\rm d}\boldsymbol{\theta},{\rm d}\boldsymbol{w}\right),
\]
which replaces the empirical measure ${\rm Emp}\left(\left\{ \boldsymbol{\theta}_{1,i},\boldsymbol{w}_{1,i},w_{2,ji}\right\} _{i\in\left[n_{1}\right]}\right)$
with $\nu_{2,j}\rho_{1}$, by the self-averaging property. Here we
note that $h_{2,j}$'s for different neuron $j$'s involve the same
$\boldsymbol{\Theta}_{1}$ and $\boldsymbol{W}_{1}$ of the first
layer. This is reflected by the use of $\rho_{1}$ independent of
$j$. By setting
\[
f_{2,j}\left(\boldsymbol{\theta},\boldsymbol{w}\right)={\rm CE}\left\{ \nu_{2,j}\right\} \left(\boldsymbol{\theta},\boldsymbol{w}\right)=\int w\nu_{2,j}\left({\rm d}w\middle|\boldsymbol{\theta},\boldsymbol{w}\right),
\]
we obtain 
\[
h_{2,j}\approx H_{2}\left(f_{2,j};\boldsymbol{x},\rho_{1}\right).
\]
Consequently,
\begin{align*}
h_{3,j} & =\frac{1}{n_{2}}\left\langle \boldsymbol{w}_{3,j},\sigma_{2}\left(\boldsymbol{\Theta}_{2},\boldsymbol{h}_{2}\right)\right\rangle \approx\frac{1}{n_{2}}\sum_{i=1}^{n_{2}}w_{3,ji}\sigma_{2}\left(\boldsymbol{\theta}_{2,i},H_{2}\left(f_{2,i};\boldsymbol{x},\rho_{1}\right)\right).
\end{align*}
The same argument then gives us:
\[
h_{3,j}\approx\int w\sigma_{2}\left(\boldsymbol{\theta},H_{2}\left(f;\boldsymbol{x},\rho_{1}\right)\right)\nu_{3,j}\left({\rm d}w\middle|\boldsymbol{\theta},f\right)\rho_{2}\left({\rm d}\boldsymbol{\theta},{\rm d}f\right)=H_{3}\left(f_{3,j};\boldsymbol{x},\rho_{1},\rho_{2}\right),
\]
in which we set $f_{3,j}={\rm CE}\left\{ \nu_{3,j}\right\} $. Similarly,
performing a chain of the same argument, we thus get, for $\ell=2,...,L$:
\begin{equation}
h_{\ell,j}\approx H_{\ell}\left(f_{\ell,j};\boldsymbol{x},\left\{ \rho_{i}\right\} _{i=1}^{\ell-1}\right),\qquad f_{\ell,j}={\rm CE}\left\{ \nu_{\ell,j}\right\} .\label{eq:multilayer-connection-forward-1}
\end{equation}
This yields
\[
\hat{y}_{n}\left(\boldsymbol{x};{\cal W}\right)=\frac{1}{n_{L}}\sum_{i=1}^{n_{L}}\sigma_{L}\left(\boldsymbol{\theta}_{L,i},h_{L,i}\right)\approx\int\sigma_{L}\left(\boldsymbol{\theta},H_{L}\left(f;\boldsymbol{x},\left\{ \rho_{i}\right\} _{i=1}^{L-1}\right)\right)\rho_{L}\left({\rm d}\boldsymbol{\theta},{\rm d}f\right)=\hat{y}\left(\boldsymbol{x};\underbar{\ensuremath{\rho}}\right).
\]
Note that $\rho_{1}$ is a surrogate measure of ${\rm Emp}\left(\left\{ \boldsymbol{\theta}_{1,i},\boldsymbol{w}_{1,i}\right\} _{i\in\left[n_{1}\right]}\right)$,
and for $\ell=2,...,L$, $\rho_{\ell}$ is that of ${\rm Emp}\left(\left\{ \boldsymbol{\theta}_{\ell,i},{\rm CE}\left\{ \nu_{\ell,i}\right\} \right\} _{i\in\left[n_{\ell}\right]}\right)$.

\subsubsection*{Backward pass.}

We have from Eq. (\ref{eq:multilayer-connection-forward-1}):
\begin{align}
\left(\tilde{\nabla}_{\boldsymbol{\boldsymbol{\Theta}}_{L}}\hat{y}_{n}\left(\boldsymbol{x};{\cal W}\right)\right)_{j} & \approx\nabla_{1}\sigma_{L}\left(\boldsymbol{\theta}_{L,j},H_{L}\left(f_{L,j};\boldsymbol{x},\left\{ \rho_{i}\right\} _{i=1}^{L-1}\right)\right)=\Delta_{\boldsymbol{\theta},L}\left(\boldsymbol{\theta}_{L,j},f_{L,j};\boldsymbol{x},\left\{ \rho_{i}\right\} _{i=1}^{L-1}\right),\label{eq:multilayer-connection-backward-1}\\
\left(\tilde{\nabla}_{\boldsymbol{h}_{L}}\hat{y}_{n}\left(\boldsymbol{x};{\cal W}\right)\right)_{j} & \approx\partial_{2}\sigma_{L}\left(\boldsymbol{\theta}_{L,j},H_{L}\left(f_{L,j};\boldsymbol{x},\left\{ \rho_{i}\right\} _{i=1}^{L-1}\right)\right)=\Delta_{H,L}\left(\boldsymbol{\theta}_{L,j},f_{L,j};\boldsymbol{x},\left\{ \rho_{i}\right\} _{i=1}^{L-1}\right).\nonumber 
\end{align}
Consider $\tilde{\nabla}_{\boldsymbol{\Theta}_{L-1}}\hat{y}_{n}\left(\boldsymbol{x};{\cal W}\right)$:
\begin{align*}
\left(\tilde{\nabla}_{\boldsymbol{\Theta}_{L-1}}\hat{y}_{n}\left(\boldsymbol{x};{\cal W}\right)\right)_{j} & =\left(\frac{1}{n_{L}}\sum_{k=1}^{n_{L}}w_{L,kj}\left(\tilde{\nabla}_{\boldsymbol{h}_{L}}\hat{y}_{n}\left(\boldsymbol{x};{\cal W}\right)\right)_{k}\right)\nabla_{1}\sigma_{L-1}\left(\boldsymbol{\theta}_{L-1,j},h_{L-1,j}\right)\\
 & \approx\left(\frac{1}{n_{L}}\sum_{k=1}^{n_{L}}w_{L,kj}\Delta_{H,L}\left(\boldsymbol{\theta}_{L,k},f_{L,k};\boldsymbol{x},\left\{ \rho_{i}\right\} _{i=1}^{L-1}\right)\right)\\
 & \qquad\times\nabla_{1}\sigma_{L-1}\left(\boldsymbol{\theta}_{L-1,j},H_{L-1}\left(f_{L-1,j};\boldsymbol{x},\left\{ \rho_{i}\right\} _{i=1}^{L-2}\right)\right).
\end{align*}
Recall that $w_{L,kj}\sim\nu_{L,k}\left(\cdot\middle|\boldsymbol{\theta}_{L-1,j},f_{L-1,j}\right)$,
$f_{L,k}={\rm CE}\left\{ \nu_{L,k}\right\} $, and that neuron $k$
of layer $L$ is represented by $\left(\boldsymbol{\theta}_{L,k},\nu_{L,k}\right)$.
By the self-averaging property:
\begin{align*}
\left(\tilde{\nabla}_{\boldsymbol{\Theta}_{L-1}}\hat{y}_{n}\left(\boldsymbol{x};{\cal W}\right)\right)_{j} & \approx\left(\int f\left(\boldsymbol{\theta}_{L-1,j},f_{L-1,j}\right)\Delta_{H,L}\left(\boldsymbol{\theta},f;\boldsymbol{x},\left\{ \rho_{i}\right\} _{i=1}^{L-1}\right)\rho_{L}\left({\rm d}\boldsymbol{\theta},{\rm d}f\right)\right)\\
 & \qquad\times\nabla_{1}\sigma_{L-1}\left(\boldsymbol{\theta}_{L-1,j},H_{L-1}\left(f_{L-1,j};\boldsymbol{x},\left\{ \rho_{i}\right\} _{i=1}^{L-2}\right)\right)\\
 & =\Delta_{\boldsymbol{\theta},L-1}\left(\boldsymbol{\theta}_{L-1,j},f_{L-1,j};\boldsymbol{x},\underbar{\ensuremath{\rho}}\right),
\end{align*}
and similarly,
\[
\left(\tilde{\nabla}_{\boldsymbol{h}_{L-1}}\hat{y}_{n}\left(\boldsymbol{x};{\cal W}\right)\right)_{j}\approx\Delta_{H,L-1}\left(\boldsymbol{\theta}_{L-1,j},f_{L-1,j};\boldsymbol{x},\underbar{\ensuremath{\rho}}\right).
\]
Performing the same argument and recalling that $f_{\ell,j}={\rm CE}\left\{ \nu_{\ell,j}\right\} $,
we obtain that
\begin{align}
\left(\tilde{\nabla}_{\boldsymbol{\Theta}_{\ell}}\hat{y}_{n}\left(\boldsymbol{x};{\cal W}\right)\right)_{j} & \approx\Delta_{\boldsymbol{\theta},\ell}\left(\boldsymbol{\theta}_{\ell,j},f_{\ell,j};\boldsymbol{x},\underbar{\ensuremath{\rho}}\right),\qquad\ell=2,...,L-1,\label{eq:multilayer-connection-backward-2}\\
\left(\tilde{\nabla}_{\boldsymbol{h}_{\ell}}\hat{y}_{n}\left(\boldsymbol{x};{\cal W}\right)\right)_{j} & \approx\Delta_{H,\ell}\left(\boldsymbol{\theta}_{\ell,j},f_{\ell,j};\boldsymbol{x},\underbar{\ensuremath{\rho}}\right),\qquad\ell=2,...,L-1,\nonumber \\
\left(\tilde{\nabla}_{\boldsymbol{\Theta}_{1}}\hat{y}_{n}\left(\boldsymbol{x};{\cal W}\right)\right)_{j} & \approx\Delta_{\boldsymbol{\theta},1}\left(\boldsymbol{\theta}_{1,j},\boldsymbol{w}_{1,j};\boldsymbol{x},\underbar{\ensuremath{\rho}}\right),\label{eq:multilayer-connection-backward-3}\\
\left(\tilde{\nabla}_{\boldsymbol{h}_{1}}\hat{y}_{n}\left(\boldsymbol{x};{\cal W}\right)\right)_{j} & \approx\Delta_{H,1}\left(\boldsymbol{\theta}_{1,j},\boldsymbol{w}_{1,j};\boldsymbol{x},\underbar{\ensuremath{\rho}}\right).\nonumber 
\end{align}
Finally, it is then easy to see that
\begin{align}
\left(\tilde{\nabla}_{\boldsymbol{W}_{L}}\hat{y}_{n}\left(\boldsymbol{x};{\cal W}\right)\right)_{ij} & \approx\Delta_{W,L}\left(\boldsymbol{\theta}_{L,i},f_{L,i},\boldsymbol{\theta}_{L-1,i},f_{L-1,i};\boldsymbol{x},\left\{ \rho_{i}\right\} _{i=1}^{L-1}\right),\label{eq:multilayer-connection-backward-4}\\
\left(\tilde{\nabla}_{\boldsymbol{W}_{\ell}}\hat{y}_{n}\left(\boldsymbol{x};{\cal W}\right)\right)_{ij} & \approx\Delta_{W,\ell}\left(\boldsymbol{\theta}_{\ell,i},f_{\ell,i},\boldsymbol{\theta}_{\ell-1,j},f_{\ell-1,j};\boldsymbol{x},\underbar{\ensuremath{\rho}}\right),\qquad\ell=3,...,L-1,\label{eq:multilayer-connection-backward-5}\\
\left(\tilde{\nabla}_{\boldsymbol{W}_{2}}\hat{y}_{n}\left(\boldsymbol{x};{\cal W}\right)\right)_{ij} & \approx\Delta_{W,2}\left(\boldsymbol{\theta}_{2,i},f_{2,i},\boldsymbol{\theta}_{1,j},\boldsymbol{w}_{1,j};\boldsymbol{x},\underbar{\ensuremath{\rho}}\right),\label{eq:multilayer-connection-backward-6}\\
\left(\tilde{\nabla}_{\boldsymbol{W}_{1}}\hat{y}_{n}\left(\boldsymbol{x};{\cal W}\right)\right)_{i} & \approx\Delta_{W,1}\left(\boldsymbol{\theta}_{1,i},\boldsymbol{w}_{1,i};\boldsymbol{x},\underbar{\ensuremath{\rho}}\right),\label{eq:multilayer-connection-backward-7}
\end{align}
where $\left(\tilde{\nabla}_{\boldsymbol{W}_{1}}\hat{y}_{n}\left(\boldsymbol{x};{\cal W}\right)\right)_{i}$
is the $i$-th row of $\tilde{\nabla}_{\boldsymbol{W}_{1}}\hat{y}_{n}\left(\boldsymbol{x};{\cal W}\right)$.

\subsubsection*{Learning dynamics.}

We now connect the evolution dynamics of the formal system with the
SGD dynamics of the neural network. Similar to the three-layers case
in Section \ref{subsec:three-layers-derivation}, we first take $t=k\alpha$
and $\alpha\downarrow0$ to obtain time continuum and in-expectation
property w.r.t. ${\cal P}$ from the SGD dynamics:
\begin{align*}
\frac{{\rm d}}{{\rm d}t}\boldsymbol{W}_{\ell}^{t} & =-\mathbb{E}_{{\cal P}}\left\{ \partial_{2}{\cal L}\left(y,\hat{y}_{n}\left(\boldsymbol{x};{\cal W}^{t}\right)\right)\tilde{\nabla}_{\boldsymbol{W}_{\ell}}\hat{y}_{n}\left(\boldsymbol{x};{\cal W}^{t}\right)\right\} ,\\
\frac{{\rm d}}{{\rm d}t}\boldsymbol{\Theta}_{\ell}^{t} & =-\mathbb{E}_{{\cal P}}\left\{ \partial_{2}{\cal L}\left(y,\hat{y}_{n}\left(\boldsymbol{x};{\cal W}^{t}\right)\right)\tilde{\nabla}_{\boldsymbol{\Theta}_{\ell}}\hat{y}_{n}\left(\boldsymbol{x};{\cal W}^{t}\right)\right\} ,\qquad\ell=1,...,L,
\end{align*}
where ${\cal W}^{t}=\left\{ \boldsymbol{W}_{1}^{t},...,\boldsymbol{W}_{L}^{t},\boldsymbol{\Theta}_{1}^{t},...,\boldsymbol{\Theta}_{L}^{t}\right\} $.
Given this, at any time $t$, we represent neuron $i$ of the first
layer and neuron $j$ of layer $\ell\geq2$ with respectively $\left(\boldsymbol{\theta}_{1,i}^{t},\boldsymbol{w}_{1,i}^{t}\right)$
and $\left(\boldsymbol{\theta}_{\ell,j}^{t},\nu_{\ell,j}^{t}\right)$,
whose meanings accord with the representation described at the beginning
of Appendix \ref{sec:multilayer-derivation}. We also define $f_{\ell,j}^{t}={\rm CE}\left\{ \nu_{\ell,j}^{t}\right\} $
for $j\in\left[n_{\ell}\right]$ and $\ell=2,...,L$. We let $\tilde{\rho}_{1}^{t}$
be the surrogate measure of ${\rm Emp}\left(\left\{ \boldsymbol{\theta}_{1,i}^{t},\boldsymbol{w}_{1,i}^{t}\right\} _{i\in\left[n_{1}\right]}\right)$
and $\tilde{\rho}_{\ell}^{t}$ be that of ${\rm Emp}\left(\left\{ \boldsymbol{\theta}_{\ell,i}^{t},{\rm CE}\left\{ \nu_{\ell,i}^{t}\right\} \right\} _{i\in\left[n_{\ell}\right]}\right)$
for $\ell=2,...,L$. Let $\underbar{\ensuremath{\tilde{\rho}}}^{t}=\left\{ \tilde{\rho}_{\ell}^{t}\right\} _{\ell=1}^{L}$.

It is easy to see from Eq. (\ref{eq:multilayer-connection-backward-1})-(\ref{eq:multilayer-connection-backward-7})
that
\begin{align}
\frac{{\rm d}}{{\rm d}t}\boldsymbol{w}_{1,j}^{t} & \approx G_{W,1}\left(\boldsymbol{\theta}_{1,j}^{t},\boldsymbol{w}_{1,j}^{t};\tilde{\underbar{\ensuremath{\rho}}}^{t}\right),\label{eq:multilayer-connection-dynamics-1}\\
\frac{{\rm d}}{{\rm d}t}w_{2,ij}^{t} & \approx G_{f,2}\left(\boldsymbol{\theta}_{2,i}^{t},f_{2,i}^{t},\boldsymbol{\theta}_{1,j}^{t},\boldsymbol{w}_{1,j}^{t};\tilde{\underbar{\ensuremath{\rho}}}^{t}\right),\label{eq:multilayer-connection-dynamics-2}\\
\frac{{\rm d}}{{\rm d}t}w_{\ell,ij}^{t} & \approx G_{f,\ell}\left(\boldsymbol{\theta}_{\ell,i}^{t},f_{\ell,i}^{t},\boldsymbol{\theta}_{\ell-1,j}^{t},f_{\ell-1,j}^{t};\tilde{\underbar{\ensuremath{\rho}}}^{t}\right),\qquad\ell=3,...,L,\label{eq:multilayer-connection-dynamics-3}\\
\frac{{\rm d}}{{\rm d}t}\boldsymbol{\theta}_{1,j}^{t} & \approx G_{\boldsymbol{\theta},1}\left(\boldsymbol{\theta}_{1,j}^{t},\boldsymbol{w}_{1,j}^{t};\tilde{\underbar{\ensuremath{\rho}}}^{t}\right),\label{eq:multilayer-connection-dynamics-4}\\
\frac{{\rm d}}{{\rm d}t}\boldsymbol{\theta}_{\ell,j}^{t} & \approx G_{\boldsymbol{\theta},\ell}\left(\boldsymbol{\theta}_{\ell,j}^{t},f_{\ell,j}^{t};\tilde{\underbar{\ensuremath{\rho}}}^{t}\right),\qquad\ell=2,...,L.\label{eq:multilayer-connection-dynamics-5}
\end{align}
Applying the marginal uniformity property to Eq. (\ref{eq:multilayer-connection-dynamics-1})
and Eq. (\ref{eq:multilayer-connection-dynamics-4}), we have ${\rm Law}\left(\boldsymbol{\theta}_{1,j}^{t},\boldsymbol{w}_{1,j}^{t}\right)$
is independent of $j$, and thanks to the self-averaging property
in addition, we obtain ${\rm Law}\left(\boldsymbol{\theta}_{1,j}^{t},\boldsymbol{w}_{1,j}^{t}\right)\approx\tilde{\rho}_{1}^{t}$. 

Observe that in Eq. (\ref{eq:multilayer-connection-dynamics-2}),
the right-hand side does not depend on $w_{2,ij}^{t}$, and since
$w_{2,ij}^{t}\sim\nu_{2,i}^{t}\left(\cdot\middle|\boldsymbol{\theta}_{1,j}^{t},\boldsymbol{w}_{1,j}^{t}\right)$,
we get for $\Delta t\to0$ and any event $E\subseteq\mathbb{R}$,
\[
\nu_{2,i}^{t+\Delta t}\left(E+G_{f,2}\left(\boldsymbol{\theta}_{2,i}^{t},f_{2,i}^{t},\boldsymbol{\theta}_{1,j}^{t},\boldsymbol{w}_{1,j}^{t};\tilde{\underbar{\ensuremath{\rho}}}^{t}\right)\Delta t\middle|\boldsymbol{\theta}_{1,j}^{t+\Delta t},\boldsymbol{w}_{1,j}^{t+\Delta t}\right)\approx\nu_{2,i}^{t}\left(E\middle|\boldsymbol{\theta}_{1,j}^{t},\boldsymbol{w}_{1,j}^{t}\right).
\]
This gives us
\begin{align*}
\frac{{\rm d}}{{\rm d}t}\left(f_{2,i}^{t}\left(\boldsymbol{\theta}_{1,j}^{t},\boldsymbol{w}_{1,j}^{t}\right)\right) & =\lim_{\Delta t\to0}\frac{1}{\Delta t}\left(\int w\nu_{2,i}^{t+\Delta t}\left({\rm d}w\middle|\boldsymbol{\theta}_{1,j}^{t+\Delta t},\boldsymbol{w}_{1,j}^{t+\Delta t}\right)-\int w\nu_{2,i}^{t}\left({\rm d}w\middle|\boldsymbol{\theta}_{1,j}^{t},\boldsymbol{w}_{1,j}^{t}\right)\right)\\
 & \approx\lim_{\Delta t\to0}\frac{1}{\Delta t}\Biggl(\int\left(w+G_{f,2}\left(\boldsymbol{\theta}_{2,i}^{t},f_{2,i}^{t},\boldsymbol{\theta}_{1,j}^{t},\boldsymbol{w}_{1,j}^{t};\tilde{\underbar{\ensuremath{\rho}}}^{t}\right)\Delta t\right)\nu_{2,i}^{t}\left({\rm d}w\middle|\boldsymbol{\theta}_{1,j}^{t},\boldsymbol{w}_{1,j}^{t}\right)\\
 & \qquad\qquad\qquad-\int w\nu_{2,i}^{t}\left({\rm d}w\middle|\boldsymbol{\theta}_{1,j}^{t},\boldsymbol{w}_{1,j}^{t}\right)\Biggl)\\
 & =G_{f,2}\left(\boldsymbol{\theta}_{2,i}^{t},f_{2,i}^{t},\boldsymbol{\theta}_{1,j}^{t},\boldsymbol{w}_{1,j}^{t};\tilde{\underbar{\ensuremath{\rho}}}^{t}\right).
\end{align*}
On the other hand, from Eq. (\ref{eq:multilayer-connection-dynamics-1})
and (\ref{eq:multilayer-connection-dynamics-4}),
\begin{align*}
\frac{{\rm d}}{{\rm d}t}\left(f_{2,i}^{t}\left(\boldsymbol{\theta}_{1,j}^{t},\boldsymbol{w}_{1,j}^{t}\right)\right) & =\left(\partial_{t}f_{2,i}^{t}\right)\left(\boldsymbol{\theta}_{1,j}^{t},\boldsymbol{w}_{1,j}^{t}\right)+\left\langle \nabla_{1}f_{2,i}^{t}\left(\boldsymbol{\theta}_{1,j}^{t},\boldsymbol{w}_{1,j}^{t}\right),G_{\boldsymbol{\theta},1}\left(\boldsymbol{\theta}_{1,j}^{t},\boldsymbol{w}_{1,j}^{t};\tilde{\underbar{\ensuremath{\rho}}}^{t}\right)\right\rangle \\
 & \qquad+\left\langle \nabla_{2}f_{2,i}^{t}\left(\boldsymbol{\theta}_{1,j}^{t},\boldsymbol{w}_{1,j}^{t}\right),G_{W,1}\left(\boldsymbol{\theta}_{1,j}^{t},\boldsymbol{w}_{1,j}^{t};\tilde{\underbar{\ensuremath{\rho}}}^{t}\right)\right\rangle .
\end{align*}
We thus get:
\begin{equation}
\partial_{t}f_{2,i}^{t}\left(\boldsymbol{\theta},\boldsymbol{w}\right)\approx{\cal G}_{2}\left(\boldsymbol{\theta}_{2,i}^{t},f_{2,i}^{t},\boldsymbol{\theta},\boldsymbol{w};\tilde{\underbar{\ensuremath{\rho}}}^{t}\right),\label{eq:multilayer-connection-dynamics-6}
\end{equation}
for $\tilde{\rho}_{1}^{t}$-a.e. $\left(\boldsymbol{\theta},\boldsymbol{w}\right)$,
recalling ${\rm Law}\left(\boldsymbol{\theta}_{1,j}^{t},\boldsymbol{w}_{1,j}^{t}\right)\approx\tilde{\rho}_{1}^{t}$
for any $j\in\left[n_{1}\right]$. Observe that values of $f_{2,i}^{t}$
at $\left(\boldsymbol{\theta},\boldsymbol{w}\right)\notin{\rm supp}\left(\tilde{\rho}_{1}^{t}\right)$
are used in the computation of neither the forward pass nor the backward
pass at time $t$. As such, one can extend the dynamic (\ref{eq:multilayer-connection-dynamics-6})
to all $\left(\boldsymbol{\theta},\boldsymbol{w}\right)\in\mathbb{R}^{q}\times{\cal F}_{1}$
without affecting the prediction stated in Section \ref{subsec:multilayer-formalism}.
Applying the marginal uniformity and self-averaging properties again,
we then obtain ${\rm Law}\left(\boldsymbol{\theta}_{2,i}^{t},f_{2,i}^{t}\right)\approx\tilde{\rho}_{2}^{t}$
for any $i\in\left[n_{2}\right]$. One can then perform a similar
argument on Eq. (\ref{eq:multilayer-connection-dynamics-3}) inductively
for $\ell=3,...,L$ and get:
\begin{align*}
\frac{{\rm d}}{{\rm d}t}\left(f_{\ell,i}^{t}\left(\boldsymbol{\theta}_{\ell-1,j}^{t},f_{\ell-1,j}^{t}\right)\right) & \approx G_{f,\ell}\left(\boldsymbol{\theta}_{\ell,i}^{t},f_{\ell,i}^{t},\boldsymbol{\theta}_{\ell-1,j}^{t},f_{\ell-1,j}^{t};\tilde{\underbar{\ensuremath{\rho}}}^{t}\right),\\
\frac{{\rm d}}{{\rm d}t}\left(f_{\ell,i}^{t}\left(\boldsymbol{\theta}_{\ell-1,j}^{t},f_{\ell-1,j}^{t}\right)\right) & =\left(\partial_{t}f_{\ell,i}^{t}\right)\left(\boldsymbol{\theta}_{\ell-1,j}^{t},f_{\ell-1,j}^{t}\right)+\left\langle \nabla_{1}f_{\ell,i}^{t}\left(\boldsymbol{\theta}_{\ell-1,j}^{t},f_{\ell-1,j}^{t}\right),G_{\boldsymbol{\theta},\ell}\left(\boldsymbol{\theta}_{\ell-1,j}^{t},f_{\ell-1,j}^{t};\tilde{\underbar{\ensuremath{\rho}}}^{t}\right)\right\rangle \\
 & \qquad+\mathscr{D}_{2}f_{\ell,i}^{t}\left\{ \boldsymbol{\theta}_{\ell-1,j}^{t},f_{\ell-1,j}^{t}\right\} \left(\partial_{t}f_{\ell-1,j}^{t}\right),
\end{align*}
which yields
\[
\partial_{t}f_{\ell,i}^{t}\left(\boldsymbol{\theta},f\right)\approx{\cal G}_{\ell}\left(\boldsymbol{\theta}_{\ell,i}^{t},f_{\ell,i}^{t},\boldsymbol{\theta},f;\underbar{\ensuremath{\rho}}^{t}\right)\qquad\forall\left(\boldsymbol{\theta},f\right)\in\mathbb{R}^{q}\times{\cal F}_{\ell-1},\qquad{\rm Law}\left(\boldsymbol{\theta}_{\ell,i}^{t},f_{\ell,i}^{t}\right)\approx\tilde{\rho}_{\ell}^{t},
\]
for all $i\in\left[n_{\ell}\right]$. Hence, if $\tilde{\rho}_{\ell}^{0}=\rho_{\ell}^{0}$
for all $\ell\in\left[L\right]$, then $\tilde{\rho}_{\ell}^{t}\approx\rho_{\ell}^{t}$
for all $\ell\in\left[L\right]$ at any time $t$. This completes
the derivation.

Finally we note that the initialization in the prediction statement
in Section \ref{subsec:multilayer-formalism} is sufficient to ensure
firstly that symmetry among the neurons is attained at initialization
and hence at all subsequent time, and secondly $\tilde{\rho}_{\ell}^{0}=\rho_{\ell}^{0}$
for all $\ell\in\left[L\right]$.

\section{Statics: equivalence of the optima for multilayer networks\label{sec:multilayer-statics}}

We recall the multilayer network (\ref{eq:multilayer-net}) and its
formal system (\ref{eq:multilayer-formal-net}). Also recall that
$n_{\ell}=n_{\ell}\left(n\right)\to\infty$ as $n\to\infty$. In the
following, we argue that
\[
\lim_{n\to\infty}\inf_{{\cal W}}\mathbb{E}_{{\cal P}}\left\{ {\cal L}\left(y,\hat{y}_{n}\left(\boldsymbol{x};{\cal W}\right)\right)\right\} =\inf_{\underbar{\ensuremath{\rho}}}\mathbb{E}_{{\cal P}}\left\{ {\cal L}\left(y,\hat{y}\left(\boldsymbol{x};\underbar{\ensuremath{\rho}}\right)\right)\right\} ,
\]
under certain assumptions to be stated below. This result is similar
to, though not as quantitative as, Eq. (\ref{eq:two-layers-statics})
of the two-layers case. To show the above, it is decomposed into the
two inequalities:
\begin{align}
\inf_{\underbar{\ensuremath{\rho}}}\mathbb{E}_{{\cal P}}\left\{ {\cal L}\left(y,\hat{y}\left(\boldsymbol{x};\underbar{\ensuremath{\rho}}\right)\right)\right\}  & \leq\inf_{{\cal W}}\mathbb{E}_{{\cal P}}\left\{ {\cal L}\left(y,\hat{y}_{n}\left(\boldsymbol{x};{\cal W}\right)\right)\right\} ,\label{eq:multilayer-statics-1}\\
\limsup_{n\to\infty}\inf_{{\cal W}}\mathbb{E}_{{\cal P}}\left\{ {\cal L}\left(y,\hat{y}_{n}\left(\boldsymbol{x};{\cal W}\right)\right)\right\}  & \leq\inf_{\underbar{\ensuremath{\rho}}}\mathbb{E}_{{\cal P}}\left\{ {\cal L}\left(y,\hat{y}\left(\boldsymbol{x};\underbar{\ensuremath{\rho}}\right)\right)\right\} .\label{eq:multilayer-statics-2}
\end{align}
Then the thesis follows immediately.

Below we shall let $C$ denote any (immaterial) constant, and similarly,
$c=c\left(\epsilon\right)$ denote any ``constant'' that depends
on $\epsilon$ \textendash{} a parameter that is to be defined below.
That is, $C$ and $c\left(\epsilon\right)$ are constants that may
change from line to line. We state our assumptions:
\begin{itemize}
\item $\sigma_{L}$ satisfies that for some constant $C>0$:
\begin{align}
\left|\sigma_{L}\left(\boldsymbol{\theta}_{1},h_{1}\right)-\sigma_{L}\left(\boldsymbol{\theta}_{2},h_{2}\right)\right| & \leq C\left(1+\left\Vert \boldsymbol{\theta}_{1}\right\Vert _{2}+\left\Vert \boldsymbol{\theta}_{2}\right\Vert _{2}\right)\left(\left\Vert \boldsymbol{\theta}_{1}-\boldsymbol{\theta}_{2}\right\Vert _{2}+\left|h_{1}-h_{2}\right|\right),\label{eq:multilayer-statics-assum_sL-1}\\
\left|\sigma_{L}\left(\boldsymbol{\theta},h\right)\right| & \leq C\left(1+\left\Vert \boldsymbol{\theta}\right\Vert _{2}\right),\label{eq:multilayer-statics-assum_sL-2}
\end{align}
for any $\boldsymbol{\theta},\boldsymbol{\theta}_{1},\boldsymbol{\theta}_{2}\in\mathbb{R}^{q}$
and $h,h_{1},h_{2}\in\mathbb{R}$.
\item For each $\ell=1,...,L-1$, $\sigma_{\ell}$ satisfies that for some
constant $C>0$:
\begin{align}
\left|\sigma_{\ell}\left(\boldsymbol{\theta}_{1},h_{1}\right)-\sigma_{\ell}\left(\boldsymbol{\theta}_{2},h_{2}\right)\right| & \leq C\left(\left\Vert \boldsymbol{\theta}_{1}-\boldsymbol{\theta}_{2}\right\Vert _{2}+\left|h_{1}-h_{2}\right|\right),\label{eq:multilayer-statics-assum_s_ell-1}\\
\left\Vert \sigma_{\ell}\right\Vert _{\infty} & \leq C,\label{eq:multilayer-statics-assum_s_ell_2}
\end{align}
for any $\boldsymbol{\theta}_{1},\boldsymbol{\theta}_{2}\in\mathbb{R}^{q}$
and $h_{1},h_{2}\in\mathbb{R}$.
\item ${\cal L}$ satisfies that for some constant $C>0$:
\begin{equation}
\left|{\cal L}\left(y_{1},y_{2}\right)-{\cal L}\left(y_{3},y_{4}\right)\right|\leq C\left(1+\left|y_{1}\right|+\left|y_{2}\right|+\left|y_{3}\right|+\left|y_{4}\right|\right)\left(\left|y_{1}-y_{3}\right|+\left|y_{2}-y_{4}\right|\right),\label{eq:multilayer-statics-assum_L}
\end{equation}
for any $y_{1},y_{2},y_{3},y_{4}\in\mathbb{R}$.
\item There exists a constant $C$ such that the data satisfies:
\begin{equation}
{\cal P}\left(\left|y\right|>C\right)=0.\label{eq:multilayer-statics-assum_y}
\end{equation}
\item For all $\epsilon>0$ sufficiently small, there exists $\underbar{\ensuremath{\rho}}=\left\{ \rho_{\ell}\right\} _{\ell=1}^{L}$
such that 
\begin{equation}
\mathbb{E}_{{\cal P}}\left\{ {\cal L}\left(y,\hat{y}\left(\boldsymbol{x};\underbar{\ensuremath{\rho}}\right)\right)\right\} \leq\inf_{\underbar{\ensuremath{\rho}}'}\mathbb{E}_{{\cal P}}\left\{ {\cal L}\left(y,\hat{y}\left(\boldsymbol{x};\underbar{\ensuremath{\rho}}'\right)\right)\right\} +\epsilon,\label{eq:multilayer-statics-assum_inf-1}
\end{equation}
as well as that
\begin{equation}
\rho_{L}\left(\left\{ \left\Vert \boldsymbol{\theta}\right\Vert _{2}>c\right\} \right)=0,\qquad\rho_{\ell}\left(\left\{ f\in{\cal F}_{\ell}:\;\left\Vert f\right\Vert _{\infty}>c\right\} \right)=0,\quad\ell=2,...,L,\label{eq:multilayer-statics-assum_inf-2}
\end{equation}
for some $c=c\left(\epsilon\right)$.
\end{itemize}

\subsection{Derivation of Eq. (\ref{eq:multilayer-statics-1})}

For $\epsilon>0$, we take ${\cal W}=\left\{ \boldsymbol{W}_{1},...,\boldsymbol{W}_{L},\boldsymbol{\Theta}_{1},...,\boldsymbol{\Theta}_{L}\right\} $
that yields 
\[
\mathbb{E}_{{\cal P}}\left\{ {\cal L}\left(y,\hat{y}_{n}\left(\boldsymbol{x};{\cal W}\right)\right)\right\} \leq\inf_{{\cal W}'}\mathbb{E}_{{\cal P}}\left\{ {\cal L}\left(y,\hat{y}_{n}\left(\boldsymbol{x};{\cal W}'\right)\right)\right\} +\epsilon.
\]
Let $\boldsymbol{\Theta}_{\ell}=\left(\boldsymbol{\theta}_{\ell,i}\right)_{i\in\left[n_{\ell}\right]}$,
$\boldsymbol{W}_{1}=\left(\boldsymbol{w}_{1,i}\right)_{i\in\left[n_{1}\right]}$
and $\boldsymbol{W}_{\ell}=\left(w_{\ell,ij}\right)_{i\in\left[n_{\ell}\right],j\in\left[n_{\ell-1}\right]}$.
We construct $\underbar{\ensuremath{\rho}}=\left\{ \rho_{\ell}\right\} _{\ell=1}^{L}$
as follows. Take
\[
\rho_{1}=\frac{1}{n_{1}}\sum_{i=1}^{n_{1}}\delta_{\boldsymbol{\theta}_{1,i},\boldsymbol{w}_{1,i}}.
\]
We choose (any) $f_{2,i}\in{\cal F}_{2}$, for each $i\in\left[n_{2}\right]$,
such that for any $j\in\left[n_{1}\right]$,
\[
f_{2,i}\left(\boldsymbol{\theta}_{1,j},\boldsymbol{w}_{1,j}\right)=\frac{1}{\left|S_{1,j}\right|}\sum_{k\in S_{1,j}}w_{2,ik},\qquad S_{1,j}=\left\{ k\in\left[n_{1}\right]:\;\left(\boldsymbol{\theta}_{1,k},\boldsymbol{w}_{1,k}\right)=\left(\boldsymbol{\theta}_{1,j},\boldsymbol{w}_{1,j}\right)\right\} .
\]
We then take
\[
\rho_{2}=\frac{1}{n_{2}}\sum_{i=1}^{n_{2}}\delta_{\boldsymbol{\theta}_{2,i},f_{2,i}}.
\]
We continue this process inductively, i.e. for $\ell\geq3$, we choose
$f_{\ell,i}\in{\cal F}_{\ell}$, for each $i\in\left[n_{\ell}\right]$,
such that for any $j\in\left[n_{\ell-1}\right]$,
\[
f_{\ell,i}\left(\boldsymbol{\theta}_{\ell-1,j},f_{\ell-1,j}\right)=\frac{1}{\left|S_{\ell-1,j}\right|}\sum_{k\in S_{\ell-1,j}}w_{\ell,ik},\qquad S_{\ell-1,j}=\left\{ k\in\left[n_{\ell-1}\right]:\;\left(\boldsymbol{\theta}_{\ell-1,k},f_{\ell-1,k}\right)=\left(\boldsymbol{\theta}_{\ell-1,j},f_{\ell-1,j}\right)\right\} ,
\]
and we take
\[
\rho_{\ell}=\frac{1}{n_{\ell}}\sum_{i=1}^{n_{\ell}}\delta_{\boldsymbol{\theta}_{\ell,i},f_{\ell,i}}.
\]
Then it is easy to check that
\[
\mathbb{E}_{{\cal P}}\left\{ {\cal L}\left(y,\hat{y}_{n}\left(\boldsymbol{x};{\cal W}\right)\right)\right\} =\mathbb{E}_{{\cal P}}\left\{ {\cal L}\left(y,\hat{y}\left(\boldsymbol{x};\underbar{\ensuremath{\rho}}\right)\right)\right\} .
\]
This yields Eq. (\ref{eq:multilayer-statics-1}), since $\epsilon$
is arbitrary.

\subsection{Derivation of Eq. (\ref{eq:multilayer-statics-2})}

For $\epsilon>0$ sufficiently small, we take $\underbar{\ensuremath{\rho}}=\left\{ \rho_{\ell}\right\} _{\ell=1}^{L}$
that satisfies Assumptions (\ref{eq:multilayer-statics-assum_inf-1})
and (\ref{eq:multilayer-statics-assum_inf-2}). Let us generate ${\cal W}=\left\{ \boldsymbol{W}_{1},...,\boldsymbol{W}_{L},\boldsymbol{\Theta}_{1},...,\boldsymbol{\Theta}_{L}\right\} $
at random as follows. First we generate $\left\{ \left(\boldsymbol{\theta}_{1,i},\boldsymbol{w}_{1,i}\right)\right\} _{i\in\left[n_{1}\right]}\sim\rho_{1}$
i.i.d. and form $\boldsymbol{W}_{1}=\left(\boldsymbol{w}_{1,i}\right)_{i\in\left[n_{1}\right]}$
and $\boldsymbol{\Theta}_{1}=\left(\boldsymbol{\theta}_{1,i}\right)_{i\in\left[n_{1}\right]}$.
Then inductively for $\ell=2,...,L$, we generate $\left\{ \left(\boldsymbol{\theta}_{\ell,i},f_{\ell,i}\right)\right\} _{i\in\left[n_{\ell}\right]}\sim\rho_{\ell}$
i.i.d., all independently of each other and of $\left\{ \left(\boldsymbol{\theta}_{1,i},\boldsymbol{w}_{1,i}\right)\right\} _{i\in\left[n_{1}\right]}$,
and form $\boldsymbol{W}_{2}=\left(f_{2,i}\left(\boldsymbol{\theta}_{1,j},\boldsymbol{w}_{1,j}\right)\right)_{i\in\left[n_{2}\right],j\in\left[n_{1}\right]}$,
$\boldsymbol{W}_{\ell}=\left(f_{\ell,i}\left(\boldsymbol{\theta}_{\ell-1,j},f_{\ell-1,j}\right)\right)_{i\in\left[n_{\ell}\right],j\in\left[n_{\ell-1}\right]}$
and $\boldsymbol{\Theta}_{\ell}=\left(\boldsymbol{\theta}_{\ell,i}\right)_{i\in\left[n_{\ell}\right]}$.
We shall argue that
\begin{equation}
\mathbb{E}_{{\cal W}}\left\{ \mathbb{E}_{{\cal P}}\left\{ {\cal L}\left(y,\hat{y}_{n}\left(\boldsymbol{x};{\cal W}\right)\right)\right\} \right\} \leq\mathbb{E}_{{\cal P}}\left\{ {\cal L}\left(y,\hat{y}\left(\boldsymbol{x};\underbar{\ensuremath{\rho}}\right)\right)\right\} +c\left(\epsilon\right)\sum_{\ell=1}^{L}\frac{1}{\sqrt{n_{\ell}}}.\label{eq:multilayer-statics-derive-2-1}
\end{equation}
This implies
\[
\inf_{{\cal W}}\mathbb{E}_{{\cal P}}\left\{ {\cal L}\left(y,\hat{y}_{n}\left(\boldsymbol{x};{\cal W}\right)\right)\right\} \leq\inf_{\underbar{\ensuremath{\rho}}}\mathbb{E}_{{\cal P}}\left\{ {\cal L}\left(y,\hat{y}\left(\boldsymbol{x};\underbar{\ensuremath{\rho}}\right)\right)\right\} +c\left(\epsilon\right)\sum_{\ell=1}^{L}\frac{1}{\sqrt{n_{\ell}}}+\epsilon,
\]
which immediately gives Eq. (\ref{eq:multilayer-statics-2}) by taking
$n\to\infty$ and then $\epsilon\to0$. To that end, for each $\ell=2,...,L$,
let us define $\hat{y}_{n}^{\left(\ell\right)}:\;\mathbb{R}^{n_{\ell}}\mapsto\mathbb{R}$
such that 
\[
\hat{y}_{n}^{\left(\ell\right)}\left(\boldsymbol{h}_{\ell}\right)=\hat{y}_{n}\left(\boldsymbol{x};{\cal W}\right).
\]
Here recall that
\begin{align*}
\hat{y}_{n}\left(\boldsymbol{x};{\cal W}\right) & =\frac{1}{n_{L}}\sum_{i=1}^{n_{L}}\sigma_{L}\left(\boldsymbol{\theta}_{L,i},h_{L,i}\right),\\
\boldsymbol{h}_{\ell} & =\frac{1}{n_{\ell-1}}\boldsymbol{W}_{\ell}\sigma_{\ell-1}\left(\boldsymbol{\Theta}_{\ell-1},\boldsymbol{h}_{\ell-1}\right),\qquad\ell=2,...,L,\\
\boldsymbol{h}_{1} & =\boldsymbol{W}_{1}\boldsymbol{x}.
\end{align*}
Note that in the above definition, $\hat{y}_{n}^{\left(\ell\right)}$
depends on $\boldsymbol{\Theta}_{\ell},...,\boldsymbol{\Theta}_{L},\boldsymbol{W}_{\ell+1},...,\boldsymbol{W}_{L}$
and $\boldsymbol{x}$, which are not displayed to lighten the notation.
In the following, the dependency on ${\cal W}$ or $\text{\ensuremath{\underbar{\ensuremath{\rho}}}}$
is also not displayed. The derivation of Eq. (\ref{eq:multilayer-statics-derive-2-1})
contains several steps.

\paragraph{Step 1.}

We argue that for each $\ell\geq2$ and any $\boldsymbol{u},\boldsymbol{v}\in\mathbb{R}^{n_{\ell}}$,
\begin{equation}
\left|\hat{y}_{n}^{\left(\ell\right)}\left(\boldsymbol{u}\right)-\hat{y}_{n}^{\left(\ell\right)}\left(\boldsymbol{v}\right)\right|\leq\frac{c\left(\epsilon\right)}{\sqrt{n_{\ell}}}\left\Vert \boldsymbol{u}-\boldsymbol{v}\right\Vert _{2}.\label{eq:multilayer-statics-derive-2-2}
\end{equation}
First, we have:
\begin{align*}
\left|\hat{y}_{n}^{\left(L\right)}\left(\boldsymbol{u}\right)-\hat{y}_{n}^{\left(L\right)}\left(\boldsymbol{v}\right)\right| & =\left|\frac{1}{n_{L}}\sum_{i=1}^{n_{L}}\sigma_{L}\left(\boldsymbol{\theta}_{L,i},u_{i}\right)-\frac{1}{n_{L}}\sum_{i=1}^{n_{L}}\sigma_{L}\left(\boldsymbol{\theta}_{L,i},v_{i}\right)\right|\\
 & \stackrel{\left(a\right)}{\leq}\frac{C}{n_{L}}\sum_{i=1}^{n_{L}}\left(1+\left\Vert \boldsymbol{\theta}_{L,i}\right\Vert _{2}\right)\left|u_{i}-v_{i}\right|\\
 & \leq\frac{C}{n_{L}}\sqrt{n_{L}+\left\Vert \boldsymbol{\Theta}_{L}\right\Vert _{{\rm F}}^{2}}\left\Vert \boldsymbol{u}-\boldsymbol{v}\right\Vert _{2}\\
 & \stackrel{\left(b\right)}{\leq}\frac{c\left(\epsilon\right)}{\sqrt{n_{L}}}\left\Vert \boldsymbol{u}-\boldsymbol{v}\right\Vert _{2},
\end{align*}
where $\left(a\right)$ is from Assumption (\ref{eq:multilayer-statics-assum_sL-1}),
and $\left(b\right)$ is from Assumption (\ref{eq:multilayer-statics-assum_inf-2}).
We then show the thesis by induction. Indeed, assuming the claim for
$\ell+1$, we have:
\begin{align*}
\left|\hat{y}_{n}^{\left(\ell\right)}\left(\boldsymbol{u}\right)-\hat{y}_{n}^{\left(\ell\right)}\left(\boldsymbol{v}\right)\right| & =\left|\hat{y}_{n}^{\left(\ell+1\right)}\left(\frac{1}{n_{\ell}}\boldsymbol{W}_{\ell+1}\sigma_{\ell}\left(\boldsymbol{\Theta}_{\ell},\boldsymbol{u}\right)\right)-\hat{y}_{n}^{\left(\ell+1\right)}\left(\frac{1}{n_{\ell}}\boldsymbol{W}_{\ell+1}\sigma_{\ell}\left(\boldsymbol{\Theta}_{\ell},\boldsymbol{v}\right)\right)\right|\\
 & \stackrel{\left(a\right)}{\leq}\frac{c\left(\epsilon\right)}{n_{\ell}\sqrt{n_{\ell+1}}}\left\Vert \boldsymbol{W}_{\ell+1}\sigma_{\ell}\left(\boldsymbol{\Theta}_{\ell},\boldsymbol{u}\right)-\boldsymbol{W}_{\ell+1}\sigma_{\ell}\left(\boldsymbol{\Theta}_{\ell},\boldsymbol{v}\right)\right\Vert _{2}\\
 & \stackrel{\left(b\right)}{\leq}\frac{c\left(\epsilon\right)}{n_{\ell}\sqrt{n_{\ell+1}}}\left\Vert \boldsymbol{W}_{\ell+1}\right\Vert _{{\rm F}}\left\Vert \boldsymbol{u}-\boldsymbol{v}\right\Vert _{2}\\
 & \stackrel{\left(c\right)}{\leq}\frac{c\left(\epsilon\right)}{\sqrt{n_{\ell}}}\left\Vert \boldsymbol{u}-\boldsymbol{v}\right\Vert _{2},
\end{align*}
where $\left(a\right)$ is by the induction hypothesis, $\left(b\right)$
is by Assumption (\ref{eq:multilayer-statics-assum_s_ell-1}), and
$\left(c\right)$ is from Assumption (\ref{eq:multilayer-statics-assum_inf-2}).
This shows the thesis.

\paragraph{Step 2.}

We argue that
\begin{equation}
\mathbb{E}\left\{ \left|\hat{y}_{n}\left(\boldsymbol{x}\right)-\hat{y}\left(\boldsymbol{x}\right)\right|\right\} \leq c\left(\epsilon\right)\sum_{\ell=1}^{L}\frac{1}{\sqrt{n_{\ell}}}.\label{eq:multilayer-statics-derive-2-3}
\end{equation}
Notice that
\begin{align*}
\hat{y}_{n}^{\left(L\right)}\left(\left(H_{L}\left(f_{L,i};\boldsymbol{x}\right)\right)_{i\in\left[n_{L}\right]}\right) & =\frac{1}{n_{L}}\sum_{i=1}^{n_{L}}\sigma_{L}\left(\boldsymbol{\theta}_{L,i},H_{L}\left(f_{L,i};\boldsymbol{x}\right)\right),\\
\hat{y}_{n}^{\left(\ell\right)}\left(\left(H_{\ell}\left(f_{\ell,i};\boldsymbol{x}\right)\right)_{i\in\left[n_{\ell}\right]}\right) & =\hat{y}_{n}^{\left(\ell+1\right)}\left(\frac{1}{n_{\ell}}\boldsymbol{W}_{\ell+1}\sigma_{\ell}\left(\boldsymbol{\Theta}_{\ell},\left(H_{\ell}\left(f_{\ell,i};\boldsymbol{x}\right)\right)_{i\in\left[n_{\ell}\right]}\right)\right),\quad\ell=2,...,L-1,\\
\hat{y}_{n}^{\left(2\right)}\left(\frac{1}{n_{1}}\boldsymbol{W}_{2}\sigma_{1}\left(\boldsymbol{\Theta}_{1},\boldsymbol{h}_{1}\right)\right) & =\hat{y}_{n}\left(\boldsymbol{x}\right).
\end{align*}
We thus have the following decomposition:
\begin{align*}
 & \mathbb{E}\left\{ \left|\hat{y}_{n}\left(\boldsymbol{x}\right)-\hat{y}\left(\boldsymbol{x}\right)\right|\right\} \\
 & \leq\mathbb{E}\left\{ \left|\hat{y}\left(\boldsymbol{x}\right)-\frac{1}{n_{L}}\sum_{i=1}^{n_{L}}\sigma_{L}\left(\boldsymbol{\theta}_{L,i},H_{L}\left(f_{L,i};\boldsymbol{x}\right)\right)\right|\right\} \\
 & \qquad+\sum_{\ell=2}^{L-1}\mathbb{E}\left\{ \left|\hat{y}_{n}^{\left(\ell+1\right)}\left(\left(H_{\ell+1}\left(f_{\ell+1,i};\boldsymbol{x}\right)\right)_{i\in\left[n_{\ell+1}\right]}\right)-\hat{y}_{n}^{\left(\ell+1\right)}\left(\frac{1}{n_{\ell}}\boldsymbol{W}_{\ell+1}\sigma_{\ell}\left(\boldsymbol{\Theta}_{\ell},\left(H_{\ell}\left(f_{\ell,i};\boldsymbol{x}\right)\right)_{i\in\left[n_{\ell}\right]}\right)\right)\right|\right\} \\
 & \qquad+\mathbb{E}\left\{ \left|\hat{y}_{n}^{\left(2\right)}\left(\left(H_{2}\left(f_{2,i};\boldsymbol{x}\right)\right)_{i\in\left[n_{2}\right]}\right)-\hat{y}_{n}^{\left(2\right)}\left(\frac{1}{n_{1}}\boldsymbol{W}_{2}\sigma_{1}\left(\boldsymbol{\Theta}_{1},\boldsymbol{h}_{1}\right)\right)\right|\right\} \\
 & \equiv A_{L}+\sum_{\ell=2}^{L-1}A_{\ell}+A_{1}.
\end{align*}
From Eq. (\ref{eq:multilayer-statics-derive-2-2}), we have for $\ell=2,...,L-1$:
\begin{align*}
A_{\ell}^{2} & \leq\mathbb{E}\left\{ \left|\hat{y}_{n}^{\left(\ell+1\right)}\left(\left(H_{\ell+1}\left(f_{\ell+1,i};\boldsymbol{x}\right)\right)_{i\in\left[n_{\ell+1}\right]}\right)-\hat{y}_{n}^{\left(\ell+1\right)}\left(\frac{1}{n_{\ell}}\boldsymbol{W}_{\ell+1}\sigma_{\ell}\left(\boldsymbol{\Theta}_{\ell},\left(H_{\ell}\left(f_{\ell,i};\boldsymbol{x}\right)\right)_{i\in\left[n_{\ell}\right]}\right)\right)\right|^{2}\right\} \\
 & \leq\frac{c\left(\epsilon\right)}{n_{\ell+1}}\sum_{i=1}^{n_{\ell+1}}\mathbb{E}\left\{ \left(H_{\ell+1}\left(f_{\ell+1,i};\boldsymbol{x}\right)-\frac{1}{n_{\ell}}\sum_{j=1}^{n_{\ell}}f_{\ell+1,i}\left(\boldsymbol{\theta}_{\ell,j},f_{\ell,j}\right)\sigma_{\ell}\left(\boldsymbol{\theta}_{\ell,j},H_{\ell}\left(f_{\ell,j};\boldsymbol{x}\right)\right)\right)^{2}\right\} \\
 & \stackrel{\left(a\right)}{=}\frac{c\left(\epsilon\right)}{n_{\ell}n_{\ell+1}}\sum_{i=1}^{n_{\ell+1}}\mathbb{E}\left\{ {\rm Var}_{\ell}\left\{ f_{\ell+1,i}\left(\boldsymbol{\theta},f\right)\sigma_{\ell}\left(\boldsymbol{\theta},H_{\ell}\left(f;\boldsymbol{x}\right)\right)\right\} \right\} \\
 & \leq\frac{c\left(\epsilon\right)}{n_{\ell}n_{\ell+1}}\sum_{i=1}^{n_{\ell+1}}\mathbb{E}\left\{ \mathbb{E}_{\ell}\left\{ f_{\ell+1,i}^{2}\left(\boldsymbol{\theta},f\right)\sigma_{\ell}^{2}\left(\boldsymbol{\theta},H_{\ell}\left(f;\boldsymbol{x}\right)\right)\right\} \right\} \\
 & \stackrel{\left(b\right)}{\leq}\frac{c\left(\epsilon\right)}{n_{\ell}}
\end{align*}
where ${\rm Var}_{\ell}$ and $\mathbb{E}_{\ell}$ indicate the variance
and the mean w.r.t. $\left(\boldsymbol{\theta},f\right)\sim\rho_{\ell}$
. Here the factor $1/n_{\ell}$ in step $\left(a\right)$ is due to
$\left\{ \left(\boldsymbol{\theta}_{\ell,i},f_{\ell,i}\right)\right\} _{i\in\left[n_{\ell}\right]}\sim\rho_{\ell}$
i.i.d., and step $\left(b\right)$ is due to Assumptions (\ref{eq:multilayer-statics-assum_inf-2})
and (\ref{eq:multilayer-statics-assum_s_ell_2}). Similarly we have
$A_{1}^{2}\leq c\left(\epsilon\right)/n_{1}$ and $A_{L}^{2}\leq c\left(\epsilon\right)/n_{L}$.
The thesis then follows.

\paragraph{Step 3.}

From Assumptions (\ref{eq:multilayer-statics-assum_L}) and (\ref{eq:multilayer-statics-assum_y}),
we have:
\[
\left|\mathbb{E}_{{\cal W}}\left\{ \mathbb{E}_{{\cal P}}\left\{ {\cal L}\left(y,\hat{y}_{n}\left(\boldsymbol{x}\right)\right)\right\} \right\} -\mathbb{E}_{{\cal P}}\left\{ {\cal L}\left(y,\hat{y}\left(\boldsymbol{x}\right)\right)\right\} \right|\leq C\mathbb{E}\left\{ \left(1+\left|\hat{y}_{n}\left(\boldsymbol{x}\right)\right|+\left|\hat{y}\left(\boldsymbol{x}\right)\right|\right)\left|\hat{y}_{n}\left(\boldsymbol{x}\right)-\hat{y}\left(\boldsymbol{x}\right)\right|\right\} .
\]
Notice that from Assumptions (\ref{eq:multilayer-statics-assum_sL-2})
and (\ref{eq:multilayer-statics-assum_inf-2}),
\[
\left|\hat{y}_{n}\left(\boldsymbol{x}\right)\right|\leq\frac{1}{n_{L}}\sum_{i=1}^{n_{L}}\left|\sigma_{L}\left(\boldsymbol{\theta}_{L,i},h_{L,i}\right)\right|\leq\frac{C}{n_{L}}\sum_{i=1}^{n_{L}}\left(1+\left\Vert \boldsymbol{\theta}_{L,i}\right\Vert _{2}\right)\leq c\left(\epsilon\right).
\]
Similarly, $\left|\hat{y}\left(\boldsymbol{x}\right)\right|\leq c\left(\epsilon\right)$.
Combining with Eq. (\ref{eq:multilayer-statics-derive-2-3}), we arrive
at Eq. (\ref{eq:multilayer-statics-derive-2-1}).

\section{Mean field limit in multilayer convolutional neural networks\label{sec:LLN-CNNs}}

We consider convolutional neural networks (CNNs), which are most interesting
when they have many layers. In a CNN, the number of neurons $n_{\ell}$
at layer $\ell$ is the number of filters at that layer. A simple
convolutional analog of the three-layers fully-connected network (\ref{eq:three-layers-net})
can be described by the following:
\[
\hat{y}_{n}\left(\boldsymbol{x};{\cal W}\right)=\frac{1}{n_{2}}\left\langle \boldsymbol{\beta},\sigma_{2}\left(\boldsymbol{h}_{2}\right)\right\rangle ,\quad\boldsymbol{h}_{2}=\frac{1}{n_{1}}\boldsymbol{W}_{2}\circledast\sigma_{1}\left(\boldsymbol{h}_{1}\right),\quad\boldsymbol{h}_{1}=\boldsymbol{W}_{1}\circledast\boldsymbol{x}.
\]
Here $\circledast$ is the operator defined by
\[
\boldsymbol{W}\circledast\boldsymbol{u}=\left(\sum_{j=1}^{m_{1}}w_{ij}*u_{j}\right)_{i\in\left[m_{2}\right]},\qquad\boldsymbol{u}\in\left(\mathbb{R}^{p}\right)^{m_{1}},\quad\boldsymbol{W}=\left(w_{ij}\right)_{i\in\left[m_{2}\right],j\in\left[m_{1}\right]}\in\left(\mathbb{R}^{s}\right)^{m_{2}\times m_{1}},
\]
for a convolutional operator $*$, in which $w*u\in\mathbb{R}^{r\left(s,p\right)}$
for any $u\in\mathbb{R}^{p}$ and $w\in\mathbb{R}^{s}$, and $r\left(s,p\right)$
is an integer to be determined by the exact operation of $*$. It
can be made $r\left(s,p\right)=p$ with appropriate paddings and no
striding. In our context,
\begin{itemize}
\item $\boldsymbol{x}\in\left(\mathbb{R}^{p}\right)^{d}$ a $p$-pixels
$d$-channels ($1$-dimensional) input image (e.g. $d=3$ for an RGB
image),
\item $\boldsymbol{W}_{1}\in\left(\mathbb{R}^{s_{1}}\right)^{n_{1}\times d}$
where each entry is an element in $\mathbb{R}^{s_{1}}$ and the receptive
field size is $s_{1}$,
\item $\boldsymbol{h}_{1}\in\left(\mathbb{R}^{r\left(s_{1},p\right)}\right)^{n_{1}}$,
$\sigma_{1}:\;\mathbb{R}^{r\left(s_{1},p\right)}\mapsto\mathbb{R}^{p_{1}}$
some nonlinear mapping for $p_{1}=p_{1}\left(r\left(s_{1},p\right)\right)$,
and $\sigma_{1}\left(\boldsymbol{h}_{1}\right)\in\left(\mathbb{R}^{p_{1}}\right)^{n_{1}}$
element-wise,
\item $\boldsymbol{W}_{2}\in\left(\mathbb{R}^{s_{2}}\right)^{n_{2}\times n_{1}}$
where each entry is an element in $\mathbb{R}^{s_{2}}$ and the receptive
field size is $s_{2}$,
\item $\boldsymbol{h}_{2}\in\left(\mathbb{R}^{r\left(s_{2},p_{1}\right)}\right)^{n_{2}}$,
$\sigma_{2}:\;\mathbb{R}^{r\left(s_{2},p_{1}\right)}\mapsto\mathbb{R}^{p_{2}}$
some nonlinear mapping for $p_{2}=p_{2}\left(r\left(s_{2},p_{1}\right)\right)$,
and $\sigma_{2}\left(\boldsymbol{h}_{2}\right)\in\left(\mathbb{R}^{p_{2}}\right)^{n_{2}}$
element-wise,
\item $\boldsymbol{\beta}\in\mathbb{R}^{n_{2}p_{2}}$ and $\left\langle \cdot,\cdot\right\rangle $
computes the usual Euclidean inner product after vectorizing its arguments.
\end{itemize}
The nonlinear mapping can be, as in the usual practice, a composition
of a pooling operation and a scalar nonlinear activation. Now observe
the similarity between this network and its fully-connected counterpart,
especially the summation structure shared by both, as evident from
the definition of the $\circledast$ operator. Recall that this summation
structure is key to the self-averaging property in light of Eq. (\ref{eq:self-averaging}).
The only difference is that local operations (such as the $*$ operator)
are no longer scalar-valued, but rather vector-valued (or matrix-valued).
As discussed in Section \ref{subsec:Discussions-three-layers}, we
expect that this difference is not very critical and the MF limit
behavior still occurs, provided that the number of filters $n_{\ell}\to\infty$
while all other dimensions are kept constant. The scalings can be
deduced from Section \ref{subsec:multilayer-setting}.

In the following, we present an experimental validation of the existence
of the MF limit in multilayer CNNs on the CIFAR-10 classification
task. We construct 8-layers networks according to Table \ref{tab:CNN}.
Note that we apply a stride of size $\left(2,2\right)$ in the first
two layers, which reduces the spatial dimensions by a considerable
amount and hence limits memory consumption at the cost of the networks'
performance. We use the same number of filters at each layer $n$,
where $n$ is to be varied among the networks. We normalize each RGB
value in the image to the range $\left[-1,+1\right]$. We use the
cross-entropy loss ${\cal L}$, and use the whole training set of
size $50\times10^{3}$. To train the networks, we use mini-batch SGD
with an annealed learning rate $\alpha_{k}=0.08k^{-0.1}$, where $k\geq1$
is the SGD iteration, and a batch size of $100$. We initialize the
networks in a similar fashion to those in Section \ref{subsec:Dynamics-validation},
i.e. the first layer weight entries are initialized with $\mathsf{N}\left(0,2/\left(9d\right)\right)$,
the other layers' weight entries are initialized with $\mathsf{N}\left(1,0.1\right)$,
and all biases are initialized to zero.

\begin{figure}
\begin{centering}
\includegraphics[width=1\columnwidth]{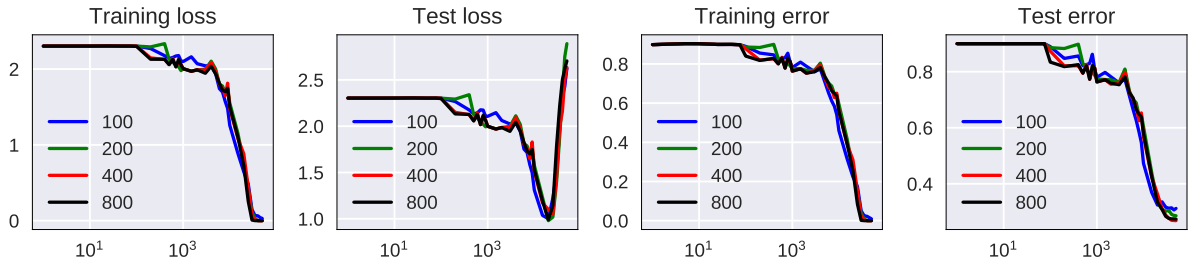}
\par\end{centering}
\caption{The performance of 8-layers CNNs on CIFAR-10 classification, plotted
against training iteration. For each network, $n=100,200,400,800$
respectively.}
\label{fig:cifar-CNN}
\end{figure}

The result is shown in Fig. \ref{fig:cifar-CNN}. We observe the good
match among the networks \textendash{} the larger $n$, the better
match. The performance is also realistic: the test error rate for
the network with $n=400$ is about $27\%$. This is similar to the
performance of a 259-layers CNN reported in \cite{xiao2018dynamical},
which attains a test error rate of $30\%$. It has a similar vanilla
structure, is initialized with i.i.d. Gaussian weights of zero mean
and carefully selected variance, is trained without regularization,
but is not under our scalings. Hence the introduced scalings do not
trivialize the performance of the networks.

\begin{table}
\begin{centering}
\begin{tabular}{|c|c|c|}
\hline 
\multirow{2}{*}{Layer} & \multirow{2}{*}{Structure} & Output\tabularnewline
 &  & spatial dimension\tabularnewline
\hline 
\hline 
1 & CONV-$\left(3,3,n\right)$, stride $\left(2,2\right)$ \textemdash{}
ReLU & $16\times16$\tabularnewline
\hline 
2 & CONV-$\left(3,3,n\right)$, stride $\left(2,2\right)$ \textemdash{}
ReLU & $8\times8$\tabularnewline
\hline 
3 & CONV-$\left(3,3,n\right)$, stride $\left(1,1\right)$ \textemdash{}
ReLU & $8\times8$\tabularnewline
\hline 
4 & CONV-$\left(3,3,n\right)$, stride $\left(1,1\right)$ \textemdash{}
ReLU & $8\times8$\tabularnewline
\hline 
5 & CONV-$\left(3,3,n\right)$, stride $\left(1,1\right)$ \textemdash{}
ReLU \textemdash{} POOL-$\left(3,3\right)$, stride $\left(2,2\right)$ & $4\times4$\tabularnewline
\hline 
6 & CONV-$\left(3,3,n\right)$, stride $\left(1,1\right)$ \textemdash{}
ReLU & $4\times4$\tabularnewline
\hline 
7 & CONV-$\left(3,3,n\right)$, stride $\left(1,1\right)$ \textemdash{}
ReLU \textemdash{} POOL-$\left(3,3\right)$, stride $\left(2,2\right)$ & $2\times2$\tabularnewline
\hline 
8 & FC-$\left(10,4n\right)$ & \textendash{}\tabularnewline
\hline 
\end{tabular}
\par\end{centering}
\caption{Structure of the CNNs. Here CONV-$\left(s,s,n\right)$ is a convolutional
layer with a receptive field size of $s$ and $n$ filters. POOL-$\left(s,s\right)$
is the max pooling operation over a spatial region of size $s\times s$.
Stride $\left(s,s\right)$ is the stride of size $s$ in each dimension,
applied to the accompanied operation. FC-$\left(10,4n\right)$ is
a fully-connected layer with dimensions $10\times4n$. ReLU indicates
an entry-wise rectifier linear unit nonlinearity. All layers have
trainable biases. We apply appropriate paddings to obtain the corresponding
output spatial dimension.}

\label{tab:CNN}
\end{table}

\bibliographystyle{amsalpha}
\bibliography{LLN_multilayer_nnet}

\newcommand{\etalchar}[1]{$^{#1}$}
\providecommand{\bysame}{\leavevmode\hbox to3em{\hrulefill}\thinspace}
\providecommand{\MR}{\relax\ifhmode\unskip\space\fi MR }
\providecommand{\MRhref}[2]{%
  \href{http://www.ams.org/mathscinet-getitem?mr=#1}{#2}
}
\providecommand{\href}[2]{#2}
\begin{thebibliography}{dGMHR{\etalchar{+}}18}

\bibitem[ABGM14]{arora2014provable}
Sanjeev Arora, Aditya Bhaskara, Rong Ge, and Tengyu Ma, \emph{Provable bounds
  for learning some deep representations}, International Conference on Machine
  Learning, 2014, pp.~584--592.

\bibitem[AZLL18]{allen2018learning}
Zeyuan Allen-Zhu, Yuanzhi Li, and Yingyu Liang, \emph{Learning and
  generalization in overparameterized neural networks, going beyond two
  layers}, arXiv preprint arXiv:1811.04918 (2018).

\bibitem[AZLS18]{allen2018convergence}
Zeyuan Allen-Zhu, Yuanzhi Li, and Zhao Song, \emph{A convergence theory for
  deep learning via over-parameterization}, arXiv preprint arXiv:1811.03962
  (2018).

\bibitem[BRV{\etalchar{+}}06]{bengio2006convex}
Yoshua Bengio, Nicolas~L Roux, Pascal Vincent, Olivier Delalleau, and Patrice
  Marcotte, \emph{Convex neural networks}, Advances in neural information
  processing systems, 2006, pp.~123--130.

\bibitem[CB18a]{chizat2018note}
Lenaic Chizat and Francis Bach, \emph{A note on lazy training in supervised
  differentiable programming}, arXiv preprint arXiv:1812.07956 (2018).

\bibitem[CB18b]{chizat2018}
L\'{e}na\"{\i}c Chizat and Francis Bach, \emph{On the global convergence of
  gradient descent for over-parameterized models using optimal transport},
  Advances in Neural Information Processing Systems, 2018, pp.~3040--3050.

\bibitem[CHM{\etalchar{+}}15]{choromanska2015loss}
Anna Choromanska, Mikael Henaff, Michael Mathieu, G{\'e}rard~Ben Arous, and
  Yann LeCun, \emph{The loss surfaces of multilayer networks}, Artificial
  Intelligence and Statistics, 2015, pp.~192--204.

\bibitem[Coo18]{cooper2018loss}
Yaim Cooper, \emph{The loss landscape of overparameterized neural networks},
  arXiv preprint arXiv:1804.10200 (2018).

\bibitem[CPS18]{chen2018dynamical}
Minmin Chen, Jeffrey Pennington, and Samuel Schoenholz, \emph{Dynamical
  isometry and a mean field theory of {RNN}s: Gating enables signal propagation
  in recurrent neural networks}, Proceedings of the 35th International
  Conference on Machine Learning, vol.~80, 2018, pp.~873--882.

\bibitem[dGMHR{\etalchar{+}}18]{matthews2018gaussian}
Alexander~G. de~G.~Matthews, Jiri Hron, Mark Rowland, Richard~E. Turner, and
  Zoubin Ghahramani, \emph{Gaussian process behaviour in wide deep neural
  networks}, International Conference on Learning Representations, 2018.

\bibitem[DL18]{du2018power}
Simon Du and Jason Lee, \emph{On the power of over-parametrization in neural
  networks with quadratic activation}, Proceedings of the 35th International
  Conference on Machine Learning, vol.~80, 2018, pp.~1329--1338.

\bibitem[DLL{\etalchar{+}}18]{du2018gradientDeep}
Simon~S Du, Jason~D Lee, Haochuan Li, Liwei Wang, and Xiyu Zhai, \emph{Gradient
  descent finds global minima of deep neural networks}, arXiv preprint
  arXiv:1811.03804 (2018).

\bibitem[DZPS19]{du2018gradient}
Simon~S. Du, Xiyu Zhai, Barnabas Poczos, and Aarti Singh, \emph{Gradient
  descent provably optimizes over-parameterized neural networks}, International
  Conference on Learning Representations, 2019.

\bibitem[FB17]{freeman2016topology}
C~Daniel Freeman and Joan Bruna, \emph{Topology and geometry of half-rectified
  network optimization}, International Conference on Learning Representations
  (2017).

\bibitem[GARA19]{garriga2018deep}
Adri{\`a} Garriga-Alonso, Carl~Edward Rasmussen, and Laurence Aitchison,
  \emph{Deep convolutional networks as shallow gaussian processes},
  International Conference on Learning Representations, 2019.

\bibitem[GJS{\etalchar{+}}19]{geiger2019scaling}
Mario Geiger, Arthur Jacot, Stefano Spigler, Franck Gabriel, Levent Sagun,
  St{\'e}phane d'Ascoli, Giulio Biroli, Cl{\'e}ment Hongler, and Matthieu
  Wyart, \emph{Scaling description of generalization with number of parameters
  in deep learning}, arXiv preprint arXiv:1901.01608 (2019).

\bibitem[GSd{\etalchar{+}}18]{geiger2018jamming}
Mario Geiger, Stefano Spigler, St{\'e}phane d'Ascoli, Levent Sagun, Marco
  Baity-Jesi, Giulio Biroli, and Matthieu Wyart, \emph{The jamming transition
  as a paradigm to understand the loss landscape of deep neural networks},
  arXiv preprint arXiv:1809.09349 (2018).

\bibitem[Han18]{hanin2018neural}
Boris Hanin, \emph{Which neural net architectures give rise to exploding and
  vanishing gradients?}, Advances in Neural Information Processing Systems 31,
  2018, pp.~580--589.

\bibitem[HJ15]{hazan2015steps}
Tamir Hazan and Tommi Jaakkola, \emph{Steps toward deep kernel methods from
  infinite neural networks}, arXiv preprint arXiv:1508.05133 (2015).

\bibitem[HNP{\etalchar{+}}18]{ho2018neural}
Nhat Ho, Tan Nguyen, Ankit Patel, Anima Anandkumar, Michael~I Jordan, and
  Richard~G Baraniuk, \emph{Neural rendering model: Joint generation and
  prediction for semi-supervised learning}, arXiv preprint arXiv:1811.02657
  (2018).

\bibitem[HR18]{hanin2018start}
Boris Hanin and David Rolnick, \emph{How to start training: The effect of
  initialization and architecture}, Advances in Neural Information Processing
  Systems 31, 2018, pp.~569--579.

\bibitem[JGH18]{jacot2018neural}
Arthur Jacot, Franck Gabriel, and Clement Hongler, \emph{Neural tangent kernel:
  Convergence and generalization in neural networks}, Advances in Neural
  Information Processing Systems 31, 2018, pp.~8580--8589.

\bibitem[JMM19]{javanmard2019analysis}
Adel Javanmard, Marco Mondelli, and Andrea Montanari, \emph{Analysis of a
  two-layer neural network via displacement convexity}, arXiv preprint
  arXiv:1901.01375 (2019).

\bibitem[LBH15]{lecun2015deep}
Yann LeCun, Yoshua Bengio, and Geoffrey Hinton, \emph{Deep learning}, nature
  \textbf{521} (2015), no.~7553, 436.

\bibitem[LL18]{li2018learning}
Yuanzhi Li and Yingyu Liang, \emph{Learning overparameterized neural networks
  via stochastic gradient descent on structured data}, Advances in Neural
  Information Processing Systems, 2018, pp.~8168--8177.

\bibitem[LN19]{li2018on}
Ping Li and Phan-Minh Nguyen, \emph{On random deep weight-tied autoencoders:
  Exact asymptotic analysis, phase transitions, and implications to training},
  International Conference on Learning Representations, 2019.

\bibitem[LSdP{\etalchar{+}}18]{lee2017deep}
Jaehoon Lee, Jascha Sohl-dickstein, Jeffrey Pennington, Roman Novak, Sam
  Schoenholz, and Yasaman Bahri, \emph{Deep neural networks as gaussian
  processes}, International Conference on Learning Representations, 2018.

\bibitem[Mal16]{mallat2016understanding}
St{\'e}phane Mallat, \emph{Understanding deep convolutional networks},
  Philosophical Transactions of the Royal Society A: Mathematical, Physical and
  Engineering Sciences \textbf{374} (2016), no.~2065, 20150203.

\bibitem[MBM18]{mei2018landscape}
Song Mei, Yu~Bai, and Andrea Montanari, \emph{The landscape of empirical risk
  for nonconvex losses}, The Annals of Statistics \textbf{46} (2018), no.~6A,
  2747--2774.

\bibitem[MMN18]{mei2018mean}
Song Mei, Andrea Montanari, and Phan-Minh Nguyen, \emph{A mean field view of
  the landscape of two-layers neural networks}, Proceedings of the National
  Academy of Sciences, vol. 115, 2018, pp.~7665--7671.

\bibitem[MP16]{mhaskar2016deep}
Hrushikesh~N Mhaskar and Tomaso Poggio, \emph{Deep vs. shallow networks: An
  approximation theory perspective}, Analysis and Applications \textbf{14}
  (2016), no.~06, 829--848.

\bibitem[NH17]{nguyen2017loss}
Quynh Nguyen and Matthias Hein, \emph{The loss surface of deep and wide neural
  networks}, Proceedings of the 34th International Conference on Machine
  Learning, vol.~70, 2017, pp.~2603--2612.

\bibitem[NH18]{nguyen2018optimization}
Quynh Nguyen and Matthias Hein, \emph{Optimization landscape and expressivity
  of deep cnns}, International Conference on Machine Learning, 2018,
  pp.~3727--3736.

\bibitem[NXB{\etalchar{+}}19]{novak2018bayesian}
Roman Novak, Lechao Xiao, Yasaman Bahri, Jaehoon Lee, Greg Yang, Daniel~A.
  Abolafia, Jeffrey Pennington, and Jascha Sohl-dickstein, \emph{Bayesian deep
  convolutional networks with many channels are gaussian processes},
  International Conference on Learning Representations, 2019.

\bibitem[PLR{\etalchar{+}}16]{poole2016exponential}
Ben Poole, Subhaneil Lahiri, Maithra Raghu, Jascha Sohl-Dickstein, and Surya
  Ganguli, \emph{Exponential expressivity in deep neural networks through
  transient chaos}, Advances in neural information processing systems, 2016,
  pp.~3360--3368.

\bibitem[PSG17]{pennington2017resurrecting}
Jeffrey Pennington, Samuel Schoenholz, and Surya Ganguli, \emph{Resurrecting
  the sigmoid in deep learning through dynamical isometry: theory and
  practice}, Advances in neural information processing systems, 2017,
  pp.~4785--4795.

\bibitem[RDS{\etalchar{+}}15]{russakovsky2015imagenet}
Olga Russakovsky, Jia Deng, Hao Su, Jonathan Krause, Sanjeev Satheesh, Sean Ma,
  Zhiheng Huang, Andrej Karpathy, Aditya Khosla, Michael Bernstein, Alexander
  Berg, and Fei-Fei Li, \emph{Imagenet large scale visual recognition
  challenge}, International Journal of Computer Vision \textbf{115} (2015),
  no.~3, 211--252.

\bibitem[RVE18]{rotskoff2018neural}
Grant~M Rotskoff and Eric Vanden-Eijnden, \emph{Neural networks as interacting
  particle systems: Asymptotic convexity of the loss landscape and universal
  scaling of the approximation error}, arXiv preprint arXiv:1805.00915 (2018).

\bibitem[SC16]{soudry2016no}
Daniel Soudry and Yair Carmon, \emph{No bad local minima: Data independent
  training error guarantees for multilayer neural networks}, arXiv preprint
  arXiv:1605.08361 (2016).

\bibitem[SGd{\etalchar{+}}18]{spigler2018jamming}
Stefano Spigler, Mario Geiger, St{\'e}phane d'Ascoli, Levent Sagun, Giulio
  Biroli, and Matthieu Wyart, \emph{A jamming transition from under-to
  over-parametrization affects loss landscape and generalization}, arXiv
  preprint arXiv:1810.09665 (2018).

\bibitem[SGGSD17]{schoenholz2016deep}
Samuel~S Schoenholz, Justin Gilmer, Surya Ganguli, and Jascha Sohl-Dickstein,
  \emph{Deep information propagation}, International Conference on Learning
  Representations (2017).

\bibitem[SJL18]{soltanolkotabi2018theoretical}
Mahdi Soltanolkotabi, Adel Javanmard, and Jason~D Lee, \emph{Theoretical
  insights into the optimization landscape of over-parameterized shallow neural
  networks}, IEEE Transactions on Information Theory (2018).

\bibitem[SS16]{safran2016quality}
Itay Safran and Ohad Shamir, \emph{On the quality of the initial basin in
  overspecified neural networks}, International Conference on Machine Learning,
  2016, pp.~774--782.

\bibitem[SS18a]{sirignano2018mean}
Justin Sirignano and Konstantinos Spiliopoulos, \emph{Mean field analysis of
  neural networks}, arXiv preprint arXiv:1805.01053 (2018).

\bibitem[SS18b]{sirignano2018CLT}
\bysame, \emph{Mean field analysis of neural networks: A central limit
  theorem}, arXiv preprint arXiv:1808.09372 (2018).

\bibitem[SZ15]{simonyan2014very}
Karen Simonyan and Andrew Zisserman, \emph{Very deep convolutional networks for
  large-scale image recognition}, International Conference on Learning
  Representations (2015).

\bibitem[SZT17]{shwartz2017opening}
Ravid Shwartz-Ziv and Naftali Tishby, \emph{Opening the black box of deep
  neural networks via information}, arXiv preprint arXiv:1703.00810 (2017).

\bibitem[VBB18]{venturi2018spurious}
Luca Venturi, Afonso Bandeira, and Joan Bruna, \emph{Spurious valleys in
  two-layer neural network optimization landscapes}, arXiv preprint
  arXiv:1802.06384 (2018).

\bibitem[WLLM18]{wei2018margin}
Colin Wei, Jason~D Lee, Qiang Liu, and Tengyu Ma, \emph{On the margin theory of
  feedforward neural networks}, arXiv preprint arXiv:1810.05369 (2018).

\bibitem[XBSD{\etalchar{+}}18]{xiao2018dynamical}
Lechao Xiao, Yasaman Bahri, Jascha Sohl-Dickstein, Samuel Schoenholz, and
  Jeffrey Pennington, \emph{Dynamical isometry and a mean field theory of
  {CNN}s: How to train 10,000-layer vanilla convolutional neural networks},
  Proceedings of the 35th International Conference on Machine Learning,
  vol.~80, 2018, pp.~5393--5402.

\bibitem[YPR{\etalchar{+}}19]{yang2018a}
Greg Yang, Jeffrey Pennington, Vinay Rao, Jascha Sohl-Dickstein, and Samuel~S.
  Schoenholz, \emph{A mean field theory of batch normalization}, International
  Conference on Learning Representations, 2019.

\bibitem[YS17]{yang2017mean}
Ge~Yang and Samuel Schoenholz, \emph{Mean field residual networks: On the edge
  of chaos}, Advances in neural information processing systems, 2017,
  pp.~7103--7114.

\bibitem[YSJ19]{yun2018small}
Chulhee Yun, Suvrit Sra, and Ali Jadbabaie, \emph{Small nonlinearities in
  activation functions create bad local minima in neural networks},
  International Conference on Learning Representations, 2019.

\bibitem[ZCZG18]{zou2018stochastic}
Difan Zou, Yuan Cao, Dongruo Zhou, and Quanquan Gu, \emph{Stochastic gradient
  descent optimizes over-parameterized deep relu networks}, arXiv preprint
  arXiv:1811.08888 (2018).

\bibitem[ZSJ{\etalchar{+}}17]{zhong2017recovery}
Kai Zhong, Zhao Song, Prateek Jain, Peter~L. Bartlett, and Inderjit~S. Dhillon,
  \emph{Recovery guarantees for one-hidden-layer neural networks}, Proceedings
  of the 34th International Conference on Machine Learning, vol.~70, 2017,
  pp.~4140--4149.

\end{thebibliography}

\end{document}